\documentclass[letterpaper]{article} 
\usepackage{aaai23}  
\usepackage{times}  
\usepackage{helvet}  
\usepackage{courier}  
\usepackage[hyphens]{url}  
\usepackage{graphicx} 
\usepackage{natbib}  
\usepackage{caption} 
\usepackage{algorithm}
\usepackage{algorithmic}

\usepackage{newfloat}
\usepackage{listings}
\DeclareCaptionStyle{ruled}{labelfont=normalfont,labelsep=colon,strut=off} 
\floatstyle{ruled}
\newfloat{listing}{tb}{lst}{}
\floatname{listing}{Listing}
\title{CDMA: A Practical Cross-Device Federated Learning Algorithm for General Minimax Problems}

\author{Jiahao Xie\textsuperscript{\rm 1} ,
        Chao Zhang\thanks{Corresponding author.}\textsuperscript{\rm 2},
        Zebang Shen\textsuperscript{\rm 3},
        Weijie Liu\textsuperscript{\rm 4,1},
        Hui Qian\textsuperscript{\rm 1,5}\\
}
\affiliations{
\textsuperscript{\rm 1}College of Computer Science and Technology, Zhejiang University\\
\textsuperscript{\rm 2}Advanced Technology Institute, Zhejiang University \\
\textsuperscript{\rm 3}ETH Zürich\\
\textsuperscript{\rm 4}Qiushi Academy for Advanced Studies, Zhejiang University \\
\textsuperscript{\rm 5}State Key Lab of CAD\&CG, Zhejiang University\\
xiejh@zju.edu.cn, zczju@zju.edu.cn, zebang.shen@inf.ethz.ch, westonhunter@zju.edu.cn, qianhui@zju.edu.cn
}

\usepackage[caption=false]{subfig}
\usepackage{booktabs} 
\usepackage{threeparttable}
\usepackage{nicefrac}       
\usepackage{mathtools}
\usepackage{amsmath}
\usepackage{amssymb}
\usepackage{amsthm}
\usepackage{multirow}
\usepackage{subfiles}
\usepackage{etoolbox}
\usepackage{enumerate}
\usepackage[shortlabels]{enumitem}
\usepackage{siunitx}
\input{required.tex}

\newcommand{\FrameworkName}{Cross-Device Minimax Averaging}

\newcommand{\FrameworkAbbr}{CDMA}
\newcommand{\AlgNaiveFull}{{\FrameworkAbbr} with No Correction}
\newcommand{\AlgNaive}{CDMA-NC}
\newcommand{\AlgMBFull}{CDMA with $\beta = 1$ and $\alpha_t \equiv 1$}
\newcommand{\AlgMB}{CDMA-ONE}
\newcommand{\AlgSTORMFull}{CDMA with $\beta = 1$ and $\alpha_t \not\equiv 1$}
\newcommand{\AlgSTORM}{CDMA-ADA}

\newcommand{\Lot}{L_f}

\newcounter{framework}


\newcounter{estimator}


\newcommand{\SUB}[1]{\ENSURE \hspace{-0.15in} \textbf{#1}}

\newcommand{\algorithmicdoinparallel}{\textbf{do in parallel}}
\makeatletter
\AtBeginEnvironment{algorithmic}{%
  \newcommand{\FORALLP}[2][default]{\ALC@it\algorithmicforall\ #2\ %
    \algorithmicdoinparallel\ALC@com{#1}\begin{ALC@for}}%
}
\makeatother

\newcommand{\bftab}{\fontseries{b}\selectfont} 
\makeatletter
\patchcmd{\Ginclude@eps}{"#1"}{#1}{}{}
\makeatother

\allowdisplaybreaks

\begin{document}

\maketitle

\begin{abstract}
    Minimax problems arise in a wide range of important applications including robust adversarial learning and Generative Adversarial Network (GAN) training.
    Recently, algorithms for minimax problems in the Federated Learning (FL) paradigm have received considerable interest.
    Existing federated algorithms for general minimax problems require
    the full aggregation (i.e., aggregation of local model information from all clients) in each training round.
    Thus, they are inapplicable to an important setting of FL known as the cross-device setting, which involves numerous unreliable mobile/IoT devices.
    In this paper, we develop the first practical algorithm named CDMA for general minimax problems in the cross-device FL setting.
    CDMA is based on a Start-Immediately-With-Enough-Responses mechanism, in which the server first signals a subset of clients to perform local computation and then starts to aggregate the local results reported by clients once it receives responses from enough clients in each round.
    With this mechanism, CDMA is resilient to the low client availability.
    In addition, CDMA is incorporated with a lightweight global correction in the local update steps of clients, which mitigates the impact of slow network connections.
    We establish theoretical guarantees of CDMA under different choices of hyperparameters and conduct experiments on AUC maximization, robust adversarial network training, and GAN training tasks.
    Theoretical and experimental results demonstrate the efficiency of CDMA.
\end{abstract}


\section{Introduction}  \label{section_introduction}

During the last few years, minimax problems have found a surge of important applications such as AUC (area under the ROC curve) maximization~\citep{ying2016stochastic, liu2020stochastic}, robust adversarial learning~\cite{madry2018towards}, and Generative Adversarial Network (GAN)~\citep{goodfellow2014generative}.
Recently, distributed minimax algorithms in the Federated Learning (FL) paradigm have drawn significant attention due to the increasing concern of data privacy and the rapid growth of data volume~\citep{mohri2019agnostic, deng2020distributionally, reisizadeh2020robust, rasouli2020fedgan, beznosikov2020distributed, hou2021efficient, guo2020communication, yuan2021federated, deng2021local}.
Generally, these federated learning algorithms optimize a global minimax problem over multiple clients (data sources) under the coordination of a central server without transferring any client's private data~\cite{kairouz2019advances}, which helps protect data privacy.
Besides, the local update strategy (i.e., performing multiple local update steps on clients before communicating with the server) is usually adopted in each training round of these algorithms to save communication efforts, which makes them scalable to large-scale problems.

Despite that existing federated minimax algorithms have achieved success in certain applications, there is a lack of practical algorithms for general minimax problems in an important FL setting, known as the \emph{cross-device} setting.
In this setting, the clients are numerous (up to $10^{10}$~\cite{kairouz2019advances}) unreliable mobile/IoT devices with relatively slow network connections.
Actually, most existing federated minimax algorithms are designed for another setting known as the \emph{cross-silo} setting~\cite{reisizadeh2020robust, rasouli2020fedgan, beznosikov2020distributed, hou2021efficient, guo2020communication, yuan2021federated, deng2021local, sharma2022federated, sun2022communication}, where the clients are a relatively small number of organizations or data centers with reliable network connections~\citep{kairouz2019advances}.
Compared with the cross-silo setting, it is generally more challenging to solve a problem in the cross-device setting due to the low client availability and relatively slow network connections~\citep{kairouz2019advances}.
\citet{mohri2019agnostic} and~\citet{deng2020distributionally} investigate a special constrained problem, where the objective function is linear w.r.t.\ the dual variable, and propose two cross-device federated minimax algorithms.
However, these algorithms heavily rely on the linear structure, which makes them infeasible to general minimax problems including AUC maximization, robust adversarial learning, and GAN training.

In this paper, we develop the first practical federated algorithm named {\FrameworkName} ({\FrameworkAbbr}) that applies to general minimax problems in the cross-device setting.
Each training round of {\FrameworkAbbr} consists of two phases: (i) the gradient collection phase and (ii) the parameter update phase.
In the first phase, the server sends the global model parameters to a subset of clients, and aggregates local gradients computed at these clients to construct a global correction for subsequent local updates.
This correction mitigates the issue of \emph{client drift} (i.e., the phenomenon that local models on clients are shifted away from the global model~\cite{karimireddy2020scaffold}), and in turn mitigates the influence of slow network connections.
In the second phase, the server sends the global correction and model parameters to another subset of clients, which perform multiple local updates before sending the resulting local models back to the server.
Note that only a small subset of clients is required in each training round, which provides robustness against the low client availability in the cross-device setting.
Our major contributions are summarized as follows.
\begin{enumerate}[1.]
    \item We design a Start-Immediately-With-Enough-Responses (SIWER) mechanism for the server.
    That is, in each round, the server first sends a synchronization signal to a subset of clients and then starts constructing the global correction/updating the global model as long as it receives local results from enough  clients.
    With this mechanism, {\FrameworkAbbr} is tolerant of client failure, and thus significantly mitigates the impact of the low client availability of the cross-device FL setting.

    \item
    We construct a stochastic recursive momentum as the global correction term.
    As this correction term utilizes historical gradient information in a recursive manner, it only needs to aggregate information from a small portion of clients, and can greatly reduce the client drift without much additional communication cost.

    \item 
    We establish theoretical guarantees for {\FrameworkAbbr} under different settings of hyperparameters.
    In particular, we prove that, to reach a stationary point up to $\varepsilon$ accuracy, the communication cost of {\FrameworkAbbr} has an $\tilde{\OM}(1/\varepsilon^3)$\footnote{The symbol $\tilde{\OM}$ suppresses a logarithmic factor in $\varepsilon^{-1}$.} dependence on $\varepsilon$.
    We also prove that under appropriate assumptions, the communication cost of {\FrameworkAbbr} reduces as the number of local steps increases, which is the first such guarantee for federated minimax algorithms.
\end{enumerate}
Empirical studies on AUC maximization, robust adversarial neural network training, and GAN training tasks demonstrate the efficiency of the proposed algorithm.

\paragraph{Notation.} We use bold lowercase symbols (e.g., $\xB$) to denote vectors.
For a vector $\xB$, we denote its $j$-th coordinate as $[\xB]_j$.
For a function $f(\xB, \yB): \RBB^{p} \times \RBB^{q} \rightarrow \RBB$, $\nabla_{\xB} f$ and $\nabla_{\yB} f$ denote its partial derivatives with respect to the first and second variables, respectively.
For ease of notation, we also denote $\zB := (\xB^T, \yB^T)^T$, $f(\zB) := f(\xB, \yB)$, and $\nabla f(\xB, \yB) := (\nabla_{\xB} f(\xB, \yB)^T, \nabla_{\yB} f(\xB, \yB)^T)^T$.
The Euclidean norm of a vector $\xB$ is denoted by $\|\xB\|$.


\section{Related Works}  \label{section_related_work}
\paragraph{Single-machine minimax optimization algorithms.}
Minimax optimization has a long history dating back to~\cite{brown1951iterative}.
The majority of existing works on minimax problems focus on the convex-concave regime~\cite{korpelevich1976extragradient, nemirovski2004prox,nedic2009subgradient, liu2015projection, palaniappan2016stochastic, chavdarova2019reducing}.
Recently, there have emerged a surge of algorithms for more general nonconvex minimax problems\footnote{Nonconvex minimax problems include nonconvex-concave and nonconvex-nonconcave problems.}~\cite{sanjabi2018solving, nouiehed2019solving, lin2020gradient, luo2020stochastic, xu2020enhanced, qiu2020single, yang2020global}.

\paragraph{Federated minimax optimization algorithms.}
To tackle large-scale minimax problems, a few distributed minimax algorithms have been proposed in conventional centralized and decentralized settings~\cite{srivastava2011distributed, mateos2015distributed, liu2020decentralized, beznosikov2020distributed, rogozin2021decentralized}.
Recently, with the growing concern of privacy, distributed minimax optimization in the FL paradigm has received considerable interest.
Several algorithms based on the local update strategy are proposed to solve minimax problems in the cross-silo FL setting~\citep{reisizadeh2020robust, rasouli2020fedgan, beznosikov2020distributed, hou2021efficient, guo2020communication, yuan2021federated, deng2021local, sharma2022federated, sun2022communication}.
Though the algorithms in~\citep{reisizadeh2020robust, rasouli2020fedgan, beznosikov2020distributed, hou2021efficient, guo2020communication, yuan2021federated, deng2021local} perform multiple local update steps in each round to reduce the communication cost, they require \emph{full aggregation} (i.e., aggregation of local model information from all clients), and thus do not apply to the cross-device setting.
For a special constrained minimax problem of the form $\min_{\xB} \max_{\yB \in \RBB_+^N, \|\yB\|_1 = 1} \sum_{i=1}^{N} [\yB]_i f_i(\xB)$, where $N$ denotes the total number of clients and $f_i$ is the local loss function of client $i$,
\citet{mohri2019agnostic} and~\citet{deng2020distributionally} propose algorithms that in each round allow only a subset of clients to participate in training.
However, the algorithms in~\cite{mohri2019agnostic} and~\cite{deng2020distributionally} heavily depend on the linear structure of the above special constrained problem, and do not apply to general minimax problems. 

\section{Problem Setting and Algorithm}  \label{section_algorithms}

We consider the following general unconstrained minimax optimization problem in the cross-device FL setting:
\begin{equation}  \label{eq_problem}
    \min_{\xB \in \RBB^{p}} \max_{\yB \in \RBB^{q}} 
    \left\{f(\xB, \yB) := \EBB_{i \sim \DM} [f_i(\xB, \yB)] \right\},
\end{equation}
where 
$\DM$ represents the client distribution,
$
    f_i(\xB, \yB)
    := \frac{1}{n_i} \sum_{j = 1}^{n_i} F_i(\xB, \yB; \zeta_{i, j})
$ is the local loss function of client $i$,
and $\zeta_{i, 1}, \ldots, \zeta_{i, n_i}$ denote the local data points on client $i$.
Generally, $\DM$ can be any distribution over $N$ clients.
A typical example of $\DM$ is the uniform distribution, which corresponds to the loss function $f = \frac{1}{N} \sum_{i=1}^N f_i$.
Since the number of clients may be extremely large ($N \approx 10^{10}$) and the client availability is low in the cross-device setting, it is unlikely to access all clients at any one time during the learning procedure.
This precludes the usage of cross-silo FL algorithms that rely on the full aggregation strategy.

\begin{algorithm}[t]
    \caption{{\FrameworkAbbr} on the server.}
    \label{algorithm_alg_mb_server}
    \begin{algorithmic}[1]
    \SUB{Input:}
    the initial model parameters $(\xB_{-1}, \yB_{-1})$,
    the number of rounds $T$,
    and the numbers $\{S_t\}_{t=0}^{T-1}$ of received clients.

    \STATE $\xB_{0} \leftarrow \xB_{-1}$, $\yB_{0} \leftarrow \yB_{-1}$;

    \FOR{$t = 0, 1, \ldots, T - 1$}
        \STATE{\bfseries Gradient collection phase:}

        Send $\{\xB_t, \yB_t, \xB_{t-1}, \yB_{t-1}\}$ to a random subset $\hat{\SM}_t'$ of clients;

        \STATE Receive local gradient information from each client $i \in \SM_t' \subseteq \hat{\SM}_t'$, where $\SM_t'$ is the subset of $\hat{\SM}_t'$ containing the first $S_t$ clients that successfully respond;  \label{step_receive_local_gradient}

        Compute $\uB_t$ and $\vB_t$ by Eq.~\eqref{eq_storm};
        
        \STATE{\bfseries Parameter update phase:}
        \STATE Send $(\uB_t$, $\vB_t)$ and $(\xB_t, \yB_t)$ to another random subset $\hat{\SM}_t$ of clients;

        \STATE Receive local iterates $(\xB_{t, i}^{(K)}, \yB_{t, i}^{(K)})$ from each client $i \in \SM_t \subseteq \hat{\SM}_t$, where $\SM_t$ is the subset of $\hat{\SM}_t$ containing the first $S_t$ clients that successfully respond;

        \STATE 
        $
        \xB_{t + 1} \!\leftarrow\! \frac{1}{|\SM_t|} \sum_{i \in \SM_t} \xB_{t, i}^{(K)},
        \yB_{t + 1} \!\leftarrow\! \frac{1}{|\SM_t|} \sum_{i \in \SM_t} \yB_{t, i}^{(K)}
        $;
    \ENDFOR
\end{algorithmic}
\end{algorithm}

\begin{algorithm}[t]
    \caption{{\FrameworkAbbr} on client $i \in \hat{\SM}_t$ or $\hat{\SM}_t'$.}
    \label{algorithm_alg_mb_client}
    \begin{algorithmic}[1]

    \SUB{Input:}
    the number of local steps $K$, step sizes $\eta_t$ and $\gamma_t$, weight parameters $\alpha_t$ and $\beta$.

    \vspace{.5em}
    
    \textbf{Gradient collection phase (for $i \in \hat{\SM}_t'$):} 
    \STATE Receive $(\xB_{t}, \yB_{t})$ and $(\xB_{t-1}, \yB_{t-1})$ from the server.

    \STATE Compute $\nabla f_i (\xB_t, \yB_t) - (1 - \alpha_t) \nabla f_i (\xB_{t - 1}, \yB_{t - 1})$ and send it to the server; 

    \vspace{.5em}

    \textbf{Parameter update phase (for $i \in \hat{\SM}_t$):}
    \STATE Receive $(\xB_{t}, \yB_{t})$ and $(\uB_{t}, \vB_{t})$ from the server.

    \STATE Initialize local model $\xB_{t, i}^{(0)} \leftarrow \xB_t$, $\yB_{t, i}^{(0)} \leftarrow \yB_t$; \
    
    \FOR{$k = 0, 1, \ldots, K - 1$}
        \STATE Sample a minibatch $\BM_{t, i}^{(k)}$ from local data;  \label{line_local_iteration_start}

        \STATE Compute $\dB_{\xB, t, i}^{(k)}$ and $\dB_{\yB, t, i}^{(k)}$ by~\eqref{eq_local_update_direction};
    
        \STATE $\xB_{t, i}^{(k + 1)} \leftarrow
            \xB_{t, i}^{(k)} - \eta_t \dB_{\xB, t, i}^{(k)}$;
    
        \STATE $\yB_{t, i}^{(k + 1)} \leftarrow
            \yB_{t, i}^{(k)} + \gamma_t \dB_{\yB, t, i}^{(k)}$; \label{line_local_iteration_end}
    \ENDFOR
    \STATE Send $(\xB_{t, i}^{(K)}, \yB_{t, i}^{(K)})$ to the server; 
\end{algorithmic}
\end{algorithm}

To solve problem~\eqref{eq_problem}, we propose an algorithm named {\FrameworkName} ({\FrameworkAbbr}), which is detailed in Algorithms~\ref{algorithm_alg_mb_server}-\ref{algorithm_alg_mb_client}.
{\FrameworkAbbr} is based on a Start-Immediately-With-Enough-Responses (SIWER) mechanism.
In this mechanism, the server first sends a synchronization signal to a subset of clients, collects local gradients/models from the clients, and starts to aggregate the received information to construct the global correction/update the global model once a sufficient number of devices have reported results.
The SIWER mechanism is detailed in the following two phases.

\par\smallskip
\noindent\textbf{(i) Gradient collection phase.}
In this phase, the server aggregates local gradients to construct a stochastic recursive momentum as the correction direction, which is inspired by recent global correction techniques in the federated minimization literature~\cite{karimireddy2020scaffold, karimireddy2021breaking}.
Specifically, the server sends $(\xB_t, \yB_t)$ and $(\xB_{t-1}, \yB_{t-1})$ to a random subset of clients $\hat{\SM}_t'$, collects local gradients from this subset, and computes the correction direction.
Note that as some clients may be temporarily unavailable, the server only collects messages from the first $S_t$ responded clients and then proceed.
We denote these $S_t$ clients as $\SM_t'$.
To ensure that at least $S_t$ clients respond, the server can select a large enough $\hat{\SM}_t'$.
{\FrameworkAbbr} leverages the similarity of consecutive iterates to construct a lightweight estimator named the recursive momentum-based estimator, which is recursively constructed as
\begin{equation}  \label{eq_storm}
\resizebox{.9\linewidth}{!}{$
\begin{aligned}[b]
\begin{cases}
    \uB_{t} =& (1 - \alpha_t) \uB_{t-1}
        + \frac{1}{|\SM_t'|} \sum_{i \in \SM_t'} ( \nabla_{\xB} f_i(\xB_t, \yB_t) \\
            &- (1 - \alpha_t) \nabla_{\xB} f_i(\xB_{t-1}, \yB_{t-1}) ), \\
    \vB_t =& (1 - \alpha_t) \vB_{t-1}
        + \frac{1}{|\SM_t'|} \sum_{i \in \SM_t'} ( \nabla_{\yB} f_i(\xB_t, \yB_t) \\
            &- (1 - \alpha_t) \nabla_{\yB} f_i(\xB_{t-1}, \yB_{t-1}) ),
\end{cases}
\end{aligned}
$}
\end{equation}
for $t \!\ge\! 1$, where $\alpha_t \!\in\! (0, 1]$ is a hyperparameter.
For $t \!=\! 0$, $\uB_0$ (resp., $\vB_0$) is set to $\frac{1}{|\SM_0'|} \sum_{i \in \SM_0'} \! \nabla_{\xB} f_i(\xB_0, \yB_0)$ (resp., $\frac{1}{|\SM_0'|} \sum_{i \in \SM_0'} \! \nabla_{\yB} f_i(\xB_0, \yB_0)$).
Similar recursive momentum techniques have been extensively used in the optimization literature and generally lead to reduced variance with small batch sizes~\cite{cutkosky2019momentum, xie2020efficient, zhang2020one, qiu2020single, tran2021hybrid}.
Note that if $\alpha_t \equiv 1$, \eqref{eq_storm} degenerates to the minibatch estimator, i.e., 
$
    \uB_t = \frac{1}{|\SM_t'|} \sum_{i \in \SM_t'} \nabla_{\xB} f_i(\xB_t, \yB_t)
$
and
$
    \vB_t = \frac{1}{|\SM_t'|} \sum_{i \in \SM_t'} \nabla_{\yB} f_i(\xB_t, \yB_t)
$.
The minibatch estimator is the most widely-used and the simplest estimator in the optimization literature, which usually needs a large batch size to control the variance below a desired level~\cite{johnson2013accelerating}.

\par\smallskip
\noindent\textbf{(ii) Parameter update phase.}
In this phase, clients update local parameters using both local stochastic gradients and the global correction constructed in the first phase.
Specifically, the server first sends the current parameters $(\xB_t, \yB_t)$ and the global correction direction $(\uB_t, \vB_t)$ to a random subset of clients $\hat{\SM}_t$, which is potentially different from $\hat{\SM}_t'$ in the previous phase.
Then, each active client in $\hat{\SM}_t$ performs $K$ local iterations and sends its local parameters to the server.
Again, the server only collects messages from the first $S_t$ responded clients, which we denote as $\SM_t$, as some clients may be temporarily unavailable.
Finally, the server averages the local parameters from $\SM_t$ to produce the next global parameters.
In each local iteration (line~\ref{line_local_iteration_start} to~\ref{line_local_iteration_end} in Algorithm~\ref{algorithm_alg_mb_client}), {\FrameworkAbbr} simultaneously takes a descent step on the local primal variable and an ascent step on the dual one.
The primal and dual update directions are computed as
\begin{equation}  \label{eq_local_update_direction}
\resizebox{.99\linewidth}{!}{$
\begin{aligned}
\begin{cases}
    \dB_{\xB, t, i}^{(k)}
    \!=\! \nabla_{\xB} F_i(\xB_{t, i}^{(k)}, \yB_{t, i}^{(k)}; \BM_{t, i}^{(k)})
        \!+\! \beta (\uB_t \!-\! \nabla_{\xB} F_i(\xB_t, \yB_t; \BM_{t, i}^{(k)})) \\
    \dB_{\yB, t, i}^{(k)}
    \!=\! \nabla_{\yB} F_i(\xB_{t, i}^{(k)}, \yB_{t, i}^{(k)}; \BM_{t, i}^{(k)})
       \!+\! \beta (\vB_t \!-\! \nabla_{\yB} F_i(\xB_t, \yB_t; \BM_{t, i}^{(k)})),
\end{cases}
\end{aligned}
$}
\end{equation}
where $\beta \in \{0, 1\}$ is a hyperparameter.

According to the choices of the weight parameters $\beta$ and $\alpha_t$, there are three specific versions of {\FrameworkAbbr}.
\begin{enumerate}[1.]
    \item \emph{{\AlgNaiveFull} ({\AlgNaive})}.
    If $\beta = 0$, the local update directions in~\eqref{eq_local_update_direction} are simply local minibatch gradients.
    Hence, the gradient collection phase is not needed, and {\FrameworkAbbr} degenerates to a single-phase algorithm, which we refer to as {\AlgNaiveFull} ({\AlgNaive}).

    \item \emph{{\AlgMBFull} ({\AlgMB})}.
    When $\beta = 1$ and $\alpha_t \equiv 1$, the local update directions incorporate global correction, which is computed using the degenerate minibatch estimator.
    We denote {\AlgMBFull} as {\AlgMB}.

    \item \emph{{\AlgSTORMFull} ({\AlgSTORM})}.
    In the case $\beta = 1$ and $\alpha_t \not\equiv 1$, the correction direction is computed using the recursive momentum-based estimator.
    We refer to this version of {\FrameworkAbbr} as {\AlgSTORM}.
\end{enumerate}
We highlight that since only a random subset of clients is involved in each round, {\FrameworkAbbr} is resilient to the low client availability.
Further, when $\beta = 1$, the local update directions in~\eqref{eq_local_update_direction} approximate global gradients $\nabla_{\xB} f(\xB_{t, i}^{(k)}, \yB_{t, i}^{(k)})$ and $\nabla_{\yB} f(\xB_{t, i}^{(k)}, \yB_{t, i}^{(k)})$ if $(\xB_{t, i}^{(k)}, \yB_{t, i}^{(k)})$ is close to $(\xB_t, \yB_t)$.
This mitigates the issue of client drift.
As we shall see in the next section, incorporating the global correction in local update steps also reduces the communication complexity.

We note that the concurrent work~\cite{sun2022communication} uses a similar global correction for cross-silo federated minimax learning. However, the computation of their global correction requires the full aggregation and thus do not apply to the cross-device setting.
Besides, the analysis in~\cite{sun2022communication} focuses on strongly-convex-strongly-concave minimax problems, while we consider a class of nonconvex-nonconcave problems, which are more challenging.


\section{Convergence Analysis}  \label{section_analysis}
This section establishes convergence guarantees for {\FrameworkAbbr}.
Here, we only present the major results and defer detailed analyses to Appendix~\ref{section_deferred_analysis}.
The following common assumptions are needed throughout our analyses.

\begin{assumption}[Bounded gradient dissimilarity]  \label{assumption_bgd}
There exist positive constants $\sigma_1$ and $\sigma_2$ such that $\forall \xB \in \RBB^{p}, \yB \in \RBB^{q}$,
$$
\begin{aligned}[b]
\begin{cases}
    \EBB_{i \sim \DM} [\|\nabla_{\xB} f_i(\xB, \yB) - \nabla_{\xB} f(\xB, \yB) \|^2] \le \sigma_1^2, \\
    \EBB_{i \sim \DM} [\|\nabla_{\yB} f_i(\xB, \yB) - \nabla_{\yB} f(\xB, \yB) \|^2] \le \sigma_2^2.
\end{cases}
\end{aligned}
$$
\end{assumption}

\begin{assumption}[Lipschitz continuous gradients]  \label{assumption_smooth}
There exists a positive constant $L_f > 0$ such that for any $i$ and $\zeta \in \{\zeta_{i, 1}, \ldots, \zeta_{i, n_i}\}$, the function $F_i(\cdot, \cdot; \zeta)$ has $L_f$-Lipschitz continuous gradients, i.e., $\forall \xB_1, \xB_2 \in \RBB^{p}$ and $\yB_1, \yB_2 \in \RBB^{q}$,
$$
\resizebox{\linewidth}{!}{$
    \left\|
        \begin{bmatrix}
            \nabla_{\xB} F_i(\xB_1, \yB_1; \zeta) - \nabla_{\xB} F_i(\xB_2, \yB_2; \zeta) \\
            \nabla_{\yB} F_i(\xB_1, \yB_1; \zeta) - \nabla_{\yB} F_i(\xB_2, \yB_2; \zeta)
        \end{bmatrix}
    \right\|
    \le L_f \left\|
        \begin{bmatrix}
            \xB_1 - \xB_2 \\
            \yB_1 - \yB_2
        \end{bmatrix} \right\|.
$}
$$
\end{assumption}

\begin{assumption}[Polyak-{\L}ojasiewicz (P{\L}) condition]  \label{assumption_pl}
There exists a constant $\mu > 0$ such that $\forall \xB \in \RBB^{p}, \yB \in \RBB^{q}$,
$$
    \| \nabla_{\yB} f(\xB, \yB) \|^2
    \ge 2 \mu \big(\mathrm{max}_{\yB' \in \RBB^{q}} f(\xB, \yB') - f(\xB, \yB) \big).
$$
\end{assumption}

We also assume that each client that successfully responds to the server in round $t$ follows the underlying client distribution $\DM$, which is a common assumption in the federated learning literature~\cite{li2018federated, li2019convergence, karimireddy2021breaking, acar2021federated}.
In our analyses, we denote $S := \min \{S_0, \ldots, S_{T-1}\}$.

A natural metric for measuring the performance of an algorithm on problem~\eqref{eq_problem} is the gradient norm of the function $\Phi(\xB) := \max_{\yB \in \RBB^q} f(\xB, \yB)$.
This metric is commonly used in analyzing algorithms for nonconvex-P{\L} or nonconvex-strongly-concave minimax problems~\cite{reisizadeh2020robust, deng2021local, lin2020gradient, luo2020stochastic}.
Note that $\Phi(\xB)$ has $L_{\Phi}$-Lipschitz continuous gradients according to~\cite{nouiehed2019solving}, where $L_{\Phi} := (1 + \kappa / 2) L_f$.
For some $\varepsilon > 0$, a point $\xB \in \RBB^p$ is said to be $\varepsilon$-stationary if $\| \nabla \Phi(\xB) \| \le \varepsilon$.
Once an approximate primal solution $\tilde{\xB}$ is obtained, one can easily find an approximate dual solution by solving $\max_{\yB} f(\tilde{\xB}, \yB)$.

\subsection{Analysis of {\AlgNaive}}  \label{section_analysis_naive}
We first provide the theoretical analysis for {\AlgNaive}.
Besides Assumptions~\ref{assumption_bgd}-\ref{assumption_pl}, the bounded local variance assumption below is also needed.
\begin{assumption}[Bounded local variance]  \label{assumption_intra_client_variance}
For any $(\xB, \yB) \in \RBB^{p \times q}$, and any client $i$,
$$
\begin{cases}
\frac{1}{n_i} \sum_{j=1}^{n_i} \| \nabla_{\xB} F_i(\xB, \yB; \zeta_{i, j}) - \nabla_{\xB} f_i(\xB, \yB) \|^2 \le G_1^2, \\
\frac{1}{n_i} \sum_{j=1}^{n_i} \| \nabla_{\yB} f_i(\xB, \yB; \zeta_{i, j}) - \nabla_{\yB} f_i(\xB, \yB) \|^2 \le G_2^2.
\end{cases}
$$
\end{assumption}

We summarize the convergence result of {\AlgNaive} in Theorem~\ref{theorem_alg_naive} below and defer the detailed analysis to Appendix~\ref{section_analysis_naive_appendix}.
\begin{theorem}  \label{theorem_alg_naive}
Suppose that Assumptions~\ref{assumption_bgd}, \ref{assumption_smooth}, and~\ref{assumption_intra_client_variance} hold.
Define the step sizes of {\AlgNaive} as
$\gamma_t = \min \Big\{
    \sqrt{\frac{20 \LM_0 S}{L_f T K^2 \sigma_2^2}},
    \Big( \frac{30 \LM_0}{L_f^2 (\sigma_2^2 + G_2^2) T K^3} \Big)^{1/3},
    \allowbreak
    \frac{1}{87 L_f K}
\Big\}
$
and
$
\eta_t = \min \Big\{
    \sqrt{\frac{20 \LM_0 S}{7 (L_{\Phi} + L_f) T K^2 \sigma_1^2}},
    \Big( \frac{3 \LM_0}{L_f^2 (\sigma_1^2 + G_1^2) T K^3} \Big)^{1/3},
    \frac{\gamma_t}{21 \kappa^2}
\Big\}
$.
If the minibatches $\BM_{t, i}^{(0)}, \ldots, \BM_{t, i}^{(K-1)}$ in {\AlgNaive} are drawn in a random reshuffling manner and $K$ is an integral multiple of the epoch length, then
\begin{align*}
    \frac{1}{T} \sum_{t=0}^{T-1} \EBB[\|\nabla \Phi(\xB_t)\|^2]
    =& \OM(\frac{\kappa^2}{T^{2/3}} + \frac{\kappa^2}{\sqrt{S T}}),
\end{align*}
where $\OM$ hides the dependence on $\LM_0$, $L_f$, $\sigma_1$, and $\sigma_2$.
\end{theorem}
Theorem~\ref{theorem_alg_naive} indicates that {\AlgNaive} finds an $\varepsilon$-stationary point in $\OM(\frac{\kappa^3}{\varepsilon^3} + \frac{\kappa^4}{S \varepsilon^4})$ communication rounds.
Actually, when all clients participate in each round, CDMA-NC reduces to the local SGDA algorithm in~\cite{sharma2022federated}, and the second term of our communication cost can be canceled by slightly modifying our analysis, which recovers the $\OM(\kappa^3 / \varepsilon^3)$ communication cost in~\cite{sharma2022federated}.
As our paper focuses on the cross-device setting where the total number of clients is extremely large, it is unrealistic to require all clients to participate in each round, and Theorem~\ref{theorem_alg_naive} shows that as long as $S = \OM(\kappa / \varepsilon)$, CDMA-NC achieves the same $\OM(\kappa^3 / \varepsilon^3)$ communication cost as local SGDA.

\subsection{Analysis of {\AlgMB}}  \label{section_analysis_mb}
The analysis of {\AlgMB} is similar to that of {\AlgNaive}.
Compared with {\AlgNaive}, {\AlgMB} incorporates global correction into local updates.
Consequently, {\AlgMB} has a smaller client drift compared with that in {\AlgNaive}, which in turn leads to a better convergence rate.
The theorem below establishes the convergence rate of {\AlgMB}.
\begin{theorem}  \label{theorem_alg_mb}
Define the step sizes $\eta_t$ and $\gamma_t$ of {\AlgMB} as
$
\eta_t = \min \left\{
    \frac{\gamma_t}{21 \kappa^2},
    \sqrt{\frac{40 \LM_0 S}{21 (L_{\Phi} + L_f) T K^2 \sigma_1^2}}\right\}
$ and
$\gamma_t = \min \left\{ \frac{1}{87 L_f K}, \sqrt{\frac{4 \LM_0 S}{3 L_f T K^2 \sigma_2^2}} \right\},
$
where $\LM_0 := \Phi(\xB_0) - \Phi(\xB^*) + \frac{1}{20} (\Phi(\xB_0) - f(\zB_0))$.
If Assumptions~\ref{assumption_bgd}-\ref{assumption_pl} hold, then
$$
\begin{aligned}
    &\frac{1}{T} \sum_{t=0}^{T-1} \EBB[\|\nabla \Phi(\xB_t)\|^2]
    = \OM(\frac{\kappa^2}{T} + \frac{\kappa^2}{\sqrt{S T}}),
\end{aligned}
$$
where $\OM$ hides the dependence on $\LM_0$, $L_f$, $\sigma_1$, and $\sigma_2$.
\end{theorem}

By Theorem~\ref{theorem_alg_mb}, {\AlgMB} has a communication cost of $\OM(\kappa^2 / \varepsilon^2 + \kappa^4 / (S \varepsilon^4))$, better than the $\OM(\kappa^3 / \varepsilon^3 + \kappa^4 / (S \varepsilon^4))$ cost of {\AlgNaive}.
This demonstrates that incorporating global gradient estimates into local updates is more efficient than merely using local stochastic gradients.

\begin{remark}
    When $K = 1$, {\AlgMB} coincides with Parallel SGDA, which directly parallelizes the single-machine SGDA algorithm~\citep{lin2020gradient} over the clients.
    Under the stronger assumption that $f(\xB, \cdot)$ is $\mu$-strongly concave given any $\xB$, \citet{lin2020gradient} proves that Parallel SGDA has an $\OM(\kappa^2 / \varepsilon^2 + \kappa^3 / (S \varepsilon^5))$ communication complexity\footnote{
        Note that~\citet{lin2020gradient} also shows another complexity of $\OM(\kappa^2 / \varepsilon^2)$ if $S \!=\! \OM(\kappa / \varepsilon^2)$.
        However, when $\varepsilon$ is small, such setting of $S$ is unsuitable for cross-device FL where only a fraction of clients are available at one time.
    }.
    This complexity is inferior to that of {\AlgMB} in terms of the dependence on $\varepsilon$.

\end{remark}

\subsection{Analysis of {\AlgSTORM}}  \label{section_analysis_storm}
As mentioned in the previous section,
{\AlgSTORM} utilizes the variance-reduced estimator~\eqref{eq_storm} which incorporates historical gradient information. 
This results in a more accurate global correction direction compared to the degenerate one used in {\AlgMB} and accelerates the convergence.
The convergence rate of {\AlgSTORM} is presented in the next theorem.
\begin{theorem}  \label{theorem_alg_storm_no_delta}
Define $\alpha_t$, $\eta_t$, and $\gamma_t$ in {\AlgSTORM} as
$\alpha_t = 200000 L_f^2 K^2 \gamma_t^2$,
$\eta_t = \frac{\gamma_t}{12 \kappa^2}$,
and $\gamma_t = \min\{\frac{1}{2 L_f}, \frac{\num{5.78e-4}}{L_f K}, \frac{S^{1/3} \hat{\LM}_0^{1/3}}{L_f^{2/3} K (t + 1)^{1/3} (\sigma_1^2 / \kappa^2 + \sigma_2^2)^{1/3}}\}$,
where $\hat{\LM}_0 \allowbreak := \Phi(\xB_0) - \Phi(\xB^*) + \frac{1}{7} \left( \Phi(\xB_0) - f(\zB_0) \right)$.
Suppose that Assumptions~\ref{assumption_bgd}-\ref{assumption_pl} hold and $S \ge (\sigma_1^2 / \kappa^2 + \sigma_2^2) / (L_f \hat{\LM}_0)$.
Then,
\begin{align*}  
&{\textstyle \frac{1}{K T} \sum_{t=0}^{T-1} {\textstyle \frac{1}{|\SM_t|}} \sum_{k=0}^{K-1} \sum_{i \in \SM_t}} \EBB[\|\nabla \Phi(\xB_{t, i}^{(k)})\|^2] \nonumber\\
    ={}& \tilde{\OM}(\kappa^2 / T + \kappa^2 (1 + 1/K) / (S^{1/3} T^{2/3})),
\end{align*}
where $\tilde{\OM}$ hides logarithmic factors and the dependence on $\LM_0$, $L_f$, $\sigma_1$, and $\sigma_2$.
\end{theorem}

Theorem~\ref{theorem_alg_storm_no_delta} indicates that the communication cost of {\AlgSTORM} to reach an $\epsilon$-stationary point is
\begin{equation}  \label{eq_alg_storm_complexity_no_delta}
    \tilde{\OM}(\kappa^2 / \varepsilon^2 + \kappa^3 (1 + 1/K)^{3/2} / (S^{1/2} \varepsilon^3)),
\end{equation}
which outperforms those of {\AlgNaive} and {\AlgMB} obtained in previous subsections.
Actually, the communication complexity bound of {\AlgSTORM} can be further improved in terms of the dependence on the condition number $\kappa$ under the following $\delta$-bounded Hessian dissimilarity ($\delta$-BHD) assumption~\cite{shamir2014communication, arjevani2015communication, reddi2016aide}.
\begin{assumption}[$\delta$-bounded Hessian dissimilarity]  \label{assumption_bhd}
    There exists a constant $\delta > 0$ such that for any client $i$ and any $\zeta \!\in\! \{\zeta_{i, 1}, \ldots, \zeta_{i, n_i}\}$, $\|\nabla^2 F_i(\xB, \yB; \zeta) \!-\! \nabla^2 f(\xB, \yB)\| \le 2 \delta$.
\end{assumption}
This assumption is widely used in the optimization literature to characterize the second-order similarity of objectives, which usually results in an improved convergence rate~\cite{shamir2014communication, arjevani2015communication, reddi2016aide, meng2020fast} compared with those rates obtained merely under first-order conditions.
Note that under Assumption~\ref{assumption_smooth}, Assumption~\ref{assumption_bhd} immediately holds with $\delta = L_f$.
Actually, $\delta$ reduces as the data heterogeneity on different clients decreases and $\delta = 0$ when the data on different clients follow the same distribution.
With this additional $\delta$-BHD assumption, we prove that {\AlgSTORM} has the following modified communication cost (see Appendix~\ref{section_analysis_storm_appendix} for details)
\begin{equation}  \label{eq_alg_storm_complexity}
    {\textstyle \tilde{\OM} (\frac{\kappa }{\varepsilon^2} (\frac{\kappa}{K} + c) + (1 + \frac{\kappa}{c K})^{3} \frac{c \kappa^2}{S^{1/2} \varepsilon^3}).}
\end{equation}
The complexity bound~\eqref{eq_alg_storm_complexity} decreases as the number of local steps $K$ increases.
If we set $K = 1$, the bound in~\eqref{eq_alg_storm_complexity} reads $\tilde{\OM}(\frac{ \kappa^2}{\varepsilon^2} + \frac{ \kappa^5}{S^{1/2} c^2\varepsilon^{3}})$, which is inferior to the bound in \eqref{eq_alg_storm_complexity_no_delta}.
When $K \ge \kappa / c$,
the complexity becomes $\tilde{\OM}(\frac{c \kappa}{\varepsilon^2} + \frac{c \kappa^2}{S^{1/2} \varepsilon^{3}})$ and outperforms that in \eqref{eq_alg_storm_complexity_no_delta} as $c := \max\{1, \delta / \mu\} \le \kappa$.
By contrast, in Theorem~\ref{theorem_alg_storm_no_delta} where the $\delta$-BHD assumption is absent (i.e., $\delta = L_f$), the order of the communication complexity of {\AlgSTORM} remains the same, though it also decreases as $K$ increases. 

Note that the complexities of {\AlgNaive} and {\AlgMB} obtained in Theorems~\ref{theorem_alg_naive} and~\ref{theorem_alg_mb} are independent of $K$.
Nevertheless, our experimental results show that multiple local steps result in better performance for these two algorithms (check the next section).
Actually, similar gaps between the theoretical analyses and the practical performance on the effect of local steps widely exist in the FL literature~\cite{khaled2020tighter, karimireddy2020scaffold, beznosikov2020distributed, karimireddy2021breaking}.
Recently, several studies on lower bounds of communication complexities for distributed optimization under different assumptions show that local steps do not provide speed-up in some settings~\cite{woodworth2020minibatch, beznosikov2020distributed}.
The investigation on the communication lower bound and whether local steps improve the communication complexities of {\AlgNaive} and {\AlgMB} under other assumptions is left for future work.

We compare the communication complexities obtained in this section with that of Parallel SGDA in Table~\ref{table_comparison}.

\begingroup
\renewcommand{\arraystretch}{0.1} 

\begin{table}[t]
    \begin{threeparttable}
    \begin{tabular}{@{}p{.45\textwidth}@{}}
    \centering
    \begin{tabular}{ll}
        \toprule
        Algorithm                      & Communication complexity \\
        \midrule
        Parallel SGDA    & $\OM(\frac{\kappa^2}{\varepsilon^2} + \frac{\kappa^3}{S \varepsilon^5})$$^{\dagger}$   \\
        \midrule
        \textbf{{\AlgNaive}}     & $\OM(\frac{\kappa^3}{\varepsilon^3} + \frac{\kappa^4}{S \varepsilon^4})$ \\
        \midrule
        \textbf{{\AlgMB}}   & $\OM(\frac{\kappa^2}{\varepsilon^2} + \frac{\kappa^4}{S \varepsilon^4})$   \\
        \midrule 
        \multirow{2}{*}{\textbf{{\AlgSTORM}}}  & $\tilde{\OM} (\frac{\kappa^2}{\varepsilon^2} + (1 + \frac{1}{K})^{3/2} \frac{\kappa^3}{S^{1/2} \varepsilon^3} )$  \\
        \cmidrule{2-2}
        & $\tilde{\OM} (\frac{\kappa }{\varepsilon^2} (\frac{\kappa}{K} + c) + (1 + \frac{\kappa}{c K})^{3} \frac{c \kappa^2}{S^{1/2} \varepsilon^3} )$$^{\sharp}$  \\
        \bottomrule
    \end{tabular}
    \end{tabular}
    \begin{tablenotes}
        \footnotesize
        \item $^{\dagger}$ This entry relies on the extra condition that $f(\xB, \cdot)$ is $\mu$-strongly concave $\forall \xB$, which is stronger than Assumption~\ref{assumption_pl}.
        \item $^{\sharp}$ This result requires the additional Assumption~\ref{assumption_bhd}. Note that $c := \max\{1, \delta / \mu\} \le \kappa$.
    \end{tablenotes}
    \caption{Communication costs of algorithms for minimax problems in cross-device FL.}
    \label{table_comparison}
    \end{threeparttable}
\end{table}

\endgroup


\section{Experiments}  \label{section_experiments}
To demonstrate the efficiency of the proposed algorithm, we conduct experiments on three tasks: (i) AUC maximization, (ii) robust adversarial neural network training, and (iii) GAN training.
We compare the proposed three versions of {\FrameworkAbbr} with
Parallel SGDA~\cite{lin2020gradient},
and the cross-device versions of
Extra Step Local SGD~\cite{beznosikov2020distributed},
Local SGDA+~\cite{deng2021local},
Catalyst-Scaffold-S~\cite{hou2021efficient},
CODA+,
and CODASCA~\cite{yuan2021federated}.
We note that Local SGDA+, Extra Step Local SGD, Catalyst-Scaffold-S, CODA+, and CODASCA are state-of-the-art federated minimax algorithms which rely on the full aggregation strategy and thus only apply to the cross-silo setting.
Here, we use their cross-device variants by only involving a subset of clients in each round, which are actually \emph{not} theoretically guaranteed.

In our experimental setting, there are $500$ clients in total and the client distribution $\DM$ is the uniform distribution.
Since communication is often the major bottleneck in cross-device FL and clients generally have much slower upload than download bandwidth~\cite{kairouz2019advances}, we compare the algorithms given the same amount of data transferred from clients to the server.
To ensure that each algorithm has the same amount of communication per round, for {\AlgNaive}, Parallel SGDA, Local SGDA+, Extra Step Local SGD and CODA+, which only require to send parameters from clients to the server, the server selects a random client subset of size $\hat{S} = 16$ in each round.
For the rest algorithms, in which clients send both parameters and gradient estimates to the server, the size of the randomly selected client subset is fixed to $\hat{S} = 8$ in each round.
Moreover, to simulate the low client availability, we set the size of clients that successfully respond to the server as $S_t = \ceil{p_t \hat{S}}$, where the response probability $p_t$ is uniformly sampled from $[0.5, 1]$.
Throughout the experiments, we use the same random seed for all algorithms for reproducibility.
We do grid-search on hyperparameters for the algorithms and select the best hyperparameters.
All experiments were implemented in Pytorch and run on $4$ workstations, each with $2$ Intel E5-2680 v4 CPUs ($28$ cores), $4$ NVIDIA 2080Ti GPUs, and $378$GB memory.\footnote{Source code: \url{https://github.com/xjiajiahao/federated-minimax}}

\subsection{AUC Maximization}  \label{section_experiment_auc}
In the first experiment, we consider the $\ell_2$-relaxed AUC maximization problem in the minimax formulation~\cite{ying2016stochastic, liu2020stochastic}.
This formulation is widely used in binary classification tasks, especially on imbalanced data.
We use two datasets: MNIST~\cite{lecun1998gradient} and CIFAR-10~\cite{krizhevsky2009learning}.
To make the data heterogeneous, the training data is first sorted according to the original class label and then equally partitioned into $500$ clients so that all data points on one client are from the same class.
Each algorithm runs exactly one local epoch ($K = 12$ local steps for MNIST and $K = 10$ for CIFAR-10) with the minibatch size $B = 10$ on the selected clients in each round.

\begin{figure}[t]
    \captionsetup[subfloat]{farskip=1pt,captionskip=1pt}
    \centering
    \subfloat{\includegraphics[trim={0cm 0cm 0cm 0cm}, clip, width=.49\linewidth, height=.44\linewidth]{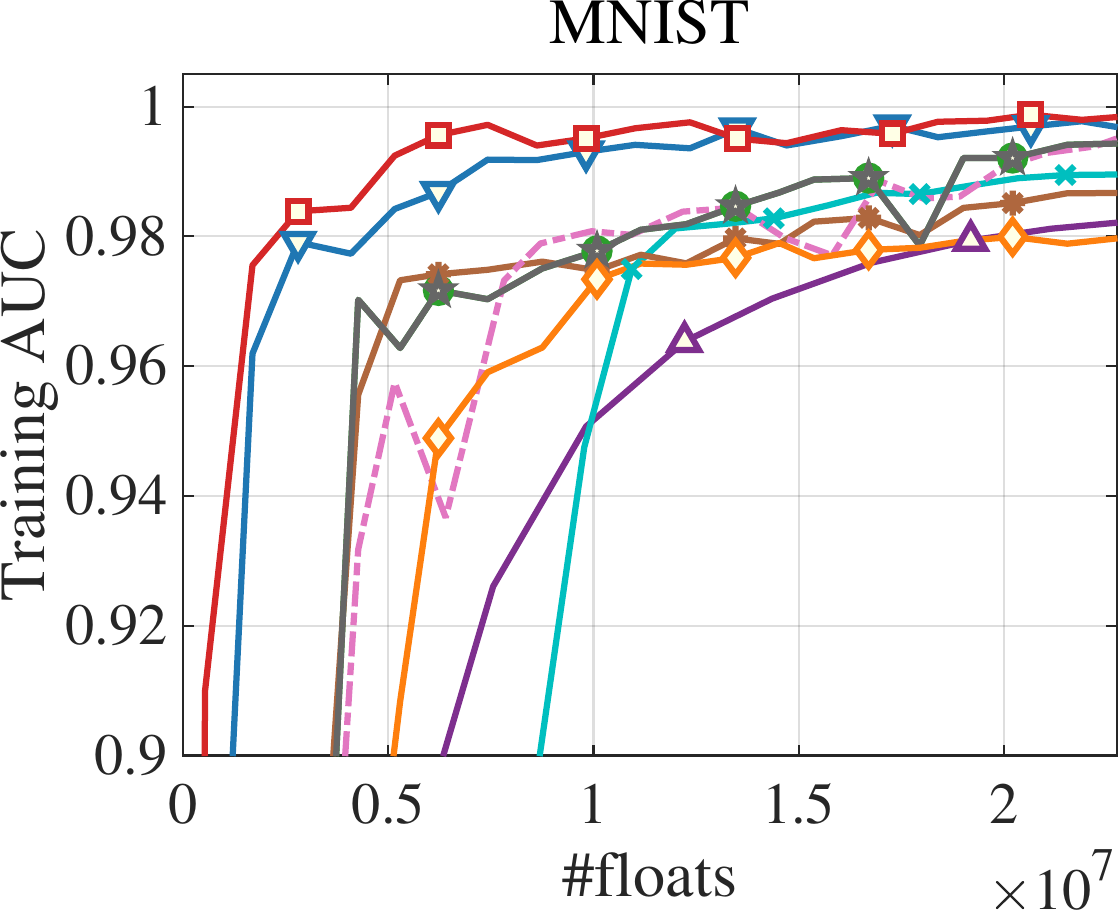}}
    \hfil
    \subfloat{\includegraphics[trim={0cm 0cm 0cm 0cm}, clip, width=.49\linewidth, height=.44\linewidth]{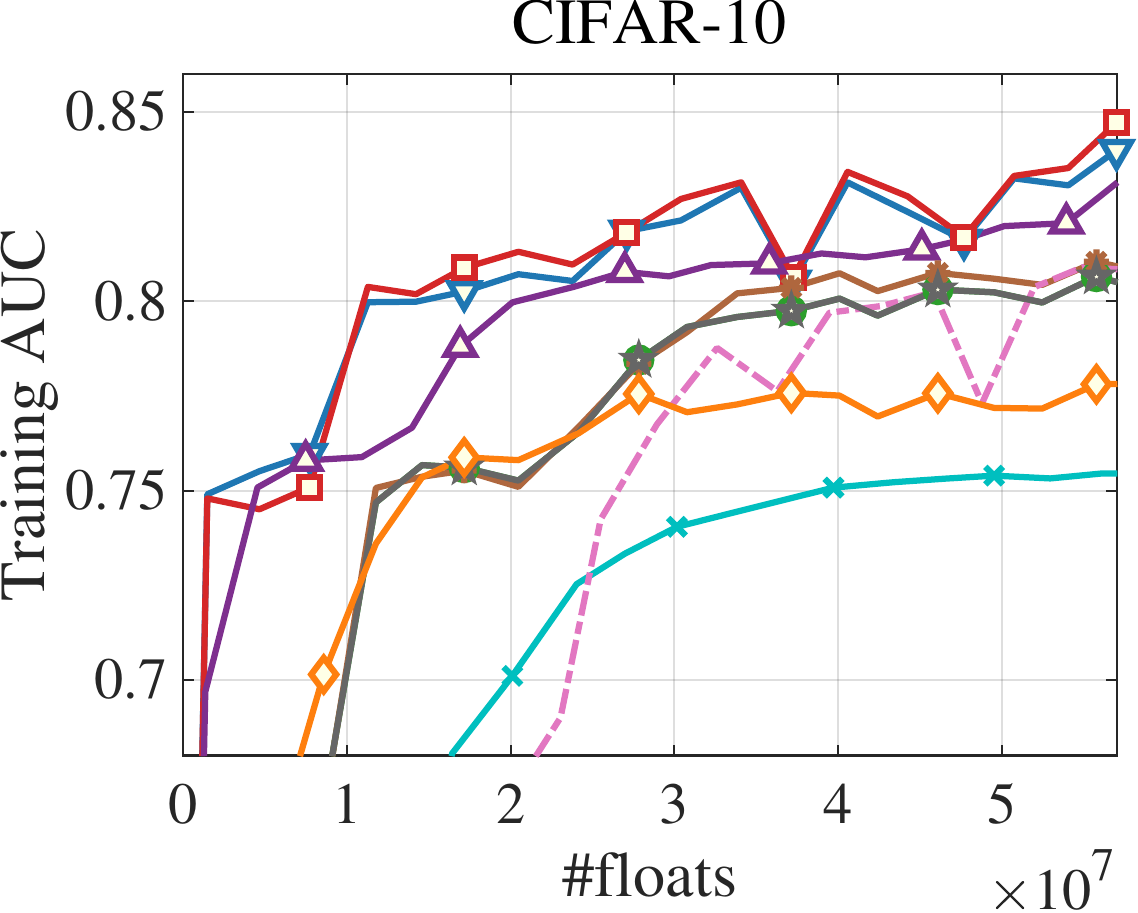}}
    \\[0.25em]
    \subfloat{\includegraphics[trim={0cm 0cm 0cm 0cm}, clip, width=.49\linewidth, height=.44\linewidth]{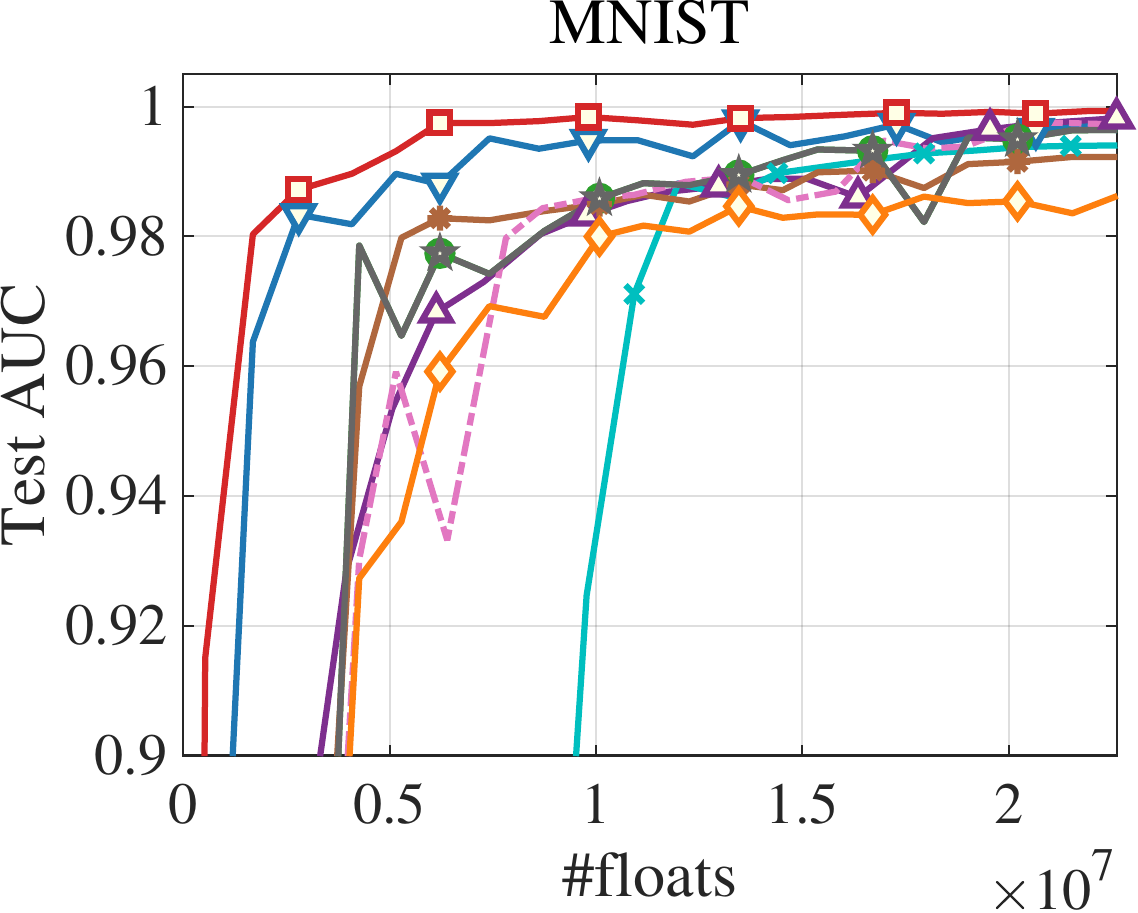}}
    \hfil
    \subfloat{\includegraphics[trim={0cm 0cm 0cm 0cm}, clip, width=.49\linewidth, height=.44\linewidth]{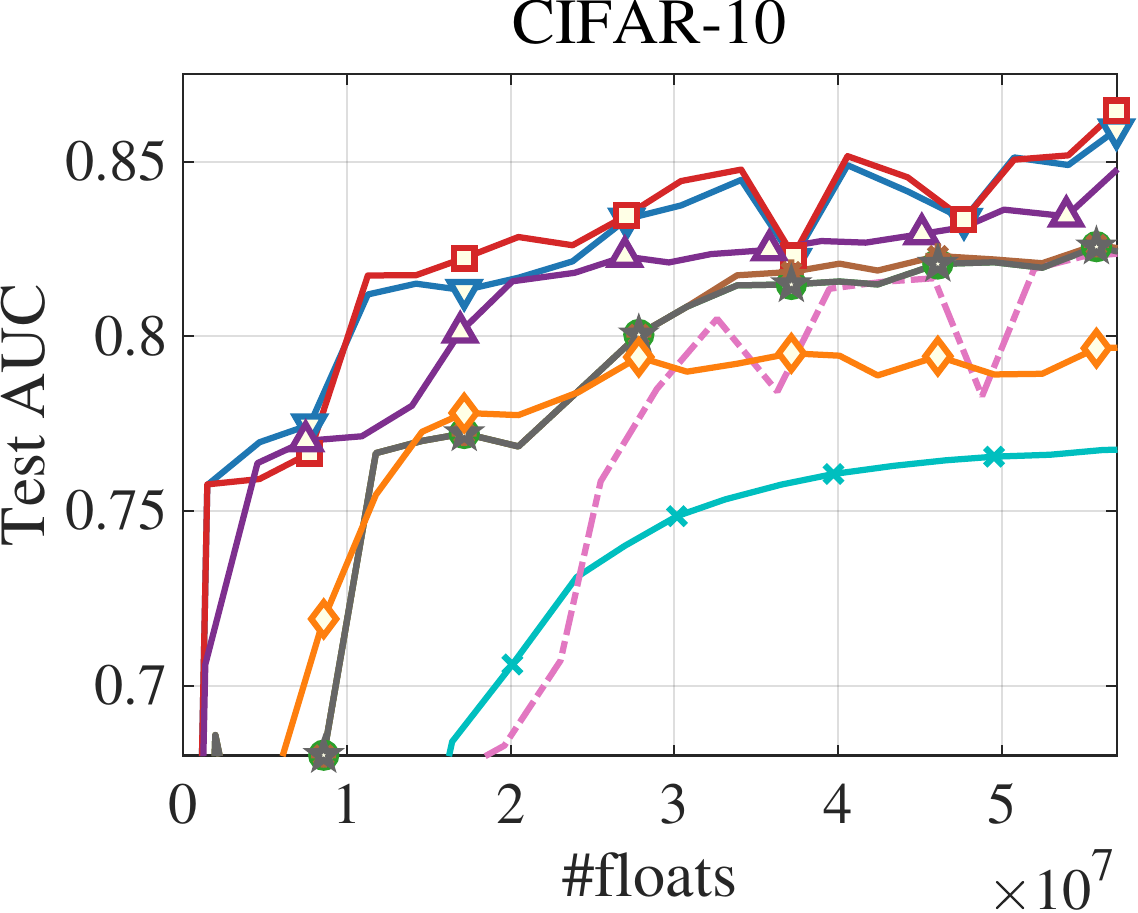}}
    \\[0.2em]
    \subfloat{\includegraphics[trim={-1.9cm 0cm 0cm 0cm}, clip, width=.96\linewidth]{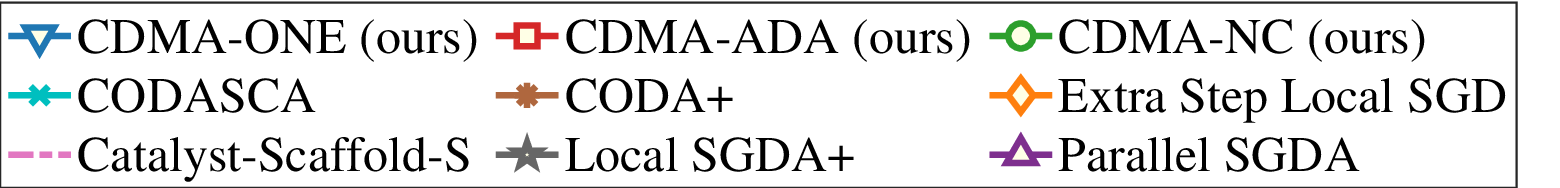}}
    \caption{
        Results on the AUC maximization task (top: training results, bottom: test results).
        The left and right columns correspond to MNIST and CIFAR-10 datasets, respectively.
    }
    \label{figure_auc}
\end{figure}

We report the AUC value on the training and test datasets versus the amount of communication (the total number of FLOATs received by the server) in Figure~\ref{figure_auc}.
We observe that {\AlgSTORM} and {\AlgMB} are superior to other algorithms on both datasets.
In addition, among the theoretically-guaranteed algorithms, {\AlgSTORM} achieves the best performance, {\AlgMB} is only inferior to {\AlgSTORM}, and {\AlgNaive} and Parallel SGDA have comparable performance.
This corroborates our theory and demonstrates the efficiency of the SWIER aggregation mechanism as well as the stocahstic recursive momentum.

\begin{figure}[t]
    \captionsetup[subfloat]{farskip=1pt,captionskip=1pt}
    \centering
    \subfloat{\includegraphics[trim={0cm 0cm 0cm 0cm}, clip, width=.48\linewidth, height=.44\linewidth]{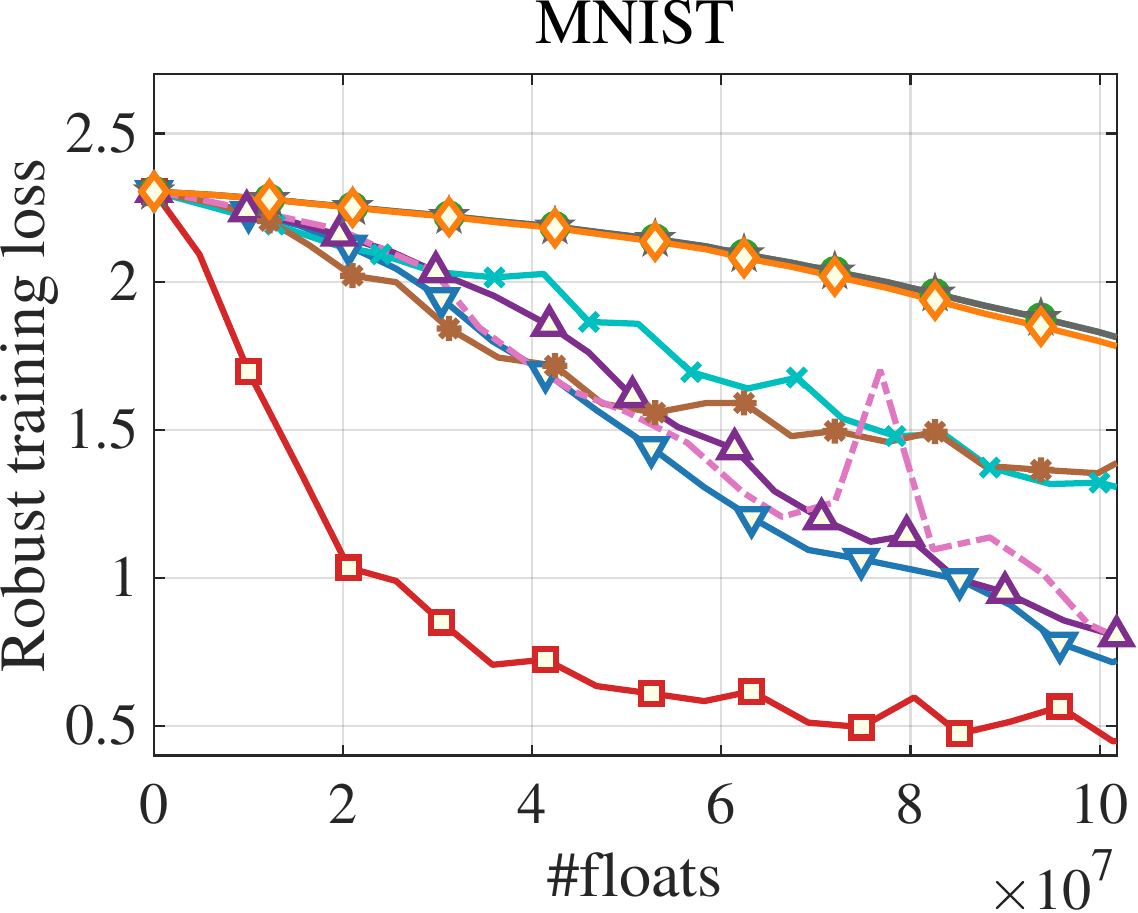}}
    \hfil
    \subfloat{\includegraphics[trim={0cm 0cm 0cm 0cm}, clip, width=.48\linewidth, height=.44\linewidth]{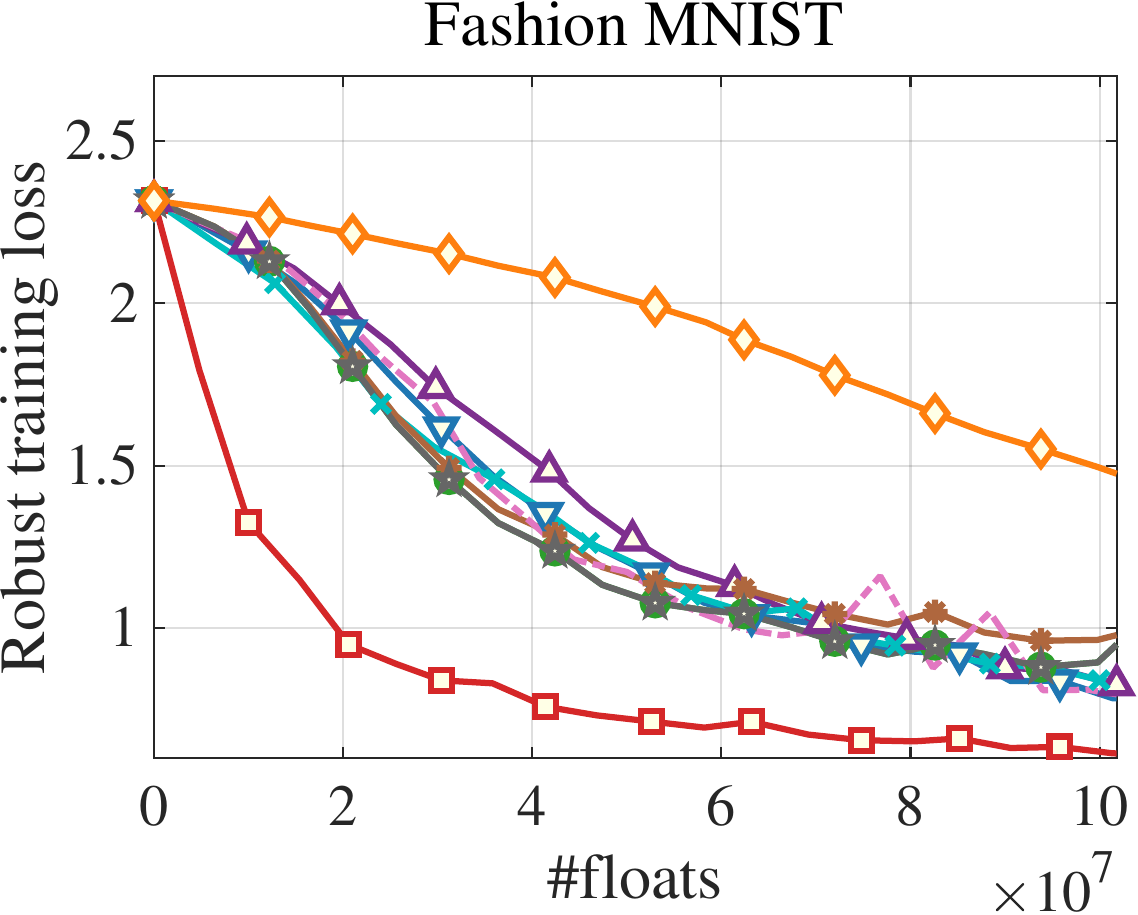}}
    \\[0.25em]
    \subfloat{\includegraphics[trim={0cm 0cm 0cm 0cm}, clip, width=.48\linewidth, height=.44\linewidth]{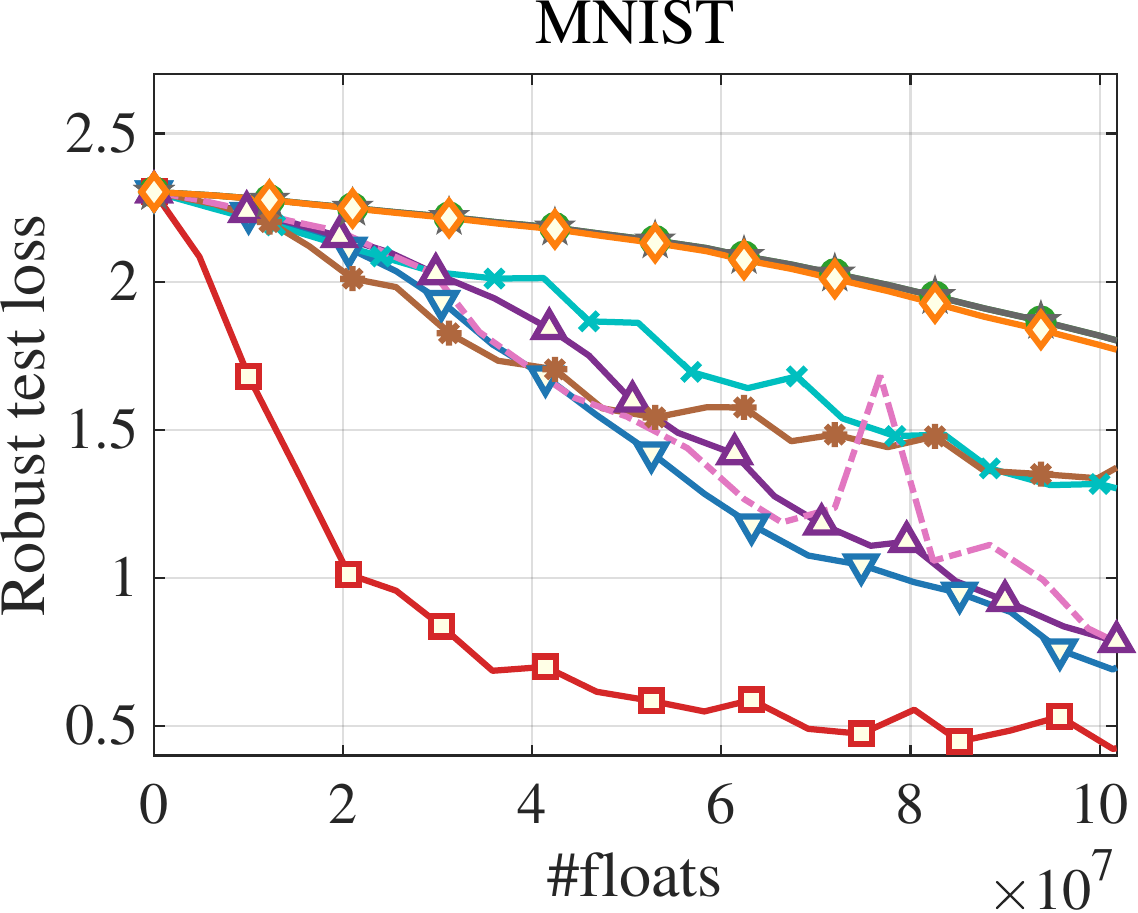}}
    \hfil
    \subfloat{\includegraphics[trim={0cm 0cm 0cm 0cm}, clip, width=.48\linewidth, height=.44\linewidth]{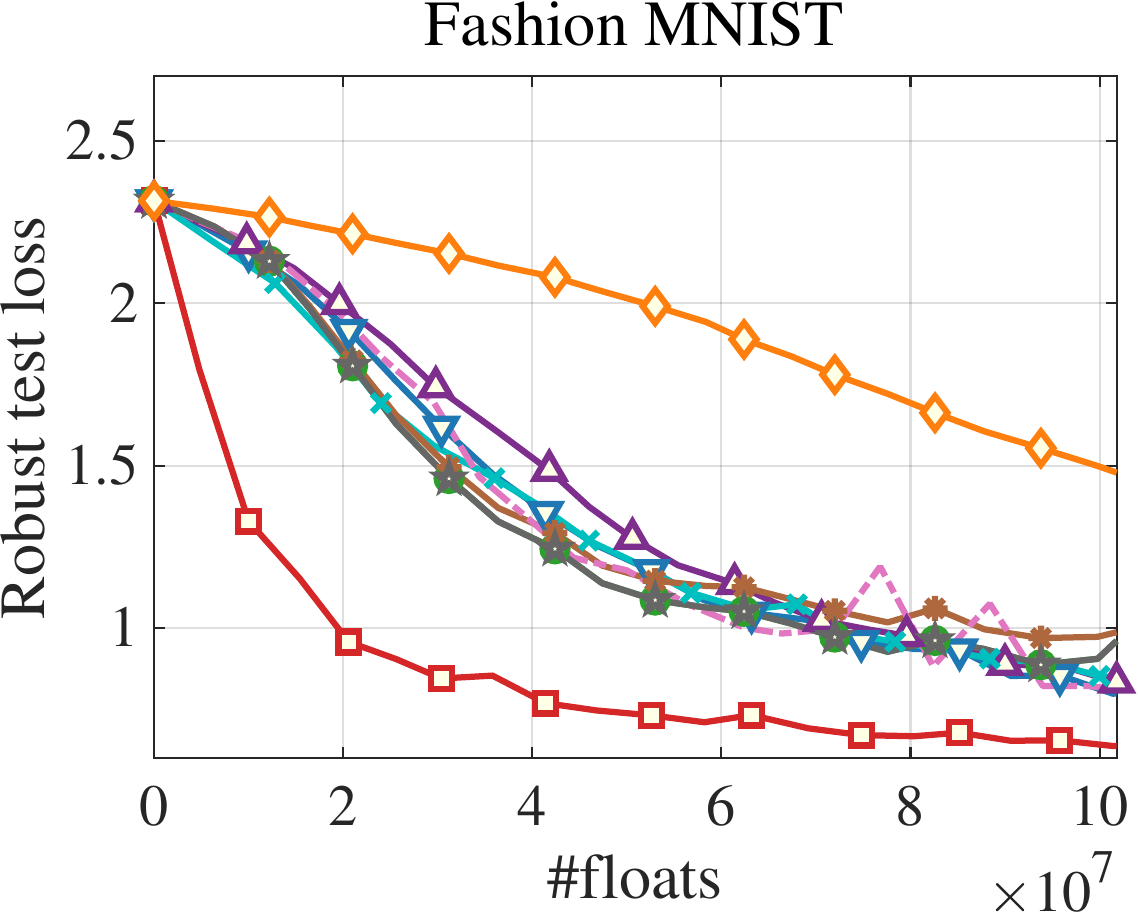}}
    \\[0.2em]
    \subfloat{\includegraphics[trim={-1.55cm 0cm 0cm 0cm}, clip, width=.95\linewidth]{./imgs_journal/legend}}
    \caption{
        Results on the robust neural network training task.
        The top (resp., bottom) row depicts the training (resp., test) results.
        The left (resp., right) column corresponds to the MNIST (resp., Fashion MNIST) dataset.
    }
    \label{figure_robustnn}
\end{figure}

\subsection{Robust Adversarial Network Training}  \label{section_experiment_robustnn}
In the second experiment, we focus on training a classification model that is robust against adversarial noise~\cite{deng2021local}.
We use MNIST and Fashion MNIST~\cite{xiao2017fashion}
datasets in this experiment.
The data partitioning scheme is the same as the AUC experiment and each pixel value of the images is rescaled to $[-1, 1]$.
The performance measure is the robust loss, which is the loss function value of the classification model on the optimally perturbed dataset.
We defer details of the model and the problem formulation to Appendix~\ref{section_experimental_setting_details} due to the space limit.
Similar to the previous experiment, each algorithm runs with the minibatch size $B = 10$ and the number of local steps $K = 12$.
Note that each selected client makes one pass of its local dataset in each round.
The experimental results on the training and test datasets are shown in Figure~\ref{figure_robustnn}.
It shows that {\AlgSTORM} achieves the best performance and {\AlgMB} outperforms the other two theoretically-guaranteed algorithms {\AlgNaive} and Parallel SGDA.

\begin{table}[t]
    \centering
    \begin{tabular}{l|r|r|r}
        \toprule
        \multirow{2}{*}{Algorithm} & IS on & IS on & FID on  \\
                                   & MNIST  & F-MNIST & CelebA  \\
        \midrule
        {\AlgSTORM} (ours)   & \bftab{8.50} & \bftab{8.75} & \bftab{35.84} \\
        {\AlgMB} (ours)      &        8.34  &        8.57  &        37.55 \\
        {\AlgNaive} (ours)   &        8.31  &        7.58  &        44.71  \\
        CODASCA              &        7.85  &        8.18  &        38.33  \\
        CODA+                &        8.33  &        8.23  &        47.61  \\
        Catalyst-Scaffold-S  &        8.13  &        8.38  &        39.37  \\
        Local SGDA+          &        8.29  &        7.61  &        79.24  \\
        Extra Step Local SGD &        7.50  &        7.61  &        59.81  \\
        Parallel SGDA        &        6.08  &        5.74  &        222.65  \\
        \bottomrule
    \end{tabular}
    \caption{
        The best IS obtained by the compared algorithms on MNIST and Fashion MNIST (F-MNIST), and the best FID obtained on the CelebA dataset. Larger IS is better, while smaller FID is better.
    }
    \label{table_gan}
\end{table}

\subsection{GAN Training}  \label{section_experiment_gan}
In the last experiment, we test the performance of our algorithms for training GANs on MNIST, Fashion MNIST, and CelebA~\cite{liu2015faceattributes} datasets.
We measure the performance of the compared algorithms with the Inception Score (IS)~\cite{chavdarova2019reducing} on MNIST and Fashion MNIST, and the more complicated Fr{\'e}chet Inception Distance (FID)~\cite{heusel2017gans} on the harder dataset CelebA.
The experimental setting in more details is deferred to Appendix~\ref{section_experimental_setting_details}.

\begin{figure}[t]
    \captionsetup[subfloat]{farskip=1pt,captionskip=1pt}
    \centering
    \subfloat{\includegraphics[trim={0cm 0cm 0cm 0cm}, clip, width=.46\linewidth]{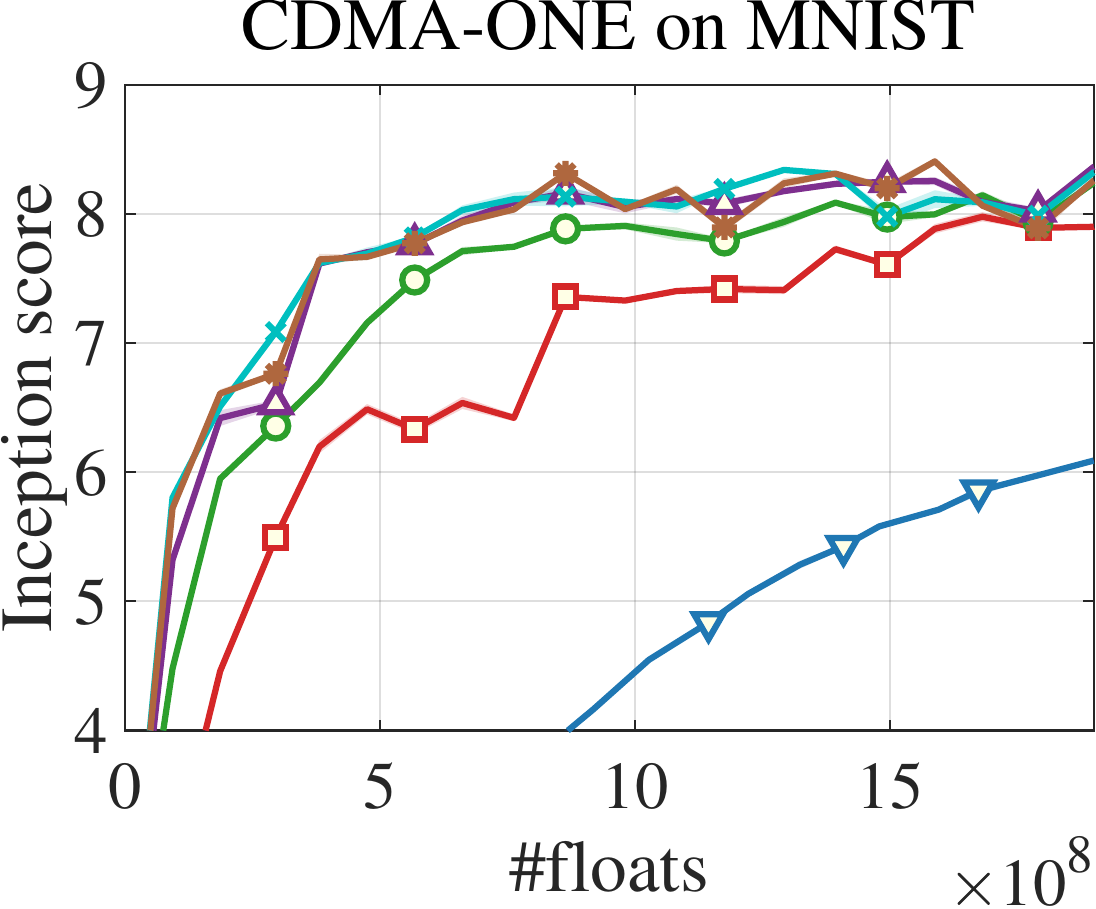}}
    \hfil
    \subfloat{\includegraphics[trim={0cm 0cm 0cm 0cm}, clip, width=.46\linewidth]{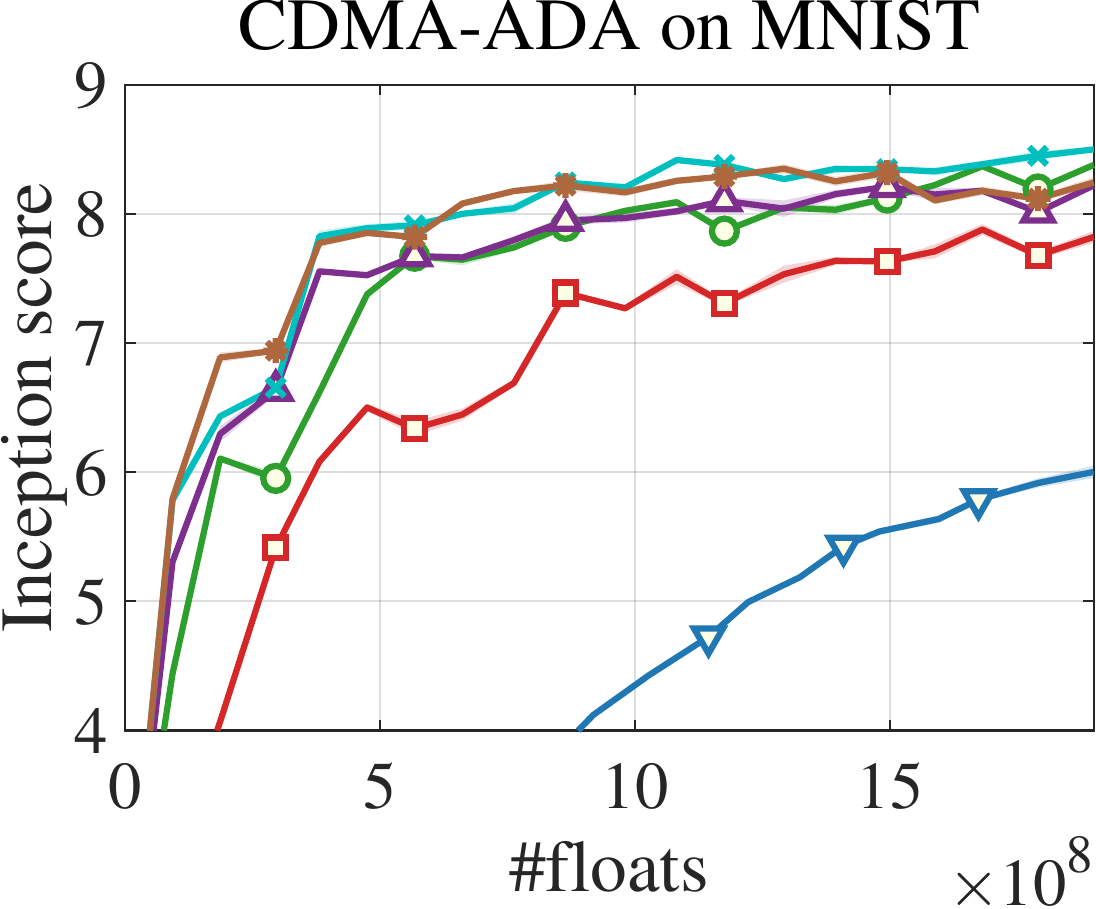}}
    \\[0.05em]
    \subfloat{\includegraphics[trim={-1.3cm 0cm 2cm 0cm}, clip, width=.98\linewidth]{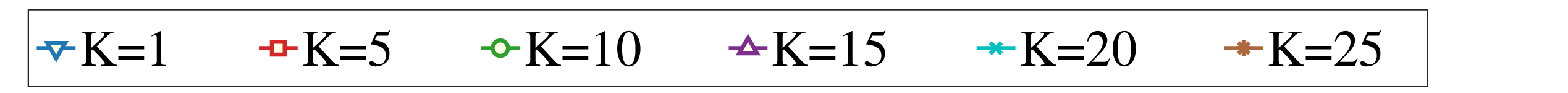}}
    \caption{
        Results of {\AlgMB} (left) and {\AlgSTORM} (right) with different $K$ for GAN training on MNIST. 
    }
    \label{figure_gan_MNIST_K}
\end{figure}

Table~\ref{table_gan} presents the best IS (resp., FID) obtained by the algorithms on MNIST and Fashion MNIST (resp., CelebA) given a fixed amount of communication for each dataset.
The results show that {\AlgSTORM} (resp., {\AlgMB}) achieves the best (resp., second best) performance among the compared algorithms.

In order to validate the effect of local steps, we also conduct empirical studies on the three versions of {\FrameworkAbbr} with different numbers of local steps $K \in \{1, 5, 10, 15, 20, 25\}$ on MNIST and Fashion MNIST.
Figure~\ref{figure_gan_MNIST_K} reports the results of {\AlgMB} and {\AlgSTORM} on MNIST.
We can see that both {\AlgMB} and {\AlgSTORM} with multiple local steps attain significantly higher inception scores than with only one local step,
and achieve the best performance when $K \in \{15, 20, 25\}$ and $K = 20$, respectively.


\section{Conclusion and Future Work}  \label{section_conclusion}
We propose the first practical algorithm for general minimax problems in the cross-device FL setting.
Equipped with the SIWER mechanism and the global correction, our algorithm is resilient to the low client availability and alleviates the impact of slow network connections.
Theoretical analyses and experimental results show the efficiency of our algorithm.
Directions for future work include the extension of {\FrameworkAbbr} to constrained problems and the study of lower bounds for minimax problems in the cross-device FL setting.

\clearpage
\section*{Acknowledgments}
This work is supported by National Key Research and Development Program of China under Grant 2020AAA0107400, Zhejiang Provincial Natural Science Foundation of China under Grant No. LZ18F020002, and National Natural Science Foundation of China (Grant No: 62206248).

\bibliography{references}

\clearpage
\onecolumn
\appendix
\newcommand*{\dictchar}[1]{
    \twocolumn[
    \centerline{\parbox[c][3cm][t]{18cm}{
        \centering
        \fontsize{15}{15}
        \selectfont
        {\bf #1}}}
    ]
}

\dictchar{Appendices of \\ CDMA: A Practical Cross-Device Federated Learning Algorithm for General Minimax Problems}

\vspace{-4em}

\section{Discussions on Challenges in the cross-device FL setting}
To develop a practical algorithm for general minimax problems in the cross-device FL setting, a number of challenges should be overcome~\cite{kairouz2019advances, karimireddy2021breaking}.
Among them, two challenges listed below are negligible in the cross-silo setting but critical to the cross-device setting.
\begin{enumerate}[(i)]
    \item \emph{Low client availability}.
    As Mobile/IoT devices may be temporarily unavailable when battery, network, or idleness requirements are violated, only a fraction of devices are eligible for training at any one time~\cite{kairouz2019advances},
    while the clients in the cross-silo setting are almost always available.
    \item \emph{Limited communication bandwidth}.
    The network connections between the server and mobile/IoT devices are relatively unreliable and slow due to limited resources such as bandwidth and energy.
    As a result, communication is a major bottleneck in the cross-device setting~\cite{li2020federated}.
    In contrast, the network connections in the cross-silo setting are usually fast and reliable.
\end{enumerate}
In addition to the above two challenges, another common challenge that arises in both the cross-device and cross-silo settings of FL is the high data heterogeneity among clients.
It has been shown that the data heterogeneity can introduce the client drift, which in turn leads to slower and unstable convergence~\cite{karimireddy2020scaffold, karimireddy2021breaking}.
To mitigate the influence of data heterogeneity, existing cross-silo minimax algorithms commonly require the full aggregation in each training round.
However, due to the low client availability and limited communication bandwidth of cross-device FL, it is generally prohibitive to synchronize model information over all clients in the cross-device setting,
which precludes the full aggregation strategy.
This makes it more difficult to train a global model in the cross-device setting.

\section{{\AlgNaiveFull}}  \label{section_alg_naive}
{\FrameworkAbbr} with $\beta = 0$, i.e., without global correction, degenerates to a single-phase algorithm detailed in Algorithm~\ref{algorithm_alg_naive}.
\begin{algorithm}[ht]
    \caption{{\AlgNaiveFull} ({\AlgNaive}).}
    \label{algorithm_alg_naive}
    \begin{algorithmic}[1]
    \SUB{Input:}
    initial point $(\xB_{0}, \yB_{0})$,
    the number of rounds $T$,
    the step sizes $\{\eta_t\}_{t=0}^{T-1}$ and $\{\gamma_t\}_{t=0}^{T-1}$,
    and the numbers $\{S_t\}_{t=0}^{T-1}$ of received clients.

    \textbf{Server executes:} 
    \begin{ALC@g}
    \FOR{$t = 0, 1, \ldots, T - 1$}
        \STATE Send $(\xB_t, \yB_t)$ to a random subset $\hat{\SM}_t$ of clients;

        \STATE Receive local iterates $(\xB_{t, i}^{(K)}, \yB_{t, i}^{(K)})$ from each client $i \in \SM_t \subseteq \hat{\SM}_t$, where $\SM_t$ is the subset of $\hat{\SM}_t$ containing the first $S_t$ clients that successfully respond;

        \STATE $\xB_{t + 1} \leftarrow \frac{1}{|\SM_t|} \sum_{i \in \SM_t} \xB_{t, i}^{(K)}$, $\yB_{t + 1} \leftarrow \frac{1}{|\SM_t|} \sum_{i \in \SM_t} \yB_{t, i}^{(K)}$;
    \ENDFOR
    \end{ALC@g}

    \vspace{1em}

    \textbf{Client $i \in \hat{\SM}_t$ executes:}
    \begin{ALC@g}
        \STATE Initialize local model $\xB_{t, i}^{(0)} \leftarrow \xB_t$, $\yB_{t, i}^{(0)} \leftarrow \yB_t$; \
    
        \FOR{$k = 0, 1, \ldots, K - 1$}
            \STATE Sample a minibatch $\BM_{t, i}^{(k)}$ from local data;
    
            \STATE $\xB_{t, i}^{(k + 1)} \leftarrow
                \xB_{t, i}^{(k)} - \eta_t \nabla_{\xB} F_i(\xB_{t, i}^{(k)}, \yB_{t, i}^{(k)}; \BM_{t, i}^{(k)})$;
    
            \STATE $\yB_{t, i}^{(k + 1)} \leftarrow
                \yB_{t, i}^{(k)} + \gamma_t \nabla_{\yB} F_i(\xB_{t, i}^{(k)}, \yB_{t, i}^{(k)}; \BM_{t, i}^{(k)})$;
        \ENDFOR
        \STATE Send $(\xB_{t, i}^{(K)}, \yB_{t, i}^{(K)})$ to the server; 
    \end{ALC@g}
\end{algorithmic}
\end{algorithm}

\section{Deferred Analysis}  \label{section_deferred_analysis}

\subsection{Discussions on Assumptions~\ref{assumption_bgd}-\ref{assumption_pl}}  \label{section_discussion_on_assumptions}

\paragraph{The bounded gradient dissimilarity assumption.}
Assumption~\ref{assumption_bgd} (or a similar assumption) is commonly used in most existing analyses of FL algorithms~\cite{kairouz2019advances, rasouli2020fedgan, hou2021efficient, yuan2021federated, deng2020distributionally}.
The Scaffold method~\cite{karimireddy2020scaffold}, which is designed for minimization problems, is an exception.
However, its convergence rate depends on $N/S$, where $N$ is the total number of clients and $S$ is the number of clients participating per round.
In typical cross-device federated learning applications, $N$ may be extremely large compared with $S$~\cite[Table 2]{kairouz2019advances}.
Thus, we follow the conventional setting and assume that the bounded gradient dissimilarity condition holds.

\paragraph{Lipschitz continuous gradients.}
Assumption~\ref{assumption_smooth} is also known as the Lipschitz smoothness assumption, and is commonly used in the minimax optimization literature~\cite{lin2020gradient, luo2020stochastic, xu2020enhanced, qiu2020single, reisizadeh2020robust, deng2021local}.
This assumption immediately implies that both $f_i$ and $f$ also have $L_f$-Lipschitz continuous gradients.

\paragraph{The Polyak-{\L}ojasiewicz (P{\L}) condition.}
The P{\L} condition in Assumption~\ref{assumption_pl} is a common condition in the optimization literature and is satisfied by many practical problems.
Actually, it holds for any strongly concave functions (e.g., the minimax formulation of the AUC maximization problem~\citep{liu2020stochastic}), and many non-concave functions arising in over-parameterized deep networks, deep networks with linear activations, one-hidden-layer networks with Leaky ReLU activation~\citep{du2019gradient, allen2019convergence, charles2018stability}, etc.
Under the P{\L} condition and Assumption~\ref{assumption_smooth}, we define the condition number of problem~\eqref{eq_problem} as $\kappa := \Lot / \mu$.
Note that $\kappa \ge 1$ (see Lemma~\ref{lemma_condition_number} below).

\subsection{Additional Notation and a Useful Lemma}
In the rest of the paper, we use $\zB$ to denote the concatenation of primal and dual variables $(\xB^\top, \yB^\top)^\top$ for ease of notation.
For example, $\zB_{t, i}^{(k)}$ represents client $i$'s local iterate $((\xB_{t, i}^{(k)})^\top, (\yB_{t, i}^{(k)})^\top)^\top$ at the $k$-th local iteration in round $t$.
In addition, We denote $\hat{\eta}_t \!:=\! \eta_t K$, $\hat{\gamma}_t \!:=\! \gamma_t K$, $\tilde{\nabla}_{\xB}^{(k)} F_i(\zB) \!:=\! \nabla_{\xB} F_i(\zB; \BM_{t, i}^{(k)})$ and $\tilde{\nabla}_{\yB}^{(k)} F_i(\zB) \!:=\! \nabla_{\yB} F_i(\zB; \BM_{t, i}^{(k)})$. 
The following property is required in our analyses.
\begin{lemma}  \label{lemma_condition_number}
    Suppose that Assumptions~\ref{assumption_smooth} and~\ref{assumption_pl} hold.
    Define the condition number of $\Phi$ as $\kappa := L_f / \mu$.
    Then, $\kappa \ge 1$.
\end{lemma}

\begin{proof}
We denote $\yB^* \in \mathrm{argmax}_{\yB \in \RBB^q} \ f(\xB, \yB)$ and define $g(\yB) := f(\xB, \yB)$ given a fixed $\xB$.
Let $\zB = \yB - \frac{1}{L_f} g(\yB)$.
By Assumption~\ref{assumption_smooth}, we have
$$
\begin{aligned}[b]
    &g(\yB^*) - g(\yB)
    = g(\yB^*) - g(\zB) + g(\zB) - g(\yB) \\
    \ge{}& \langle \nabla g(\yB), \zB - \yB \rangle - {\textstyle \frac{L_f}{2}} \| \yB - \zB \|^2
    = {\textstyle \frac{1}{2 L_f}} \| \nabla g(\yB) \|^2.
\end{aligned}
$$
Combining the above inequality and Assumption~\ref{assumption_pl} leads to
$L_f / \mu \ge 1$.
This completes the proof.
\end{proof}

\subsection{Proofs of the Results in Section~\ref{section_analysis_naive}}  \label{section_analysis_naive_appendix}
In our analysis of {\AlgNaive}, we define $\eB'_{\xB, t}$ and $\eB'_{\yB, t}$ as
$$
\begin{cases}
    \eB'_{\xB, t} \!=\! \frac{1}{|\SM_t|} \! \sum_{i \in \SM_t} \! \big( \frac{1}{K} \sum_{k\!=\!0}^{K\!-\!1} \! \nabla_{\xB} F_i(\zB_{t, i}^{(k)}; \BM_{t, i}^{(k)}) \!-\! \nabla_{\xB} f_i(\zB_t) \big), \\
    \eB'_{\yB, t} \!=\! \frac{1}{|\SM_t|} \! \sum_{i \in \SM_t} \! \big( \frac{1}{K} \sum_{k\!=\!0}^{K\!-\!1} \! \nabla_{\yB} F_i(\zB_{t, i}^{(k)}; \BM_{t, i}^{(k)}) \!-\! \nabla_{\yB} f_i(\zB_t) \big),
\end{cases}
$$
where $\zB_{t, i}^{(k)}$ denotes $((\xB_{t, i}^{(k)})^\top, (\yB_{t, i}^{(k)})^\top)^\top$.
With this notation, we establish the upper bound of the expected gradient norm of $\Phi(\xB_t)$ in the following lemma.

\begin{lemma}  \label{lemma_one_round_progress}
Suppose that Assumptions~\ref{assumption_smooth}-\ref{assumption_pl} hold and that $\eta_t \le \frac{1}{4 L_{\Phi} K}$ in {\AlgNaive}.
Then,
$$
\resizebox{\linewidth}{!}{$
\begin{aligned}[b] 
    &{\textstyle \frac{\eta_t K}{2}} \EBB\|\nabla \Phi(\xB_t)\|^2 \\
    \le{}& \EBB[\Phi(\xB_t) - \Phi(\xB_{t+1})]
        \!+\! {\textstyle \frac{\eta_t K \Lot^2}{2 \mu}} \EBB[\Phi(\xB_t) \!-\! f(\zB_t)] \\
        &\!+\! {\textstyle \frac{5 \eta_t K}{4}} \EBB\|\eB_{\xB, t}'\|^2
        \!+\! {\textstyle \frac{L_{\Phi} \eta_t^2 K^2}{S}} \sigma_1^2
        \!-\! {\textstyle \frac{\eta_t K}{8}} \EBB\|\nabla_{\xB} f(\zB_t)\|^2.
\end{aligned}
$}
$$
\end{lemma}

\begin{proof}
By the update rule of $\xB_t^{(k)}$ and $\yB_t^{(k)}$, we have
\begin{equation}  \label{eq_global_update_rule_mb}
\begin{cases}
    \xB_{t+1} = \xB_t - \hat{\eta}_t (\uB_t' + \eB'_{\xB, t}), \\
    \yB_{t+1} = \yB_t + \hat{\gamma_t} (\vB_t' + \eB'_{\yB, t}),
\end{cases}
\end{equation}
where $\uB_t'$ and $\vB_t'$ are defined as
$\uB_t' = \frac{1}{|\SM_t|} \sum_{i \in \SM_t} \nabla_{\xB} f_i(\xB_t, \yB_t)$
and $\vB_t' = \frac{1}{|\SM_t|} \sum_{i \in \SM_t} \nabla_{\yB} f_i(\xB_t, \yB_t)$, respectively.
By the smoothness of $L_{\Phi}$, we can bound
\begin{align}
    &\EBB[ \Phi(\xB_{t+1}) - \Phi(\xB_t)] \nonumber \\
    \le& \EBB \big[ \langle \nabla \Phi(\xB_t), \xB_{t+1} - \xB_t \rangle
        + {\textstyle \frac{L_{\Phi}}{2}} \|\xB_{t+1} \!-\! \xB_t\|^2 \big] \nonumber \\
    \overset{(a)}{=}& \!-\! \hat{\eta}_t \EBB[\langle \nabla \Phi(\xB_t), \uB_t' \!+\! \eB'_{\xB, t} \rangle]
        + {\textstyle \frac{L_{\Phi} \hat{\eta}_t^2}{2}} \EBB\|\uB_t' \!+\! \eB'_{\xB, t}\|^2 \nonumber \\
    \overset{(b)}{=}& \!-\! \hat{\eta}_t \EBB[\langle \nabla \Phi(\xB_t), \nabla_{\xB} f(\zB_t) \!+\! \eB'_{\xB, t} \rangle]
        \!+\! {\textstyle \frac{L_{\Phi} \hat{\eta}_t^2}{2}} \EBB\|\uB_t' \!+\! \eB'_{\xB, t}\|^2 \nonumber \\
    \overset{(c)}{\le}& \hat{\eta}_t \EBB[\|\nabla_{\xB} f(\zB_t) \!-\! \nabla \Phi(\xB_t)\|^2 \!+\! \| \eB'_{\xB, t} \|^2]
        \!-\! {\textstyle \frac{\hat{\eta}_t}{2}} \EBB\|\nabla \Phi(\xB_t)\|^2 \nonumber \\
        &\!-\! \big( {\textstyle \frac{\hat{\eta}_t}{2}} \!-\! L_{\Phi} \hat{\eta}_t^2 \big) \EBB\|\eB'_{\xB, t} \!+\! \nabla_{\xB} f(\zB_t)\|^2
        \!+\! {\textstyle \frac{L_{\Phi} \hat{\eta}_t^2}{S}} \sigma_1^2 \nonumber \\
    \overset{(d)}{\le}& \! \hat{\eta}_t \EBB[
        \Lot^2 \|\yB_t \!-\! \yB^*(\xB_t)\|^2 \!+\! \| \eB'_{\xB, t} \|^2]
        \!-\! {\textstyle \frac{\hat{\eta}_t}{2}} \EBB\|\nabla \Phi(\xB_t)\|^2 \nonumber \\
        &\!+\! {\textstyle \frac{L_{\Phi} \hat{\eta}_t^2 \sigma_1^2}{S}}
        \!-\! {\textstyle \frac{\hat{\eta}_t}{8}} \EBB\|\nabla_{\xB} f(\zB_t)\|^2
        \!+\! {\textstyle \frac{\hat{\eta}_t}{4}} \EBB\| \eB'_{\xB, t}\|^2, \nonumber
\end{align}
where (a) follows from~\eqref{eq_global_update_rule_mb},
(b) holds because of the unbiasedness of $\uB_t$,
(c) follows from Assumption~\ref{assumption_bgd} and the fact that $\|\aB + \bB\|^2 \le 2 \|\aB\|^2 + 2 \|\bB\|^2$,
and (d) follows from Assumption~\ref{assumption_smooth}, the condition $\eta_t \le 1 / (4 L_{\Phi} K)$ and the fact that $\|\aB + \bB\|^2 \ge \frac{1}{2}\|\aB\|^2 - \|\bB\|^2$.
By \cite[Theorem 2]{karimi2016linear}, the P{\L} condition (Assumption~\ref{assumption_pl}) implies that $\forall (\xB, \yB) \in \RBB^{p \times q}$,
\begin{equation}  \label{eq_pl_property}
    \Phi(\xB) - f(\xB, \yB) \ge 2 \mu \| \yB - \yB^*(\xB) \|^2.
\end{equation}
Combining the above two inequalities yields the desired result.
\end{proof}

To proceed our analysis, we provide upper bounds of the terms $\EBB\|\eB_{\xB, t}'\|^2$ and $\EBB[ \Phi(\xB_{t+1}) - f(\zB_{t+1})]$ in Lemmas~\ref{lemma_local_iterate_error_naive} and~\ref{lemma_phi_f_diff_bound} below, respectively.
\begin{lemma}  \label{lemma_local_iterate_error_naive}
Suppose that Assumptions~\ref{assumption_bgd}, \ref{assumption_smooth}, and~\ref{assumption_intra_client_variance} hold and $\eta_t, \gamma_t \le \frac{1}{2 K L_f}$ in {\AlgNaive}.
If the minibatches $\BM_{t, i}^{(0)}, \ldots, \BM_{t, i}^{(K-1)}$ in {\AlgNaive} are drawn in a random reshuffling manner and $K$ is an integral multiple of the epoch length,
then both $\EBB\|\eB'_{\xB, t}\|^2$ and $\EBB\|\eB'_{\yB, t}\|^2$ are bounded from above by
$
    \frac{4 L_f^2}{3} \eta_t^2 K^2 \big( \EBB\| \nabla_{\xB} f(\zB_t)\|^2 + \sigma_1^2 + G_1^2 \big)
        + \frac{4 L_f^2}{3} \gamma_t^2 K^2 \big( \EBB\| \nabla_{\yB} f(\zB_t)\|^2 + \sigma_2^2 + G_2^2 \big).
$
\end{lemma}

\begin{proof}
By the definition of $\eB'_{\xB, t}$, we have
\begin{align*}
    \EBB\|\eB'_{\xB, t}\|^2
    ={}& \EBB \Big\| {\textstyle \frac{1}{|\SM_t| K} \sum_{i, k}} \big( \tilde{\nabla}_{\xB}^{(k)} F_i(\zB_{t, i}^{(k)}) - \tilde{\nabla}_{\xB}^{(k)} F_i(\zB_t) \big) \\
        &\!+\! {\textstyle \frac{1}{|\SM_t|} \sum_{i \in \SM_t}} \big( {\textstyle \frac{1}{K} \sum_{k=0}^{K-1}} \tilde{\nabla}_{\xB}^{(k)} F_i(\zB_t) \!-\! \nabla_{\xB} f_i(\zB_t) \big) \Big\|^2 \\
    ={}& \EBB \Big\| {\textstyle \frac{1}{|\SM_t| K} \sum_{i, k}} \big( \tilde{\nabla}_{\xB}^{(k)} F_i(\zB_{t, i}^{(k)}) - \tilde{\nabla}_{\xB}^{(k)} F_i(\zB_t) \big) \Big\|^2 \\
    \le{}& {\textstyle \frac{1}{|\SM_t| K} \sum_{i, k}} \EBB \big\| \tilde{\nabla}_{\xB}^{(k)} F_i(\zB_{t, i}^{(k)}) - \tilde{\nabla}_{\xB}^{(k)} F_i(\zB_t) \big\|^2 \\
    \le{}& {\textstyle \frac{L_f^2}{|\SM_t| K} \sum_{i, k}} \EBB \| \zB_{t, i}^{(k)} - \zB_t \|^2,
\end{align*}
where the second equality follows from the condition that the minibatches $\BM_{t, i}^{(1)}, \ldots, \BM_{t, i}^{(K-1)}$ in {\AlgNaive} are drawn in a random reshuffling manner and $K$ is an integral multiple of the epoch length.
Similarly, we have $\EBB\|\eB'_{\yB, t}\|^2 \le \frac{L_f^2}{S K} \sum_{i, k} \EBB \| \zB_{t, i}^{(k)} - \zB_t \|^2$.

If $k \ge 1$,
$
    \EBB\|\zB_{t, i}^{(k)} - \zB_t\|^2
    = \eta_t^2 \EBB \|\sum_{k'=0}^{k-1} \tilde{\nabla}_{\xB}^{(k)} F_i(\zB_{t, i}^{(k)}) \|^2
        + \gamma_t^2 \EBB \|\sum_{k'=0}^{k-1} \tilde{\nabla}_{\yB}^{(k)} F_i(\zB_{t, i}^{(k)}) \|^2.
$
The first term on the RHS can be bounded by
\begin{align}
    &\EBB \| {\textstyle \sum_{k'=0}^{k-1}} \! \tilde{\nabla}_{\xB}^{(k)} F_i(\zB_{t, i}^{(k)}) \|^2 \nonumber \\
    \le&  2 \EBB \Big[ \Big\| {\textstyle \sum_{k'=0}^{k-1}}
        \tilde{\nabla}_{\xB}^{(k')} F_i(\zB_{t, i}^{(k')})
        - \tilde{\nabla}_{\xB}^{(k')} F_i(\zB_t)  \Big\|^2 \nonumber \\
        &+ \Big\| {\textstyle \sum_{k'=0}^{k-1}}
        \tilde{\nabla}_{\xB}^{(k')} F_i(\zB_t) \Big\|^2 \Big] \nonumber \\
    \le{}& 2 k L_f^2 {\textstyle \sum_{k'=0}^{k-1}} \EBB\| \zB_{t, i}^{(k')} - \zB_t\|^2
        + 2 \EBB \Big\|{\textstyle \sum_{k'=0}^{k-1}}
        \tilde{\nabla}_{\xB}^{(k')} F_i(\zB_t) \Big\|^2 \nonumber \\
    \le{}& 2 k L_f^2 {\textstyle \sum_{k'=0}^{k-1}} \EBB\| \zB_{t, i}^{(k')} - \zB_t\|^2
        + 2 k^2 \EBB\|\nabla_{\xB} f_i(\zB_t)\|^2 \nonumber \\
        &+ 2 k^2 \EBB \big\| \textstyle{\frac{1}{k} \sum_{k'=0}^{k-1}} \tilde{\nabla}_{\xB}^{(k')} F_i(\zB_t) - \nabla_{\xB} f_i(\zB_t) \big\|^2,   \label{eq_alg_naive_error_primal_expand}
\end{align}
where the last inequality follows the fact that $\EBB[\nabla_{\xB} F_i(\zB_t; \BM_{t, i}^{(k')})] = \EBB[\nabla_{\xB} f_i(\zB_t)]$.
Let us denote the minibatch size $|\BM_{t, i}^{(k')}| = B$.
Recall that the minibatches $\BM_{t, i}^{(0)}, \ldots, \BM_{t, i}^{(K-1)}$ are drawn in a random reshuffling manner and that $\nabla_{\xB} F_i(\zB_t; \BM_{t, i}^{(k')}) = \frac{1}{B} \sum_{\zeta \in \BM_{t, i}^{(k')}} \nabla_{\xB} F_i(\zB_t; \zeta)$.
By Assumption~\ref{assumption_intra_client_variance} and~\cite[Lemma 1]{mishchenko2020random}, we can bound
$$
\begin{aligned}
    &\EBB \Big\| {\textstyle \frac{1}{k} \sum_{k'=0}^{k-1}} \nabla_{\xB} F_i(\zB_t; \BM_{t, i}^{(k')}) - \nabla_{\xB} f_i(\zB_t) \Big\|^2 \\
    \le& \frac{n_i - 1 - \mathrm{mod}(k B - 1, n_i)}{(\mathrm{mod}(k B - 1, n_i) + 1) (n_i - 1)} G_1^2
    \le G_1^2.
\end{aligned}
$$
Plugging the above inequality into~\eqref{eq_alg_naive_error_primal_expand} yields
\begin{align*}
    &\EBB \big\| {\textstyle \sum_{k'=0}^{k-1}} \tilde{\nabla}_{\xB}^{(k')} F_i(\zB_{t, i}^{(k)}) \big\|^2 \\
    \le{}& 2 k L_f^2 {\textstyle \sum_{k'=0}^{k-1}} \EBB[\| \zB_{t, i}^{(k')} \!-\! \zB_t\|^2]
        \!+\! 2 k^2 \EBB\|\nabla_{\xB} f_i(\zB_t)\|^2
        \!+\! 2 k^2 G_1^2 \\
    \le{}& 2 k L_f^2 {\textstyle \sum_{k'=0}^{k-1}} \EBB\| \zB_{t, i}^{(k')} \!-\! \zB_t\|^2
        + 2 k^2 \EBB[\|\nabla_{\xB} f(\zB_t)\|^2 \\
            &+ \|\nabla_{\xB} f_i(\zB_t) \!-\! \nabla_{\xB} f(\zB_t)\|^2]
        + 2 k^2 G_1^2 \\
    \le{}& 2 k L_f^2 \! {\sum_{k' = 0}^{k - 1}} \! \EBB\| \zB_{t, i}^{(k')} \!-\! \zB_t\|^2
        \!+\! 2 k^2 \EBB\|\nabla_{\xB} f(\zB_t)\|^2
        \!+\! 2 k^2 (\sigma_1^2 \!+\! G_1^2),
\end{align*}
where the last inequality follows from Assumption~\ref{assumption_bgd}.
Similarly, we can show that
$$
\begin{aligned}
    &\EBB \big\| {\textstyle \sum_{k'=0}^{k-1}} \tilde{\nabla}_{\yB}^{(k')} F_i(\zB_{t, i}^{(k)}) \big\|^2 \\
    \le{}& 2 k L_f^2 \! {\sum_{k' = 0}^{k - 1}} \! \EBB\| \zB_{t, i}^{(k')} \!-\! \zB_t\|^2
        \!+\! 2 k^2 \EBB\|\nabla_{\yB} f(\zB_t)\|^2
        \!+\! 2 k^2 (\sigma_2^2 \!+\! G_2^2).
\end{aligned}
$$
Combining the above two inequalities, we have
$$
\begin{aligned}
    \EBB\|\zB_{t, i}^{(k)} - \zB_t\|^2
    \le{}& 2 k L_f^2 (\eta_t^2 + \gamma_t^2) {\textstyle \sum_{k'=0}^{k-1}} \EBB\|\zB_{t, i}^{(k')} - \zB_t\|^2 \\
        &\!+\! 2 \eta_t^2 k^2 \EBB\|\nabla_{\xB} f(\zB_t)\|^2
        \!+\! 2 \gamma_t^2 k^2 \EBB\|\nabla_{\yB} f(\zB_t)\|^2 \\
        &\!+\! 2 \eta_t^2 k^2 (\sigma_1^2 + G_1^2)
        + 2 \gamma_t^2 k^2 (\sigma_2^2 + G_2^2).
\end{aligned}
$$
Summing the above inequality from $k=1$ to $k=K-1$ yields
$$
\begin{aligned}
    &{\textstyle \sum_{k=0}^{K-1}} \EBB\|\zB_{t, i}^{(k)} - \zB_t\|^2 \\
    \le& L_f^2 (\hat{\eta}_t^2 \!+\! \hat{\gamma}_t^2) {\textstyle \sum_{k=0}^{K-1}} \EBB\|\zB_{t, i}^{(k)} \!-\! \zB_t\|^2
        \!+\! {\textstyle \frac{2 \hat{\eta}_t^2 K}{3}} \big(
            \EBB\|\nabla_{\xB} f(\zB_t)\|^2  \\
            &\!+\! \sigma_1^2 \!+\! G_1^2 \big)
        \!+\! {\textstyle \frac{2 \hat{\gamma}_t^2 K}{3}} \big(
            \EBB\|\nabla_{\yB} f(\zB_t)\|^2 \!+\! \sigma_2^2 \!+\! G_2^2 \big).
\end{aligned}
$$
Summing the above inequality for $i \in \SM_t$ and applying the condition $\eta_t, \gamma_t \le 1 / (2 L_f K)$ leads to the desired result.
\end{proof}

\begin{lemma}  \label{lemma_phi_f_diff_bound}
Under Assumptions~\ref{assumption_bgd}-\ref{assumption_pl}, let $\gamma_t \le 1 / (2 L_f K)$ and $\eta_t \le \min\{1/(2 L_f K), 1 / (4 L_{\Phi} K)\}$ in {\AlgNaive}.
Then
$$
\begin{aligned}
    &\EBB[ \Phi(\xB_{t+1}) - f(\zB_{t+1})] \\
    \le& (1 \!-\! \frac{\mu \hat{\gamma}_t}{2} \!+\! \frac{\hat{\eta}_t \Lot^2}{2 \mu}) \EBB[\Phi(\xB_t) \!-\! f(\zB_t)]
        \!-\! \frac{\hat{\eta}_t}{2} \EBB\|\nabla \Phi(\xB_t)\|^2 \\
        &\!+\! \frac{11 \hat{\eta}_t}{4} \EBB\|\eB_{\xB, t}'\|^2
        \!+\! \frac{\hat{\gamma}_t}{2} \EBB\|\eB_{\yB, t}'\|^2
        \!+\! \frac{19 \hat{\eta}_t}{8} \EBB\|\nabla_{\xB} f(\zB_t)\|^2 \\
        &\!+\! \frac{\hat{\eta}_t^2 (L_f \!+\! L_{\Phi})}{S} \sigma_1^2
        \!+\! \frac{\hat{\gamma}_t^2 L_f}{S} \sigma_2^2
        \!-\! \frac{\hat{\gamma}_t}{4} \EBB\|\nabla_{\yB} f(\zB_t)\|^2.
\end{aligned}
$$
\end{lemma}

\begin{proof}
First, we rewrite $\Phi(\xB_{t+1}) - f(\zB_{t+1})$ as
$$
\begin{aligned}
    \Phi(\xB_{t+1}) - f(\zB_{t+1})
    ={}& \Phi(\xB_{t+1}) - \Phi(\xB_t)
        + \Phi(\xB_t) - f(\zB_t) \\
        &+ f(\zB_t) - f(\zB_{t+1}).
\end{aligned}
$$
By the smoothness of $f$, we have
\begin{align*}
    &\EBB[f(\zB_t) - f(\zB_{t+1})] \\
    \overset{(a)}{\le}{}& \EBB \Big[ \hat{\eta}_t \langle \nabla_{\xB} f(\zB_t), \uB_t' + \eB_{\xB, t}' \rangle
    - \hat{\gamma}_t \langle \nabla_{\yB} f(\zB_t), \vB_t' + \eB_{\yB, t}' \rangle \\
    &+ {\textstyle \frac{L_f}{2}} \big( \hat{\eta}_t^2 \|\uB_t' \!+\! \eB_{\xB, t}'\|^2 + \hat{\gamma}_t^2 \|\vB_t' \!+\! \eB_{\yB, t}'\|^2 \big) \Big] \\
    \overset{(b)}{\le}& \EBB \Big[ \frac{3 \hat{\eta}_t}{2} \|\nabla_{\xB} f(\zB_t)\|^2
        \!+\! {\textstyle \frac{\hat{\eta}_t}{2}} \|\eB_{\xB, t}'\|^2
        \!+\! {\textstyle \frac{\hat{\gamma}_t}{2}} \|\eB_{\yB, t}'\|^2 \\
        &\!-\! {\textstyle \frac{\hat{\gamma}_t}{2}} \|\nabla_{\yB} f(\zB_t)\|^2
        \!-\! {\textstyle \frac{\hat{\gamma}_t}{2}} \|\nabla_{\yB} f(\zB_t) \!+\! \eB_{\yB, t}'\|^2 \\
        &\!+\! {\textstyle \frac{L_f}{2}} \Big( \hat{\eta}_t^2 \|\uB_t' \!+\! \eB_{\xB, t}'\|^2
            + \hat{\gamma}_t^2 \|\vB_t' \!+\! \eB_{\yB, t}'\|^2
            \Big)
        \Big] \\
    =& \EBB \Big[ {\textstyle \frac{3 \hat{\eta}_t}{2}} \|\nabla_{\xB} f(\zB_t)\|^2
        \!+\! {\textstyle \frac{\hat{\eta}_t}{2}} \|\eB_{\xB, t}'\|^2
        \!+\! {\textstyle \frac{\hat{\gamma}_t}{2}} \|\eB_{\yB, t}'\|^2 \\
        &\!-\! {\textstyle \frac{\hat{\gamma}_t}{2}} \|\nabla_{\yB} f(\zB_t)\|^2
        \!-\! {\textstyle \frac{\hat{\gamma}_t}{2}} \|\nabla_{\yB} f(\zB_t) \!+\! \eB_{\yB, t}'\|^2 \\
        &\!+\! {\textstyle \frac{L_f}{2}} \Big( \hat{\eta}_t^2 \|\uB_t' \!-\! \nabla_{\xB} f(\zB_t) \!+\! \nabla_{\xB} f(\zB_t) \!+\! \eB_{\xB, t}'\|^2 \\
            &+ \hat{\gamma}_t^2 \|\vB_t' \!-\! \nabla_{\yB} f(\zB_t) \!+\! \nabla_{\yB} f(\zB_t) \!+\! \eB_{\yB, t}'\|^2
            \Big)
        \Big] \\
    \overset{(c)}{\le}& \EBB \Big[ \big({\textstyle \frac{3 \hat{\eta}_t}{2}} + 2 \hat{\eta}_t^2 L_f \big) \|\nabla_{\xB} f(\zB_t)\|^2
        \!+\! \big({\textstyle \frac{\hat{\eta}_t}{2}} \!+\! 2 \hat{\eta}_t^2 L_f \big) \|\eB_{\xB, t}'\|^2 \\
        &\!+\! {\textstyle \frac{\hat{\gamma}_t}{2}} \|\eB_{\yB, t}'\|^2
        \!-\! \big({\textstyle \frac{\hat{\gamma}_t}{2}} \!-\! \hat{\gamma}_t^2 L_f \big) \|\nabla_{\yB} f(\zB_t) \!+\! \eB_{\yB, t}'\|^2 \Big] \\
        &\!-\! {\textstyle \frac{\hat{\gamma}_t}{2}} \EBB[\|\nabla_{\yB} f(\zB_t)\|^2]
        \!+\! {\textstyle \frac{\hat{\eta}_t^2 L_f}{|\SM_t|}} \sigma_1^2
        \!+\! {\textstyle \frac{\hat{\gamma}_t^2 L_f}{|\SM_t|}} \sigma_2^2  \\
    \overset{(d)}{\le}& \EBB \Big[ \big({\textstyle \frac{3 \hat{\eta}_t}{2}} + 2 \hat{\eta}_t^2 L_f \big) \|\nabla_{\xB} f(\zB_t)\|^2
        \!+\! \big({\textstyle \frac{\hat{\eta}_t}{2}} \!+\! 2 \hat{\eta}_t^2 L_f \big) \|\eB_{\xB, t}'\|^2 \\
        &\!+\! {\textstyle \frac{\hat{\gamma}_t}{2}} \|\eB_{\yB, t}'\|^2 \Big]
        \!-\! {\textstyle \frac{\hat{\gamma}_t}{4}} \EBB[\|\nabla_{\yB} f(\zB_t)\|^2] \\
        &\!-\! {\textstyle \frac{\hat{\gamma}_t}{4}} \EBB[\|\nabla_{\yB} f(\zB_t)\|^2]
        \!+\! {\textstyle \frac{\hat{\eta}_t^2 L_f}{|\SM_t|}} \sigma_1^2
        \!+\! {\textstyle \frac{\hat{\gamma}_t^2 L_f}{|\SM_t|}} \sigma_2^2  \\
    \overset{(e)}{\le}& \EBB \Big[ \big({\textstyle \frac{3 \hat{\eta}_t}{2}} \!+\! 2 \hat{\eta}_t^2 L_f \big) \|\nabla_{\xB} f(\zB_t)\|^2
        \!+\! \big({\textstyle \frac{\hat{\eta}_t}{2}} \!+\! 2 \hat{\eta}_t^2 L_f \big) \|\eB_{\xB, t}'\|^2 \\
        &\!+\! {\textstyle \frac{\hat{\gamma}_t}{2}} \|\eB_{\yB, t}'\|^2
        \!-\! {\textstyle \frac{\hat{\gamma}_t}{4}} \|\nabla_{\yB} f(\zB_t)\|^2
        \!-\! {\textstyle \frac{\mu \hat{\gamma}_t}{2}} (\Phi(\xB_t) - f(\zB_t)) \Big] \\
        &\!+\! {\textstyle \frac{\hat{\eta}_t^2 L_f}{S}} \sigma_1^2
        \!+\! {\textstyle \frac{\hat{\gamma}_t^2 L_f}{S}} \sigma_2^2,
\end{align*}
where (a) follows from~\eqref{eq_global_update_rule_mb} and Assumption~\ref{assumption_smooth};
(b) holds because $\EBB\langle \nabla_{\xB} f(\zB_t), \uB_t \rangle = \EBB\|\nabla_{\xB} f(\zB_t)\|^2$,
$\EBB\langle \nabla_{\yB} f(\zB_t), \vB_t \rangle = \EBB\|\nabla_{\yB} f(\zB_t)\|^2$;
(c) follows from Assumption~\ref{assumption_bgd} and the fact that $\|\aB + \bB\|^2 \le 2 \|\aB\|^2 + 2 \|\bB\|^2$;
(d) holds because $\gamma_t \le 1 / (2 L_f K)$;
(e) follows from Assumption~\ref{assumption_pl}.
Combining the above inequalities and Lemma~\ref{lemma_one_round_progress} leads to the desired result.
\end{proof}

As shown in Lemmas~\ref{lemma_local_iterate_error_naive} and~\ref{lemma_phi_f_diff_bound}, the upper bounds of $\EBB[\Phi(\xB_t) \!-\! f(\zB_t)]$ and $\EBB\|\eB'_{\xB, t}\|^2$ are complicated, and hence are difficult to be analyzed separately.
Thus, in our analysis, we construct the following potential function to handle them simultaneously,
\begin{equation}  \label{eq_lyapunov_naive}
\LM_t := \EBB \big[\Phi(\xB_t) - \Phi(\xB^*) + {\textstyle \frac{1}{20}} \big( \Phi(\xB_t) - f(\zB_t) \big) \big],
\end{equation}
where $\xB^* \in \mathrm{argmin}_{\xB \in \RBB^p} \Phi(\xB)$.
Note that $\LM_t \ge 0$ since $\Phi(\xB_t) \ge \Phi(\xB^*) \ge f(\xB_t, \yB_t)$, and $\LM_t = 0$ implies that $\xB_t$ is a minimizer of $\Phi$ and $\yB_t \in \mathrm{argmax}_{\yB \in \RBB^{q}} f(\xB_t, \yB)$ is an optimal dual variable.
By combining Lemmas~\ref{lemma_one_round_progress}-\ref{lemma_phi_f_diff_bound}, we obtain the following important recursive relation on $\LM_t$.
\begin{lemma}  \label{lemma_potential_bound_naive}
Suppose that Assumptions~\ref{assumption_bgd}-\ref{assumption_intra_client_variance} hold and that the step sizes $\eta_t, \gamma_t$ of {\AlgNaive} satisfy
$
\eta_t \le \frac{\gamma_t}{21 \kappa^2}
$
and
$
\gamma_t \le \frac{1}{87 L_f K}
$.
Besides, suppose that the conditions in Lemma~\ref{lemma_local_iterate_error_naive} hold.
Then
$$
\begin{aligned}
    \LM_{t+1}
    \le{}& \LM_t
        - {\textstyle \frac{21 \eta_t K}{40}} \EBB\|\nabla \Phi(\xB_t)\|^2 \\
        &\!+\! {\textstyle \frac{21 \eta_t^2 K^2}{20 S}} \big( L_{\Phi} \!+\! L_f \big) \sigma_1^2
        + \eta_t^2 K^3 L_f^2 \big( \eta_t \!+\! {\textstyle \frac{\gamma_t}{30}} \big) (\sigma_1^2 \!+\! G_1^2) \nonumber \\
        &\!+\! {\textstyle \frac{\gamma_t^2 K^2 L_f}{20 S}} \sigma_2^2
        \!+\! \gamma_t^2 K^2 L_f^2 \big( \eta_t \!+\! {\textstyle \frac{\gamma_t}{30}} \big) (\sigma_2^2 \!+\! G_2^2),
\end{aligned}
$$
\end{lemma}


\begin{proof} 
Recall that $\LM_t$ is defined as
$$
\LM_t = \EBB[\Phi(\xB_t) - \Phi(\xB^*) + \big( \Phi(\xB_t) - f(\xB_t, \yB_t) \big) / 20 ].
$$
Combining Lemmas~\ref{lemma_one_round_progress}, \ref{lemma_local_iterate_error_naive}, and~\ref{lemma_phi_f_diff_bound} and applying the condition $\eta_t \le \frac{\gamma_t}{21 \kappa^2} \le \min \left\{ \frac{1}{36 L_f K}, \frac{1}{4 L_{\Phi} K} \right\}$ and $\gamma_t \le \frac{1}{87 L_f K}$, we have
\begin{align}
    &\EBB[\Phi(\xB_{t+1}) - \Phi(\xB^*) + \frac{1}{20}\big( \Phi(\xB_{t+1}) - f(\zB_{t+1}) \big)] \nonumber \\
    \le{}& \EBB[\Phi(\xB_t) \!-\! \Phi(\xB^*)]
        \!+\! \frac{1}{20} \EBB[\Phi(\xB_t) \!-\! f(\zB_t)]
        \!-\! \frac{21 \hat{\eta}_t}{40} \EBB\|\nabla \Phi(\xB_t)\|^2 \nonumber \\
        &\!+\! \frac{21 \hat{\eta}_t^2}{20 S} \big( L_{\Phi} \!+\! L_f \big) \sigma_1^2
        + \big( \frac{37 \hat{\eta}_t^3L_f^2}{20} \!+\! \frac{\hat{\gamma}_t \hat{\eta}_t^2 L_f^2}{30} \big) (\sigma_1^2 \!+\! G_1^2) \nonumber \\
        &\!+\! \frac{\hat{\gamma}_t^2 L_f}{20 S} \sigma_2^2
        \!+\! \big( \frac{37 \hat{\eta}_t \hat{\gamma}_t^2 L_f^2}{20} \!+\! \frac{\gamma_t^3 K^3 L_f^2}{30} \big) (\sigma_2^2 \!+\! G_2^2),  \label{eq_lyapunov_expand_naive}
\end{align}
where $\hat{\eta} := \eta K$ and $\hat{\gamma} := \gamma K$.
This completes the proof.
\end{proof}

By telescoping Lemma~\ref{lemma_potential_bound_naive} from $t=1$ to $T-1$ and selecting appropriate step sizes, we obtain the convergence rate in Theorem~\ref{theorem_alg_naive}.
The proof is detailed as follows.
\begin{proof}[Proof of Theorem~\ref{theorem_alg_naive}]
Denote the step sizes as
$\gamma_t = \gamma := \min \Big\{
    \sqrt{\frac{20 \LM_0 S}{L_f T K^2 \sigma_2^2}},
    \Big( \frac{30 \LM_0}{L_f^2 (\sigma_2^2 + G_2^2) T K^3} \Big)^{1/3},
    \allowbreak
    \frac{1}{87 L_f K}
\Big\}
$
and
$
\eta_t = \eta := \min \Big\{
    \sqrt{\frac{20 \LM_0 S}{7 (L_{\Phi} + L_f) T K^2 \sigma_1^2}},
    \Big( \frac{3 \LM_0}{L_f^2 (\sigma_1^2 + G_1^2) T K^3} \Big)^{1/3},
    \frac{\gamma}{21 \kappa^2}
\Big\}.$
Recursively applying~\eqref{eq_lyapunov_expand_naive} yields
\begin{align*}
    &\EBB[\Phi(\xB_T) - \Phi(\xB^*)]
        + \frac{1}{20} \EBB[\Phi(\xB_T) - f(\zB_T)] \\
    \le{}& \LM_0
        \!-\! \frac{21 \hat{\eta}}{40} {\textstyle \sum_{t=0}^{T-1}} \EBB\|\nabla \Phi(\xB_t)\|^2
        \!+\! \frac{21 \hat{\eta}^2 T}{20 S} \big( L_{\Phi} \!+\! L_f \big) \sigma_1^2 \\
        &+ \frac{\hat{\gamma}^2 L_f T}{20 S} \sigma_2^2 
        \!+\! \big( \frac{37 \hat{\eta}^3 L_f^2}{20} + \frac{\hat{\gamma} \hat{\eta}^2 L_f^2}{30} \big) (\sigma_1^2 + G_1^2) T \\
        &+ \big( \frac{37 \hat{\eta} \hat{\gamma}^2 L_f^2}{20} + \frac{\hat{\gamma}^3 L_f^2}{30} \big) (\sigma_2^2 + G_2^2) T.
\end{align*}
Plugging the values of $\eta$ and $\gamma$ to the above inequality and rearranging terms, we arrive at
\begin{align*}
    & {\textstyle \frac{1}{T} \sum_{t=0}^{T-1}} \EBB\|\nabla \Phi(\xB_t)\|^2 \\
    \le{}& \frac{40 \LM_0}{21 \eta K T}
        + \frac{2 \eta K (L_{\Phi} + L_f)}{S} \sigma_1^2
        + \frac{2 \gamma^2 K L_f}{21 \eta S} \sigma_2^2 \\
        &+ \big( \frac{74 \eta^2 K^2 L_f^2}{21} + \frac{4 \gamma \eta K^2 L_f^2}{63} \big) (\sigma_1^2 + G_1^2) \\
        &+ \big( \frac{74 \gamma^2 K^2 L_f^2}{21} + \frac{4 \gamma^3 K^2 L_f^2}{63 \eta} \big) (\sigma_2^2 + G_2^2) \\
    \le{}& \OM \Big(
        \frac{\kappa^2 \LM_0 L_f}{T}
        \!+\! \sigma_1 \sqrt{\frac{\kappa \LM_0 L_f}{S T}}
        \!+\! \kappa^2 \sigma_2 \sqrt{\frac{\LM_0 L_f}{S T}}
        \!+\! \Big(
            (\sigma_1^2 \!+\! G_1^2)^{1/3} \\
            &\!+\! \frac{(\sigma_1^2 \!+\! G_1^2)^{2/3}}{(\sigma_2^2 + G_2^2)^{1/3}}
            \!+\! \kappa^2 (\sigma_2^2 \!+\! G_2^2)^{1/3}
            \Big) \Big( \frac{\LM_0 L_f}{T} \Big)^{2/3}
        \Big),
\end{align*}
which is the desired result.
\end{proof}

\subsection{Proofs of the Results in Section~\ref{section_analysis_mb}}
Define $\eB_{\xB, t}$ and $\eB_{\yB, t}$ as
$$
\begin{cases}
    \eB_{\xB, t}
    := \frac{1}{|\SM_t| K} \! \sum_{i \in \SM_t} \! \sum_{k=0}^{K-1} \big(
        \tilde{\nabla}_{\xB}^{(k)} F_i(\zB_{t, i}^{(k)})
        \!-\! \tilde{\nabla}_{\xB}^{(k)} F_i(\zB_t)
        \big), \\
    \eB_{\yB, t}
    := \frac{1}{|\SM_t| K} \! \sum_{i \in \SM_t} \! \sum_{k=0}^{K-1} \big(
        \tilde{\nabla}_{\yB}^{(k)} F_i(\zB_{t, i}^{(k)})
        \!-\! \tilde{\nabla}_{\yB}^{(k)} F_i(\zB_t)
        \big).
\end{cases}
$$
We first provide the following lemma.
\begin{lemma}  \label{lemma_local_iterate_error_mb}
Suppose that $\eta_t, \gamma_t \le \frac{1}{2 K L_f}$ in {\AlgMB}.
Then, both $\EBB\|\eB_{\xB, t}\|^2$ and $\EBB\|\eB_{\yB, t}\|^2$ are bounded from above by
$
    18 L_f^2 \Big( \hat{\eta}_t^2 \EBB \| \nabla_{\xB} f(\zB_t) \|^2
        \!+\! \hat{\gamma}_t^2 \EBB \| \nabla_{\yB} f(\zB_t) \|^2
        \!+\! \frac{\hat{\eta}_t^2 \sigma_1^2}{S}
        \!+\! \frac{\hat{\gamma}_t^2 \sigma_2^2}{S} \Big).
$
\end{lemma}

\begin{proof}
We observe that
$$
\resizebox{\linewidth}{!}{$
\begin{aligned}[b]
    &\EBB\|\eB_{\xB, t}\|^2
    = \EBB \Big\| \frac{1}{|\SM_t| K} \! \sum_{i \in \SM_t} \! \sum_{k=0}^{K-1} \big( \tilde{\nabla}_{\xB}^{(k)} F_i(\zB_{t, i}^{(k)}) \!-\! \tilde{\nabla}_{\xB}^{(k)} F_i(\zB_t) \big) \Big\|^2 \\
    \le& \frac{1}{|\SM_t| K} \! \sum_{i, k} \! \EBB \Big\| \tilde{\nabla}_{\xB}^{(k)} F_i(\zB_{t, i}^{(k)}) \!-\! \tilde{\nabla}_{\xB}^{(k)} F_i(\zB_t) \Big\|^2
    \!\le\! \frac{L_f^2}{|\SM_t| K} \! \sum_{i, k} \! \EBB \| \zB_{t, i}^{(k)} \!-\! \zB_t \|^2,
\end{aligned}
$}
$$
where the first inequality follows from Cauchy-Schwarz and the last inequality follows from Assumption~\ref{assumption_smooth}.
Similarly, we have $\EBB[\|\eB_{\yB, t}\|^2] \le \frac{L_f^2}{|\SM_t| K} \sum_{i, k} \EBB[ \| \zB_{t, i}^{(k)} - \zB_t \|^2]$.

If $K = 1$, then $\EBB\| \zB_{t, i}^{(1)} - \zB_t \|^2 = \eta_t^2 \EBB\|\uB_t\|^2 + \gamma_t^2 \EBB\|\vB_t\|^2$.
Suppose that $K \ge 2$, then for any $k \in \{0, \ldots, K-1\}$,
\begin{align*}
    &\EBB\| \zB_{t, i}^{(k+1)} \!-\! \zB_t \|^2 \\
    =& \EBB\| \xB_{t, i}^{(k)} \!-\! \xB_t - \eta_t \big( \tilde{\nabla}_{\xB}^{(k)} F_i(\zB_{t, i}^{(k)}) \!-\! \tilde{\nabla}_{\xB}^{(k)} F_i(\zB_t) \!+\! \uB_t \big) \|^2 \\
        &+ \EBB\| \yB_{t, i}^{(k)} \!-\! \yB_t \!+\! \gamma_t \big( \tilde{\nabla}_{\yB}^{(k)} F_i(\zB_{t, i}^{(k)}) - \tilde{\nabla}_{\yB}^{(k)} F_i(\zB_t) + \vB_t \big) \|^2 \\
    \le& {\textstyle \frac{K}{K \!-\! 1}} \EBB\| \zB_{t, i}^{(k)} \!-\! \zB_t \|^2
        + K \eta_t^2 \EBB\| \tilde{\nabla}_{\xB}^{(k)} F_i(\zB_{t, i}^{(k)}) - \tilde{\nabla}_{\xB}^{(k)} F_i(\zB_t) \\
            &\!+\! \uB_t \|^2
        \!+\! K \gamma_t^2 \EBB\| \nabla_{\yB} F_i(\zB_{t, i}^{(k)}; \BM_{t, i}^{(k)}) - \nabla_{\yB} F_i(\zB_t; \BM_{t, i}^{(k)}) + \vB_t \|^2 \\
    \le& {\textstyle \frac{K}{K \!-\! 1}} \EBB\| \zB_{t, i}^{(k)} \!-\! \zB_t \|^2
        \!+\! 2 K \eta_t^2 \EBB\|\uB_t\|^2 \!+\! 2 K \gamma_t^2 \EBB\|\vB_t\|^2 \\
        &\!+\! \EBB\| \tilde{\nabla}_{\xB}^{(k)} F_i(\zB_{t, i}^{(k)}) - \tilde{\nabla}_{\xB}^{(k)} F_i(\zB_t) \|^2 / (2 K L_f^2) \\
        &\!+\! \EBB\| \tilde{\nabla}_{\yB}^{(k)} F_i(\zB_{t, i}^{(k)}) \!-\! \tilde{\nabla}_{\yB}^{(k)} F_i(\zB_t) \|^2 / (2 K L_f^2) \\
    \le{}& {\textstyle \frac{K \!+\! 1}{K \!-\! 1}} \EBB\| \zB_{t, i}^{(k)} - \zB_t \|^2
        + 2 K \eta_t^2 \EBB\|\uB_t\|^2 \!+\! 2 K \gamma_t^2 \EBB\|\vB_t\|^2,
\end{align*}
where the first inequality follows from Young's inequality, the second inequality follows from the condition $\eta_t, \gamma_t \le 1/(2 L_f K)$, and the last inequality follows from Assumption~\ref{assumption_smooth}.
Recursively applying the above inequality yields
\begin{align*}
    &\EBB\| \zB_{t, i}^{(k)} - \zB_t \|^2 \\
    \le& 2 K (\eta_t^2 \EBB\|\uB_t\|^2 \!+\! \gamma_t^2 \EBB\|\vB_t\|^2)  {\textstyle \sum_{k'=0}^{k-1} (\frac{K + 1}{K - 1})^{k'}} \\
    \le& 18 \hat{\eta}_t^2 \EBB\|\uB_t\|^2
        \!+\! 18 \hat{\gamma}_t^2 \EBB\|\vB_t\|^2 \\
    \le& 18 \hat{\eta}_t^2 (\EBB\|\nabla_{\xB} f(\zB_t)\|^2
        \!+\! \frac{\sigma_1^2}{S})
        \!+\! 18 \hat{\gamma}_t^2 (\EBB\|\nabla_{\yB} f(\zB_t)\|^2
            + \frac{\sigma_2^2}{S}),
\end{align*}
where the last inequality follows from Assumption~\ref{assumption_bgd} and the condition $|\SM_t'| = S_t \ge S$.
\end{proof}

Now, we are ready to prove Theorem~\ref{theorem_alg_mb}.
\begin{proof}[Proof of Theorem~\ref{theorem_alg_mb}]
The proof is similar to that of Theorem~\ref{theorem_alg_naive}.
We note that Lemma~\ref{lemma_one_round_progress} and Lemma~\ref{lemma_phi_f_diff_bound} also apply to {\AlgMB} with $\eB_{\xB, t}'$ and $\eB_{\yB, t}'$ replaced by $\eB_{\xB, t}$ and $\eB_{\yB, t}$, respectively.
Following the same argument as~\eqref{eq_lyapunov_expand_naive}, we have
\begin{align}  \label{eq_lyapunov_expand_mb}
    \LM_{t+1}
    \le{}& \LM_t
        - {\textstyle \frac{21 \eta_t K}{40}} \EBB\|\nabla \Phi(\xB_t)\|^2 \nonumber \\
        &+ {\textstyle \frac{\eta_t^2 K^2}{S}} \big( 1.05 L_{\Phi} + 0.81 L_f \big) \sigma_1^2
        + {\textstyle \frac{3 \gamma_t^2 K^2 L_f}{4 S}} \sigma_2^2,
\end{align}
where $\LM_t$ is defined in~\eqref{eq_lyapunov_naive}.
We denote the step sizes as
$ \gamma_t = \gamma := \min \left\{ \frac{1}{87 L_f K}, \sqrt{\frac{4 \LM_0 S}{3 L_f T K^2 \sigma_2^2}} \right\}
$ and
$
\eta_t = \eta := \min \left\{
    \frac{\gamma}{21 \kappa^2},
    \sqrt{\frac{40 \LM_0 S}{21 (L_{\Phi} + L_f) T K^2 \sigma_1^2}}\right\}
$.
Recursively applying the inequality~\eqref{eq_lyapunov_expand_mb} yields
$$
\begin{aligned}[b]
    \LM_T
    \le{}& \LM_0
        - \frac{21 \hat{\eta}}{40} {\textstyle \sum_{t=0}^{T-1}} \EBB\|\nabla \Phi(\xB)\|^2
        + \frac{3 T \hat{\gamma}^2 L_f}{4 S} \sigma_2^2 \\
        &+ \frac{21 T \hat{\eta}^2 (L_{\Phi} + L_f)}{20 S} \sigma_1^2,
\end{aligned}
$$
where $\hat{\eta} := \eta K$ and $\hat{\gamma} := \gamma K$.
Plugging the values of $\eta$ and $\gamma$ into the above inequality and rearranging terms, we arrive at
\begin{align*}
    & {\textstyle \frac{1}{T} \sum_{t=0}^{T-1}} \EBB[\|\nabla \Phi(\xB_t)\|^2] \\
    \le{}& \frac{40}{21 \hat{\eta} T} \LM_0
        + \frac{2 \hat{\eta} (L_{\Phi} + L_f)}{S} \sigma_1^2 + \frac{10 \hat{\gamma}^2 L_f}{7 \hat{\eta} S} \sigma_2^2 \\
    \le{}& 4 \sqrt{\frac{40 \LM_0 (L_{\Phi} + L_f) \sigma_1^2}{21 S T}}
        + \left( \frac{20}{7} + 60 \kappa^2 \right) \sqrt{\frac{\LM_0 L_f \sigma_2^2}{3 S T}} \\
        &+ \left( 8 L_{\Phi} + (1827 \kappa^2 + 72) L_f \right) \frac{40 \LM_0}{21 T} \\
    ={}& \OM \Big( \sqrt{\frac{ \kappa \LM_0 L_f \sigma_1^2}{S T}}
        \!+\! \kappa^2 \sqrt{\frac{\LM_0 L_f \sigma_2^2}{S T}}
        \!+\! \frac{\kappa^2 \LM_0 L_f}{T} \Big),
\end{align*}
which is the desired result.
\end{proof}

\subsection{Proofs of the Results in Section~\ref{section_analysis_storm}}   \label{section_analysis_storm_appendix}
We first present the following theorem, which corresponds to the convergence rate of {\AlgSTORM} under the additional $\delta$-BHD condition.
\begin{theorem}  \label{theorem_alg_storm}
    Define $\alpha_t$, $\eta_t$, and $\gamma_t$ in {\AlgSTORM} as
    $\alpha_t = 200000 \hat{\delta}^2 K^2 \gamma_t^2$,
    $\eta_t = \frac{\gamma_t}{12 \kappa^2}$,
    and $\gamma_t = \min\{\frac{1}{2 L_f}, \frac{\num{5.78e-4}}{\hat{\delta} K}, \frac{S^{1/3} \hat{\LM}_0^{1/3}}{\hat{\delta}^{2/3} K (t + 1)^{1/3} (\sigma_1^2 / \kappa^2 + \sigma_2^2)^{1/3}}\}$,
    where $\hat{\LM}_0$ is defined in Theorem~\ref{theorem_alg_storm_no_delta} and $\hat{\delta} := \max\{\delta, \mu\}$.
    In addition, define $c := \hat{\delta} / \mu = \max\{1, \delta / \mu\}$.
    Suppose that $S \ge (\sigma_1^2 / \kappa^2 + \sigma_2^2) / (\hat{\delta} \hat{\LM}_0)$ and Assumptions~\ref{assumption_bgd}-\ref{assumption_pl} and~\ref{assumption_bhd} hold.
    Then,
    \begin{align*}
    &{\textstyle \frac{1}{K T} \sum_{t=0}^{T-1} {\textstyle \frac{1}{|\SM_t|}} \sum_{k=0}^{K-1} \sum_{i \in \SM_t}} \EBB[\|\nabla \Phi(\xB_{t, i}^{(k)})\|^2] \\
        =& \tilde{\OM} \Big( 
            \frac{\kappa^2 \hat{\LM}_0 L_f}{T} (\frac{1}{K} + \frac{c}{\kappa})
            + \frac{\kappa^{4/3} (L_f c)^{2/3} \hat{\LM}_0^{2/3} (\sigma_1^2 / \kappa^2 + \sigma_2^2)^{1/3}}{S^{1/3} T^{2/3}} \\
            &+ \frac{\kappa^2 (\sigma_1^2 / \kappa^2 + \sigma_2^2)}{S} (\frac{\kappa / (c K) + 1}{T^{2/3}} + \frac{\kappa^2 / (c^2 K^2) + 1}{T})
        \Big).
    \end{align*}
When $\hat{\LM}$, $L_f$, $\sigma_1$, and $\sigma_2$ are treated as constants, {\AlgSTORM} has a communication complexity of
\begin{equation} 
    \tilde{\OM} (\frac{\kappa }{\varepsilon^2} (\frac{\kappa}{K} + c) + (1 + \frac{\kappa}{c K})^{3} \frac{c \kappa^2}{S^{1/2} \varepsilon^3}).
\end{equation}
\end{theorem}
%
Note that Theorem~\ref{theorem_alg_storm_no_delta}, which does not rely on the $\delta$-BHD condition, is actually a direct corollary of Theorem~\ref{theorem_alg_storm} as Assumption~\ref{assumption_bhd} immediately holds with $\delta = L_f$ under Assumption~\ref{assumption_smooth}.
For ease of notation, we omit the subscript $t, i$ of the iterates in {\AlgSTORM} and denote $\xB_{t, i}^{(k)}$ (resp., $\yB_{t, i}^{(k)}$) as $\xB^{(k)}$ (resp., $\yB^{(k)}$).
Similarly, we denote $\dB_{\xB, t, i}^{(k)}$ (resp., $\dB_{\yB, t, i}^{(k)}$) defined in~\eqref{eq_local_update_direction} as $\dB_{\xB}^{(k)}$ (resp., $\dB_{\yB}^{(k)}$) when it is clear from context.
To prove Theorem~\ref{theorem_alg_storm}, we first define some auxiliary quantities and establish several lemmas.
We denote the gradient estimation errors in {\AlgSTORM} as
$$
\begin{aligned}[b]
\begin{cases}
    \varepsilon_{\xB, t} := \EBB\|\uB_t - \nabla_{\xB} f(\xB_{t}, \yB_{t})\|^2, \\
    \varepsilon_{\yB, t} := \EBB\|\vB_t - \nabla_{\yB} f(\xB_{t}, \yB_{t})\|^2,
\end{cases}
\end{aligned}
$$
where the expectation is taken over all randomness. 
These two quantities can be bounded as follows.

\begin{lemma} \label{lemma_gradient_error_storm}
$\varepsilon_{\xB, t+1}$ and $\varepsilon_{\yB, t+1}$ are bounded by
$$
\begin{aligned}[b]
\begin{cases}
    \varepsilon_{\xB, t+1}
    \le (1 \!-\! \alpha_{t+1})^2 (\varepsilon_{\xB, t} + 8 \delta^2 \EBB \|\zB_{t+1} \!-\! \zB_{t}\|^2)
        \!+\! \frac{2 \alpha_{t+1}^2 \sigma_1^2}{|\SM_{t+1}|}, \\
    \varepsilon_{\yB, t+1}
    \le (1 \!-\! \alpha_{t+1})^2 (\varepsilon_{\yB, t} + 8 \delta^2 \EBB \|\zB_{t+1} \!-\! \zB_{t}\|^2)
        + \frac{2 \alpha_{t+1}^2 \sigma_2^2}{|\SM_{t+1}|}.
\end{cases}
\end{aligned}
$$
\end{lemma}

\begin{proof}
First, we show that the function $F_i(\xB, \yB; \zeta_{i, j}) - f(\xB, \yB)$ is $2 \delta$-Lipschitz smooth.
For some $\zB, \zB' \in \RBB^{p \times q}$ and $\tau \in [0, 1]$, denote $\zB_{\tau} = (1 \!-\! \tau) \zB \!+\! \tau \zB'$.
By Assumption~\ref{assumption_bhd},
\begin{align*}
    &\|\nabla F_i(\zB'; \zeta_{i, j}) \!-\! \nabla f(\zB') \!-\! \left(\nabla F_i(\zB; \zeta_{i, j}) \!-\! \nabla f(\zB) \right) \| \\
    =& \Big\| \int_{0}^{1} \! \big( \nabla^2 F_i(\zB_{\tau}; \zeta_{i, j}) \!-\! \nabla^2 f(\zB_{\tau}) \big) (\zB' \!-\! \zB) d \tau \Big\| \\
    \le& \int_{0}^{1} \Big\| \nabla^2 F_i(\zB_{\tau}; \zeta_{i, j}) - \nabla^2 f(\zB_{\tau}) \Big\| \cdot \| \zB' - \zB \| d \tau
    \le 2 \delta \| \zB' - \zB \|,
\end{align*}
where the last inequality follows from Assumption~\ref{assumption_bhd}.
Similarly,
\begin{equation}  \label{eq_bounded_gradient_diff}
    \|\nabla F_i(\zB'; \BM) \!-\! \nabla f(\zB') \!-\! \nabla F_i(\zB; \BM) \!+\! \nabla f(\zB)\|
    \!\le\! 2 \delta \| \zB' \!-\! \zB \|,
\end{equation}
where $\BM$ is a minibatch of data samples on client $i$ and $F_i(\zB; \BM) := |\BM|^{-1} \sum_{\zeta \in \BM} F_i(\zB; \zeta)$.
Thus, the function $F_i(\xB, \yB; \BM) - f(\xB, \yB)$ is $2 \delta$-Lipschitz smooth w.r.t.\ $(\xB, \yB)$ for any client $i$ and any minibatch $\BM$ of data samples on client $i$.

By the definition of $\varepsilon_{\xB, t+1}$, we have
\begin{align*}
    &\varepsilon_{\xB, t+1} \\
    =& \EBB\|\uB_{t+1} \!-\! \nabla_{\xB} f(\zB_{t+1})\|^2 \\
    =& \EBB \Big\|(1 \!-\! \alpha_{t+1}) \big(\uB_t \!-\! \nabla_{\xB} f(\zB_{t}) \big)
        \!-\! \alpha_{t+1} \nabla_{\xB} f(\zB_{t+1}) \\
            &\!+\! {\textstyle \frac{\alpha_{t+1}}{|\SM_{t+1}'|} \sum_{i \in \SM_{t+1}'} \nabla_{\xB} f_i(\zB_{t+1})}
        \!+\! (1 \!-\! \alpha_{t+1}) \big(
            \nabla_{\xB} f(\zB_{t}) \\
            &\!-\! \nabla_{\xB} f(\zB_{t+1})
            \!+\! {\textstyle \frac{1}{|\SM_{t+1}'|} \sum_{i \in \SM_{t+1}'}} (\nabla_{\xB} f_i(\zB_{t+1}) \!-\! \nabla_{\xB} f_i(\zB_{t})) \big)
        \Big\|^2 \\
    \overset{(a)}{=}& (1 \!-\! \alpha_{t+1})^2 \varepsilon_{\xB, t}
        \!+\! \EBB \Big\|  {\textstyle \frac{\alpha_{t+1}}{|\SM_{t+1}'|} \sum_{i \in \SM_{t+1}'}} \nabla_{\xB} f_i(\zB_{t+1}) \\
            &\!-\! \alpha_{t+1} \nabla_{\xB} f(\zB_{t+1})
        \!+\! (1 \!-\! \alpha_{t+1}) \big(
            \nabla_{\xB} f(\zB_{t})
            \!-\! \nabla_{\xB} f(\zB_{t+1}) \\
            &\!+\! {\textstyle \frac{1}{|\SM_{t+1}'|} \sum_{i \in \SM_{t+1}'}} (\nabla_{\xB} f_i(\zB_{t+1}) 
                \!-\! \nabla_{\xB} f_i(\zB_{t}))
            \big)
        \Big\|^2 \\
     \le& (1 \!-\! \alpha_{t+1})^2 \varepsilon_{\xB, t}
        \!+\! 2 \alpha_{t+1}^2 \EBB \|
            {\textstyle \frac{1}{|\SM_{t+1}'|} \sum_{i \in \SM_{t+1}'}} \nabla_{\xB} f_i(\zB_{t+1}) \\
            &\!-\! \nabla_{\xB} f(\zB_{t+1}) \|^2
        \!+\! 2 (1 \!-\! \alpha_{t+1})^2 \EBB \|
            \nabla_{\xB} f(\zB_{t})
            \!-\! \nabla_{\xB} f(\zB_{t+1}) \\
            &\!+\! {\textstyle \frac{1}{|\SM_{t+1}'|} \sum_{i \in \SM_{t+1}'}} (\nabla_{\xB} f_i(\zB_{t+1}) 
            \!-\! \nabla_{\xB} f_i(\zB_{t}))
        \|^2 \\
    \overset{(b)}{\le}& (1 \!-\! \alpha_{t \!+\! 1})^2 \varepsilon_{\xB, t}
        + 8 (1 \!-\! \alpha_{t \!+\! 1})^2 \delta^2 \EBB\|\zB_{t+1} \!-\! \zB_{t}\|^2
        + {\textstyle \frac{2 \alpha_{t \!+\! 1}^2 \sigma_1^2}{|\SM_{t+1}|}},
\end{align*}
where (a) holds because $\EBB_{\SM_{t+1}'} [ \frac{1}{|\SM_{t+1}'|} \sum_{i \in \SM_{t+1}'} f_i(\zB)] = f(\zB)$ and $\SM_{t+1}'$ is independent of $\varepsilon_{\xB, t}$;
(b) follows from Assumptions~\ref{assumption_bgd} and~\eqref{eq_bounded_gradient_diff}.
This proves the first part of Lemma~\ref{lemma_gradient_error_storm}.
The second part of the lemma follows from the same argument.
\end{proof}

The next lemma shows that the expected loss value $\EBB[\Phi(\xB^{(k)})]$ decreases as $k$ increases as long as the errors $\EBB[\Phi(\xB^{(k)}) - f(\zB^{(k)})]$, $\EBB[\|\zB^{(k)} - \zB_t\|^2]$, and $\varepsilon_{\xB, t}$ are controlled in a desirable level. 
\begin{lemma}  \label{lemma_descent}
Suppose that $\eta_t \le \frac{1}{2 L_{\Phi}}$.
Under Assumptions~\ref{assumption_bgd}-\ref{assumption_pl}, $\forall k \in \{0, \ldots, K - 1\}$, we have
\begin{align*}
    &\EBB[ \Phi(\xB^{(k+1)}) - \Phi(\xB^{(k)})] \\
    \le& \!-\!  \frac{\eta_t}{2} \EBB \| \nabla \Phi(\xB^{(k)})\|^2
        \!+\! \frac{3 \Lot^2}{4 \mu}  \eta_t \EBB[ \Phi(\xB^{(k)}) - f(\zB^{(k)})]
        \!+\! \frac{7 \eta_t}{4} \varepsilon_{\xB, t} \\
        &\!+\! 7 \delta^2 \eta_t \EBB \| \zB^{(k)} \!-\! \zB_{t}\|^2
        \!-\! \frac{\eta_t}{16} \EBB\| \nabla_{\xB} f(\zB^{(k)})\|^2
        \!-\! \frac{\eta_t}{8} \EBB\|\dB_{\xB}^{(k)}\|^2.
\end{align*}
\end{lemma}

\begin{proof}
By Assumption~\ref{assumption_smooth}, both $F_i(\cdot, \cdot; \BM^{(k)})$ and $f(\cdot, \cdot)$ are $L_f$-smooth.
Thus, we can bound
\begin{align}
    &\EBB\| \dB_{\xB}^{(k)} \!-\! \nabla \Phi(\xB^{(k)}) \|^2 \nonumber \\
    =& \EBB\| \tilde{\nabla}_{\xB}^{(k)} F_i(\zB^{(k)})
        - \tilde{\nabla}_{\xB}^{(k)} F_i(\zB_{t})
        - \nabla_{\xB} f(\zB^{(k)}) \nonumber \\
        &+ \nabla_{\xB} f(\zB_{t})
        - \nabla_{\xB} f(\zB_{t})
        + \uB_t
        + \nabla_{\xB} f(\zB^{(k)}) - \nabla \Phi(\xB^{(k)}) \|^2 \nonumber \\
    \le& 3 \EBB \|\tilde{\nabla}_{\xB}^{(k)} F_i(\zB^{(k)}) \!-\! \tilde{\nabla}_{\xB}^{(k)} F_i(\zB_{t})
        \!-\! \nabla_{\xB} f(\zB^{(k)}) + \nabla_{\xB} f(\zB_{t})\|^2 \nonumber \\
        &\!+\! 3 \varepsilon_{\xB, t}
        \!+\! 3 \EBB \|\nabla_{\xB} f(\zB^{(k)}) \!-\! \nabla \Phi(\xB^{(k)})\|^2 \nonumber \\
    \le& 12 \delta^2 \EBB \|\zB^{(k)} \!-\! \zB_{t}\|^2
        \!+\! 3 \varepsilon_{\xB, t}
        \!+\! 3 \EBB \|\nabla_{\xB} f(\zB^{(k)}) \!-\! \nabla \Phi(\xB^{(k)})\|^2 \nonumber \\
    \le& 12 \delta^2 \EBB \|\zB^{(k)} \!-\! \zB_{t}\|^2
        \!+\! 3 \varepsilon_{\xB, t}
        \!+\! 3 \Lot^2 \EBB \|\yB^{(k)} \!-\! \yB^*(\xB^{(k)})\|^2 \nonumber \\
    \!\le& 12 \delta^2 \EBB \|\zB^{(k)} \!-\! \zB_{t} \! \|^2
        \!+\! 3 \varepsilon_{\xB, t}
        \!+\! {\textstyle \frac{3 \Lot^2}{2 \mu}} \! \EBB[ \Phi(\xB^{(k)}) \!-\! f(\zB^{(k)})],  \label{eq_gradient_similarity_phi}
\end{align}
where the second, third, and last inequalities follow from~\eqref{eq_bounded_gradient_diff}, Assumption~\ref{assumption_smooth},
and~\eqref{eq_pl_property}, respectively.
Similarly, we have
\begin{equation}  \label{eq_gradient_similarity_f}
\begin{cases}
\begin{aligned}
    \EBB \|\dB_{\xB}^{(k)} \!-\! \nabla_{\xB} f(\zB^{(k)})\|^2
    \le& 8 \delta^2 \EBB \|\zB^{(k)} \!-\! \zB_{t}\|^2
        \!+\! 2 \varepsilon_{\xB, t}, \\
    \EBB \|\dB_{\yB}^{(k)} \!-\! \nabla_{\yB} f(\zB^{(k)})\|^2
    \le& 8 \delta^2 \EBB \|\zB^{(k)} \!-\! \zB_{t}\|^2
        \!+\! 2 \varepsilon_{\yB, t}.
\end{aligned}
\end{cases}
\end{equation}

As $\Phi$ has $L_{\Phi}$-Lipschitz continuous gradients~\cite[Lemma 22]{nouiehed2019solving}, we have
\begin{align*}
    &\EBB[\Phi(\xB^{(k + 1)}) - \Phi(\xB^{(k)})] \nonumber \\
    \le& \EBB[\langle \nabla \Phi(\xB^{(k)}), \xB^{(k + 1)} - \xB^{(k)} \rangle
        + \frac{L_{\Phi}}{2} \|\xB^{(k+1)} - \xB^{(k)}\|^2] \nonumber \\
    =& - \eta_t \EBB \langle \nabla \Phi(\xB^{(k)}), \dB_{\xB}^{(k)} \rangle
        + \frac{L_{\Phi} \eta_t^2}{2} \EBB \| \dB_{\xB}^{(k)}\|^2 \nonumber \\
    =& \frac{\eta_t}{2} \! \EBB \Big[
        \! \| \dB_{\xB}^{(k)} \!-\! \nabla \Phi(\xB^{(k)})\|^2
        \!-\!  \|\nabla \Phi(\xB^{(k)})\|^2
        \!-\! (1 \!-\! L_{\Phi} \eta_t) \|\dB_{\xB}^{(k)}\|^2
        \! \Big] \nonumber \\
    \le& \frac{\eta_t}{2} \EBB \| \dB_{\xB}^{(k)} - \nabla \Phi(\xB^{(k)})\|^2
        - \frac{\eta_t}{2} \EBB \|\nabla \Phi(\xB^{(k)})\|^2
        - \frac{\eta_t}{4} \EBB\|\dB_{\xB}^{(k)}\|^2 \nonumber \\
    \le& \frac{\eta_t}{2} \EBB \| \dB_{\xB}^{(k)} \!-\! \nabla \Phi(\xB^{(k)})\|^2
        \!-\! \frac{\eta_t}{2} \EBB \|\nabla \Phi(\xB^{(k)})\|^2
        \!-\! \frac{\eta_t}{8} \EBB\|\dB_{\xB}^{(k)}\|^2 \nonumber \\
        &\!-\! \frac{\eta_t}{16} \EBB\| \nabla_{\xB} f(\xB^{(k)}, \yB^{(k)})\|^2
        \!+\! \frac{\eta_t}{8} \EBB\|\dB_{\xB}^{(k)} \!-\! \nabla_{\xB} f(\xB^{(k)}, \yB^{(k)})\|^2,
\end{align*}
where the first equality follows from the update rule $\xB^{(k+1)} = \xB^{(k)} - \eta_t \dB_{\xB}^{(k)}$
and the second equality follows from the fact that $-2 \langle \aB, \bB \rangle = \|\aB - \bB\|^2 - \|\aB\|^2 - \|\bB\|^2$.
Combining the above inequality, \eqref{eq_gradient_similarity_phi}, and~\eqref{eq_gradient_similarity_f} completes the proof.
\end{proof}

We provide upper bounds for the error terms $\EBB\|\zB^{(k)} - \zB_t\|^2$ and $\EBB[\Phi(\xB^{(k)}) - f(\zB^{(k)})]$ in the next two lemmas.
\begin{lemma}  \label{lemma_iterates_recursion}
Under Assumptions~\ref{assumption_bgd}-\ref{assumption_pl}, for any $k \in \{0, \ldots, K \!-\! 1\}$,
$$
\begin{aligned}[b]
\begin{cases}
    \EBB \|\xB^{(k+1)} \!-\! \xB_{t}\|^2
    \le {\textstyle \frac{K \!+\! 1}{K}} \EBB \|\xB^{(k)} \!-\! \xB_{t}\|^2
        \!+\! (1 \!+\! K) \eta_t^2 \EBB \|\dB_{\xB}^{(k)}\|^2, \\
    \EBB \|\yB^{(k+1)} \!-\! \yB_{t}\|^2
    \le {\textstyle \frac{K \!+\! 1}{K}} \EBB \|\yB^{(k)} \!-\! \yB_{t}\|^2
        \!+\! (1 \!+\! K) \gamma_t^2 \EBB \|\dB_{\yB}^{(k)}\|^2.
\end{cases}
\end{aligned}
$$
\end{lemma}

\begin{proof}
The claimed inequalities simply follow from the update rule of $(\xB^{(k)}, \yB^{(k)})$ and Young's inequality.
\end{proof}

\begin{lemma}  \label{lemma_max_residue}
Suppose that $\eta_t \le \min\{\frac{1}{2 L_f}, \frac{1}{2 L_{\Phi}}\}$ and $\gamma_t \le \frac{1}{2 L_f}$.
Under Assumptions~\ref{assumption_bgd}-\ref{assumption_pl}, for any $k \in \{0, \ldots, K \!-\! 1\}$, we have
\begin{align*}
    &\EBB[ \Phi(\xB^{(k+1)}) - f(\zB^{(k+1)})] \\
    \le& (1 \!-\! \frac{\mu \gamma_t}{2} \!+\! \frac{3 \Lot^2}{4 \mu} \eta_t) \EBB[ \Phi(\xB^{(k)}) - f(\zB^{(k)})]
        \!-\! \frac{\eta_t}{2} \EBB\|\nabla \Phi(\xB^{(k)})\|^2 \\
        &\!+\! (7 \eta_t \!+\! 4 \gamma_t) \delta^2 \EBB \|\zB^{(k)} \!-\! \zB_t\|^2
        \!+\! \frac{7 \eta_t}{16} \EBB \| \nabla_{\xB} f(\zB^{(k)})\|^2
        \!+\! \gamma_t \varepsilon_{\yB, t} \\
        &\!+\! \frac{7 \eta_t}{4} \varepsilon_{\xB, t}
        \!+\! \frac{5 \eta_t}{8} \EBB \| \dB_{\xB}^{(k)} \! \|^2
        \!-\! \frac{\gamma_t}{4} \EBB \| \dB_{\yB}^{(k)} \! \|^2
        \!-\! \frac{\gamma_t}{4} \EBB \| \nabla_{\yB} f(\zB^{(k)}) \! \|^2.
\end{align*}
\end{lemma}

\begin{proof}
First, we notice that
$$
\begin{aligned}
    \EBB[\Phi(\xB^{(k+1)}) \!-\! f(\zB^{(k+1)})]
    =& \EBB[ \Phi(\xB^{(k+1)}) \!-\! \Phi(\xB^{(k)})
        \!+\! \Phi(\xB^{(k)}) \\
        &\!-\! f(\zB^{(k)})
        \!+\! f(\zB^{(k)}) \!-\! f(\zB^{(k+1)})].
\end{aligned}
$$
In what follows, we bound $\EBB[f(\zB^{(k)}) - f(\zB^{(k+1)})]$.
\begin{align*}
    &\EBB[f(\zB^{(k)}) - f(\zB^{(k+1)})] \\
    \le& \EBB \big[ \langle \nabla_{\xB} f(\zB^{(k)}), \xB^{(k)} \!-\! \xB^{(k+1)} \rangle
        + \langle \nabla_{\yB} f(\zB^{(k)}), \yB^{(k)} \!-\! \yB^{(k+1)} \rangle \\
        &+ \frac{L_f}{2} \big( \|\xB^{(k)} \!-\! \xB^{(k+1)}\|^2 \!+\! \|\yB^{(k)} \!-\! \yB^{(k+1)}\|^2 \big)
        \big] \\
    =& \EBB \big[
        \eta_t \langle \nabla_{\xB} f(\zB^{(k)}), \dB_{\xB}^{(k)} \rangle
        - \gamma_t \langle \nabla_{\yB} f(\zB^{(k)}), \dB_{\yB}^{(k)} \rangle \\
        &+ \frac{L_f \eta_t^2}{2} \|\dB_{\xB}^{(k)}\|^2
        + \frac{L_f \gamma_t^2}{2} \|\dB_{\yB}^{(k)}\|^2
        \big] \\
    \le& \EBB \Big[
        \frac{\eta_t}{2} \! \|\nabla_{\xB} f(\zB^{(k)}) \! \|^2
        \!+\! \frac{\eta_t (1 \!+\! \eta_t L_f)}{2} \! \|\dB_{\xB}^{(k)} \! \|^2
        \!-\! \frac{\gamma_t}{2} \! \|\nabla_{\yB} f(\zB^{(k)}) \! \|^2 \\
        &\!-\! \big( \frac{\gamma_t}{2} \!-\! \frac{L_f \gamma_t^2}{2} \big) \! \|\dB_{\yB}^{(k)} \! \|^2
        \!+\! \frac{\gamma_t}{2} \|\dB_{\yB}^{(k)} \!-\! \nabla_{\yB} f(\zB^{(k)}) \! \|^2
        \Big] \\
    \le& \EBB \Big[
        \frac{\eta_t}{2} \! \|\nabla_{\xB} f(\zB^{(k)}) \! \|^2
        \!+\! \frac{3 \eta_t}{4} \! \|\dB_{\xB}^{(k)} \! \|^2
        \!-\! \frac{\gamma_t}{2} \! \|\nabla_{\yB} f(\zB^{(k)}) \! \|^2 \\
        &- \frac{\gamma_t}{4} \|\dB_{\yB}^{(k)}\|^2
        + 4 \gamma_t \delta^2 \EBB[\|\zB^{(k)} - \zB_t\|^2] + \gamma_t \varepsilon_{\yB, t}
        \Big],
\end{align*}
where the last inequality follows from the condition $\eta_t, \gamma_t \le 1 / (2 L_f)$ and~\eqref{eq_gradient_similarity_f}.
Combining the above two inequalities, Lemma~\ref{lemma_descent}, and Assumption~\ref{assumption_pl} leads to the desired result.
\end{proof}

We are now ready to prove Theorem~\ref{theorem_alg_storm}.
\begin{proof}[Proof of Theorem~\ref{theorem_alg_storm}]
We define the potential function
\begin{align*}
    \hat{\LM}_{t, i}^{(k)}
    :=& \EBB[\Phi(\xB_{t, i}^{(k)}) \!-\! \Phi(\xB^*)] \!+\! \frac{1}{7} \EBB[ \Phi(\xB_{t, i}^{(k)}) \!-\! f(\zB_{t, i}^{(k)}) ] \\
        &\!+\! \frac{45 (8 L_f \!+\! \delta)^2 \gamma_t}{49 \alpha_t} \big( 1 \!+\! \frac{2}{K} \big)^{K \!-\! k} \EBB \|\xB_{t, i}^{(k)} \!-\! \xB_t\|^2 \\
        &\!+\! \frac{892 \delta^2 K \gamma_t}{\alpha_t} \big( 1 \!+\! \frac{2}{K} \big)^{K \!-\! k} \EBB\|\yB_{t, i}^{(k)} \!-\! \yB_t\|^2.
\end{align*}
Recall that
we set $\alpha_t = 200000 \hat{\delta}^2 K^2 \gamma_t^2$,
$\eta_t = \frac{\gamma_t}{12 \kappa^2}$,
and $\gamma_t = \min\{\frac{1}{2 L_f}, \frac{\num{5.78e-4}}{\hat{\delta} K}, \frac{S^{1/3} \hat{\LM}_0^{1/3}}{\hat{\delta}^{2/3} K (t + 1)^{1/3} (\sigma_1^2 / \kappa^2 + \sigma_2^2)^{1/3}}\}$, where $\kappa = L_f / \mu \ge 1$.
Thus, $\eta_t \le \min\{\frac{1}{2 L_f}, \frac{1}{2 L_{\Phi}}\}$ and $\gamma_t \le \frac{1}{2 L_f}$ hold.
Combining Lemmas~\ref{lemma_gradient_error_storm}-\ref{lemma_max_residue}, we can bound
\begin{align}
    &\hat{\LM}^{(k+1)}
        \!+\! \frac{4.01 \eta_{t+1}}{\alpha_{t+1}} \varepsilon_{\xB, t+1}
        \!+\! \frac{0.2858 \gamma_{t+1}}{\alpha_{t+1}} \varepsilon_{\yB, t+1} \nonumber \\
    \le& \EBB[\Phi(\xB^{(k)}) - \Phi(\xB^*)]
        \!+\! c_1 \EBB[ \Phi(\xB^{(k)}) - f(\zB^{(k)})] \nonumber \\
        &\!-\! \frac{4 \eta_t}{7} \! \EBB\|\nabla_{\xB} \Phi(\xB^{(k)})\|^2
        \!+\! c_2 \EBB\|\xB^{(k)} - \xB_t\|^2
        \!+\! c_3 \EBB\|\yB^{(k)} - \yB_t\|^2 \nonumber \\
        &\!+\! c_4 \varepsilon_{\xB, t}
        \!+\! c_5 \varepsilon_{\yB, t}
        \!+\! c_6 \EBB\| \dB_{\xB}^{(k)} \! \|^2
        \!+\! c_7 \EBB\| \dB_{\yB}^{(k)} \! \|^2
        \!+\! c_8 \EBB\|\zB_{t+1} \!-\! \zB_t \! \|^2 \nonumber \\
        &\!+\! \frac{8.02 \eta_{t+1} \alpha_{t+1}}{|\SM_{t+1}'|} \sigma_1^2
        \!+\! \frac{0.5716 \gamma_{t+1} \alpha_{t+1}}{|\SM_{t+1}'|} \sigma_2^2,  \label{eq_lyapunov_storm_expand}
\end{align}
where $c_1, \ldots, c_8$ are given by
\begin{align*}
    c_1 :=& \frac{1}{7} \big( 1 - \frac{\mu \gamma_t}{2} + \frac{3 \Lot^2}{4 \mu} \eta_t \big)
        + \frac{3 \Lot^2}{4 \mu} \eta_t, \\
    c_2 :=& 7 \delta^2 \eta_t
        + \frac{1}{7}(7 \eta_t + 4 \gamma_t) \delta^2 \\
        &+ \frac{45 (8 L_f + \delta)^2 K \gamma_t}{49 \alpha_t} \big( 1 + \frac{2}{K} \big)^{K - k - 1} \big(1 + \frac{1}{K} \big), \\
    c_3 :=& 7 \delta^2 \eta_t + \frac{1}{7} (7 \eta_t + 4 \gamma_t) \delta^2 \\
        &+ \frac{892 \delta^2 K \gamma_t}{\alpha_t} \big( 1 + \frac{2}{K} \big)^{K - k - 1} \big(1 + \frac{1}{K} \big), \\
    c_4 :=& 2 \eta_t + \frac{4.01 \eta_{t+1}}{\alpha_{t+1}} (1 - \alpha_{t+1})^2, \\
    c_5 :=& \frac{\gamma_t}{7} + \frac{0.2858 \eta_{t+1}}{\alpha_{t+1}} (1 - \alpha_{t+1})^2, \\
    c_6 :=& \frac{5 \eta_t}{56} \!-\! \frac{\eta_t}{8} \!+\! \frac{45 (8 L_f \!+\! \delta)^2 K \gamma_t}{49 \alpha_t} \big( 1 \!+\! \frac{2}{K} \big)^{K \!-\! k \!-\! 1} \big(1 \!+\! K \big) \eta_t^2, \\
    c_7 :=& -\frac{\gamma_t}{28} + \frac{892 \delta^2 K \gamma_t}{\alpha_t} \big( 1 + \frac{2}{K} \big)^{K - k - 1} \big(1 + K \big) \gamma_t^2, \\
    c_8 :=& \frac{(1 - \alpha_{t+1})^2 \delta^2}{\alpha_{t+1}} (32.08 \eta_{t+1} + 2.2864 \gamma_{t+1}).
\end{align*}
Notice that $1 \le (1 + \frac{2}{K})^{K - k} \le 8$ and $\delta \le L_f \le \kappa \hat{\delta}$.
By our choice of $(\eta_t, \gamma_t, \alpha_t)$, we have $\gamma_t /2 \le \gamma_{t+1} \le \gamma_t$, $\eta_t / 2 \le \eta_{t+1} \le \eta_t \le \gamma_t$, and
\begin{align}
    \frac{1}{\alpha_{t \!+\! 1}}
    \!=& \frac{1}{(\num{2e5}) \hat{\delta}^2 \! K^2} \! \max \! \big\{
        4 L_f^2, \!
        \frac{\hat{\delta}^2 \! K^2}{(\num{5.78e-4})^2}, \hat{\delta}^2 \! K^2 (t \!+\! 2)^{2/3} \!
        \big\} \nonumber \\
    \!=& \frac{1}{(\num{2e5}) \hat{\delta}^2 \! K^2} \! \max \! \big\{
        4 L_f^2, \!
        \frac{\hat{\delta}^2 \! K^2}{(\num{5.78e-4})^2}, \hat{\delta}^2 \! K^2 ((t \!+\! 1)^{2/3} \!+\! 1) \!
        \big\} \nonumber \\
    \le& \frac{1}{(\num{2e5}) \hat{\delta}^2 K^2 \gamma_t^2} + \frac{1}{\num{2e5}}
    = \frac{1}{\alpha_t} + \frac{1}{\num{2e5}}.  \label{eq_alpha_inverse_bound}
\end{align} 
Then, we can bound
\begin{align*}
    &c_1 \le \frac{1}{7},
    c_2 \le \frac{45 (8 L_f \!+\! \delta)^2 K \gamma_t}{49 \alpha_t} \big( 1 \!+\! \frac{2}{K} \big)^{K \!-\! k}, \\
    &c_3 \le \frac{892 \delta^2 K \gamma_t}{\alpha_t} \big( 1 \!+\! \frac{2}{K} \big)^{K \!-\! k},
    c_4 \le \frac{4.01 \eta_{t+1}}{\alpha_t},
    c_5 \le \frac{0.2858 \gamma_t}{\alpha_t}, \\
    &c_6 \le 0,
    c_7 \le 0,
    c_8 \le \frac{69 \delta^2 \gamma_t}{\alpha_t}.
\end{align*}

Plugging the bounds of $c_1, \ldots, c_8$ into~\eqref{eq_lyapunov_storm_expand}, we get
\begin{align}
    &\hat{\LM}^{(k+1)}
        + \frac{4.01 \eta_{t+1}}{\alpha_{t+1}} \varepsilon_{\xB, t+1}
        + \frac{0.2858 \gamma_{t+1}}{\alpha_{t+1}} \varepsilon_{\yB, t+1} \nonumber \\
    \le& \hat{\LM}^{(k)}
        \!-\! \frac{4 \eta_t}{7} \EBB\|\nabla_{\xB} \Phi(\xB^{(k)})\|^2
        \!+\! \frac{4.01 \eta_{t}}{\alpha_{t}} \varepsilon_{\xB, t}
        \!+\! \frac{0.2858 \gamma_{t}}{\alpha_{t}} \varepsilon_{\yB, t} \nonumber \\
        &\!+\! \frac{69 \delta^2 \gamma_t}{\alpha_t} \EBB\|\zB_{t \!+\! 1} \!-\! \zB_t\|^2
        \!+\! \frac{8.02 \eta_{t+1} \alpha_{t+1}}{|\SM_{t+1}'|} \sigma_1^2 \nonumber \\
        &\!+\! \frac{0.5716 \gamma_{t+1} \alpha_{t+1}}{|\SM_{t+1}'|} \sigma_2^2.  \label{eq_lyapunov_storm_simplified}
\end{align}

Similar to~\cite{karimireddy2020scaffold, karimireddy2021breaking}, we need a technical condition that $f(\xB, \yB)$ is $\delta$-weakly concave w.r.t.\ $(\xB, \yB)$.
By this condition, the function $-f(\xB, \yB) + \frac{\delta}{2} \|\xB - \xB'\|^2 + \frac{\delta}{2} \|\yB - \yB'\|^2$ is convex w.r.t.\ $(\xB, \yB)$, $\forall (\xB', \yB') \in \RBB^{p \times q}$.
In addition, $\Phi(\xB)$ is $L_f$-weakly convex since $\Phi(\xB_1) - \Phi(\xB_2) \le f(\xB_1, \yB_1^*) - f(\xB_2, \yB_1^*) \le \langle \nabla_{\xB} f(\xB_1, \yB_1^*), \xB_1 - \xB_2 \rangle + \frac{L_f}{2} \|\xB_1 - \xB_2\|^2$, where $\yB_1^* \in \argmax_{\yB} f(\xB_1, \yB)$.
Besides, by our choice of $(\alpha_t, \gamma_t)$,
we have
$
    \frac{45 (8 L_f + \delta)^2 K \gamma_t}{49 \alpha_t}
    \ge \frac{8 L_f + \delta}{14},
    \frac{892 \delta^2 K \gamma_t}{\alpha_t} \ge \frac{\delta}{14}.
$
Therefore, the function
$
    \Phi(\xB) - \Phi(\xB^*)
        + \frac{1}{7} (\Phi(\xB) - f(\xB, \yB))
        + \frac{45 (8 L_f + \delta)^2 K \gamma_t}{49 \alpha_t} \|\xB - \xB_t\|^2
        + \frac{892 \delta^2 K \gamma_t}{\alpha_t} \|\yB - \yB_t\|^2
$
is convex w.r.t.\ $(\xB, \yB)$, which implies that
\begin{align}
    &{\textstyle \frac{1}{|\SM_t|} \! \sum_{i \in \SM_t} \! \Big( \Phi(\xB^{(K)}) \!-\! \Phi(\xB^*)}
        \!+\! {\textstyle \frac{1}{7} \! (\Phi(\xB^{(K)}) \!-\! f(\zB^{(K)}))} \nonumber \\
        &\!+\! {\textstyle \frac{45 (8 L_f \!+\! \delta)^2 \! K \! \gamma_t}{49 \alpha_t} \! \|\xB^{( \! K \! )} \!-\! \xB_t \! \|^2}
        \!+\! {\textstyle \frac{892 \delta^2 \! K \! \gamma_t}{\alpha_t} \! \|\yB^{( \! K \! )} \!-\! \yB_t \! \|^2 \Big)} \nonumber \\
    \ge& \Phi(\xB_{t+1}) \!-\! \Phi(\xB^*)
        \!+\! {\textstyle \frac{1}{7} (\Phi(\xB_{t+1}) \!-\! f(\zB_{t+1}))} \nonumber \\
        &\!+\! {\textstyle \frac{45 (8 L_f \!+\! \delta)^2 \! K \! \gamma_t}{49 \alpha_t} \! \|\xB_{t \!+\! 1} \!-\! \xB_t \! \|^2}
        \!+\! {\textstyle \frac{892 \delta^2 \! K \! \gamma_t}{\alpha_t} \! \|\yB_{t \!+\! 1} \!-\! \yB_t \! \|^2}.  \label{eq_potential_function_non_increasing}
\end{align}
Note that the weak concavity condition is only used to ensure that~\eqref{eq_potential_function_non_increasing} holds, that is, the potential function value at the last local step does not increase after parameter averaging on the server.
If we instead select only one client to participate in the parameter update phase in Algorithm~\ref{algorithm_alg_mb_server}, we will obtain the same convergence rate as shown below without the weak concavity condition.

Notice that $\|\zB_{t+1} - \zB_t\|^2 = \|\xB_{t+1} - \xB_t\|^2 + \|\yB_{t+1} - \yB_t\|^2$ and $45 (8 L_f + \delta)^2 / 49 > 69 \delta^2$.
Combining~\eqref{eq_lyapunov_storm_simplified} and~\eqref{eq_potential_function_non_increasing} yields
\begin{align*}
    &{\textstyle \frac{4 \eta_t}{7 |\SM_t| K} \sum_{i \in \SM_t} \sum_{k=0}^{K-1} \EBB[\| \nabla_{\xB} \Phi(\xB_{t, i}^{(k)})\|^2]} \\
    \le& \Big( {\textstyle \frac{1}{K}} \EBB[\Phi(\xB_t) \!-\! \Phi(\xB^*)]
        \!+\! {\textstyle \frac{1}{7 K}} \EBB[\Phi(\xB_t) \!-\! f(\zB_t)]
        \!+\! {\textstyle \frac{4.01 \eta_t}{\alpha_t}} \varepsilon_{\xB, t} \\
        &\!+\! {\textstyle \frac{0.2858 \gamma_t}{\alpha_t}} \varepsilon_{\yB, t} \Big)
        \!+\! {\textstyle \frac{8.02 \eta_{t+1} \alpha_{t+1}}{|\SM_{t+1}'|}} \sigma_1^2
        \!+\! {\textstyle \frac{0.5716 \gamma_{t \!+\! 1} \alpha_{t \!+\! 1}}{|\SM_{t+1}'|}} \sigma_2^2 \\
        &\!-\! \Big( \! {\textstyle \frac{1}{K}} \EBB[\Phi(\xB_{t+1}) \!-\! \Phi(\xB^*)]
        \!+\! {\textstyle \frac{1}{7 K}} \EBB[\Phi(\xB_{t+1}) \!-\! f(\zB_{t+1})] \\
        &\!+\! {\textstyle \frac{4.01 \eta_{t+1}}{\alpha_{t+1}}} \varepsilon_{\xB, t+1}
        \!+\! {\textstyle \frac{0.2858 \gamma_{t+1}}{\alpha_{t+1}}} \varepsilon_{\yB, t+1} \Big).
\end{align*}
Telescoping the above inequality yields
\begin{align*}
    &{\textstyle \frac{1}{K T} \sum_{t=0}^{T-1} \frac{1}{|\SM_t|} \sum_{k=0}^{K-1} \sum_{i \in \SM_t} \EBB[\|\nabla \Phi(\xB_{t, i}^{(k)})\|^2]} \\
    \le{}& {\textstyle \OM \big( \frac{\hat{\LM}_0}{K T \eta_{T-1}}
        + \frac{\eta_0 \sigma_1^2}{S \alpha_0 T \eta_{T-1}}
        + \frac{\gamma_0 \sigma_2^2}{S \alpha_0 T \eta_{T-1}}} \\
        &+ {\textstyle \frac{\sigma_1^2}{S T \eta_{T-1}} \sum_{t=0}^{T-1} \eta_{t+1} \alpha_{t+1}
        + \frac{\sigma_2^2}{S T \eta_{T-1}} \sum_{t=0}^{T-1} \gamma_{t+1} \alpha_{t+1} \big)} \\
    =& \tilde{\OM} \big( 
        {\textstyle \frac{\kappa^2 \hat{\LM}_0}{T} (\frac{L_f}{K} + \hat{\delta})
        + \frac{\kappa^2 \hat{\delta}^{2/3} \hat{\LM}_0^{2/3} (\sigma_1^2 / \kappa^2 + \sigma_2^2)^{1/3}}{S^{1/3} T^{2/3}}} \\
        &+ {\textstyle \frac{\kappa^2 (\sigma_1^2 / \kappa^2 + \sigma_2^2)}{S} (\frac{L_f / (\hat{\delta} K) + 1}{T^{2/3}} + \frac{L_f^2 / (\hat{\delta}^2 K^2) + 1}{T})}
    \big),
\end{align*}
which is the desired result.
\end{proof}


\section{Additional Details and Results of Experiments}  \label{section_appendix_experiments}

\subsection{Details of Experimental Settings}  \label{section_experimental_setting_details}

\paragraph{AUC Maximization.} 
To make the data imbalanced, we split the original MNIST dataset (resp., CIFAR-10) into two classes by treating the original ``0'' (resp., ``airplane'') class as the positive class and the remaining as negative.
In addition, we preprocess the data by rescaling each pixel value to the range $[-1, 1]$.
We use LeNet5~\cite{lecun1998gradient} as the classification model for MNIST, and a convolutional neural network from~\cite{tensorflow2021cnn} for CIFAR-10 following~\cite{mcmahan2017communication}.

\paragraph{Robust Adversarial Network Training.}
Let $h_{\xB}$ denote a 2-layer MLP with the ReLU activation, where each layer consists of 200 neurons.
Our goal is to solve problem~\eqref{eq_problem} with each $f_i$ defined as
$
    f_i(\xB, \yB)
    := \frac{1}{n_i} \sum_{j=1}^{n_i} \ell(h_{\xB}(\aB_{i, j} + \yB), b_{i, j}),
$
where $\ell$ denotes the cross entropy function and
the feature-label pair $(\aB_{i, j}, b_{i, j})$ is the $j$-th data sample on client $i$.
Here, the dual variable $\yB$ is subject to the $\ell_2$-norm ball constraint $\{\yB: \|\yB\|_2 \le 1\}$ and the goal is to find a model parameter $\xB$ robust to the adversarial noise in this unit ball.
This constrained minimax problem can be further relaxed to an unconstrained one by adding an $\ell_2$ regularization term $- \frac{\lambda}{2}\|\yB\|^2$ to each $f_i$.
The performance measure is the regularized robust loss defined as $\max_{\yB} \EBB_{i \sim \DM} [ \frac{1}{n_i} \sum_{j=1}^{n_i} \ell(h_{\xB_t}(\aB_{i, j} + \yB), b_{i, j}) - \frac{\lambda}{2} \|\yB\|^2 ]$.
To calculate this value, we run the gradient ascent algorithm following~\cite{deng2021local}.

\paragraph{GAN training.}
Denote the generator network parameterized by $\xB$ as $G_{\xB}$ and the discriminator network parameterized by $\yB$ as $D_{\yB}$.
The local loss function is defined as
$
    f_i(\xB, \yB)
    \!=\! \frac{1}{n_i} \! \sum_{j=1}^{n_i} \log \! D_{\yB}(\aB_{i, j})
    + \EBB_{\wB \!\sim\! \PM}[ \log(1 - D_{\yB}(G_{\xB}(\wB)))],
$
where $\aB_{i, j}$ is the feature of the $j$-th data point on client $i$ and $\PM$ is the prior distribution of the noise vector for generating samples.
The data partitioning and preprocessing schemes of MNIST, Fashion MNIST are the same as those in Section~\ref{section_experiment_robustnn}.
For CelebA, the training data is first sorted according to the face ID and then equally partitioned into 500 clients, and each image is resized to $64 \times 64$ and normalized to $[-1, 1]$.
Following~\cite{radford2015unsupervised}, we use the DCGAN architecture and set $\PM$ to the uniform distribution over $[-1, 1]^{100}$.
On MNIST and Fashion MNIST, each algorithm runs with the minibatch size $B = 30$ and the number of local steps $K = 20$ for $5$ local epochs.
On the more difficult dataset CelebA, we set $B = 60$ and select the best $K$ from $\{10, 15, 20, 25\}$ for each algorithm because we found that training GANs on CelebA tends to be unstable when $K$ is large for some of the compared algorithms.


\subsection{Additional Results of the GAN training task}  \label{section_additional_experimental_results_gan}

\begin{figure*}[t]
    \captionsetup[subfloat]{farskip=1pt,captionskip=1pt}
    \centering
    \subfloat{\includegraphics[trim={0cm 0cm 0cm 0cm}, clip, width=.31\linewidth]{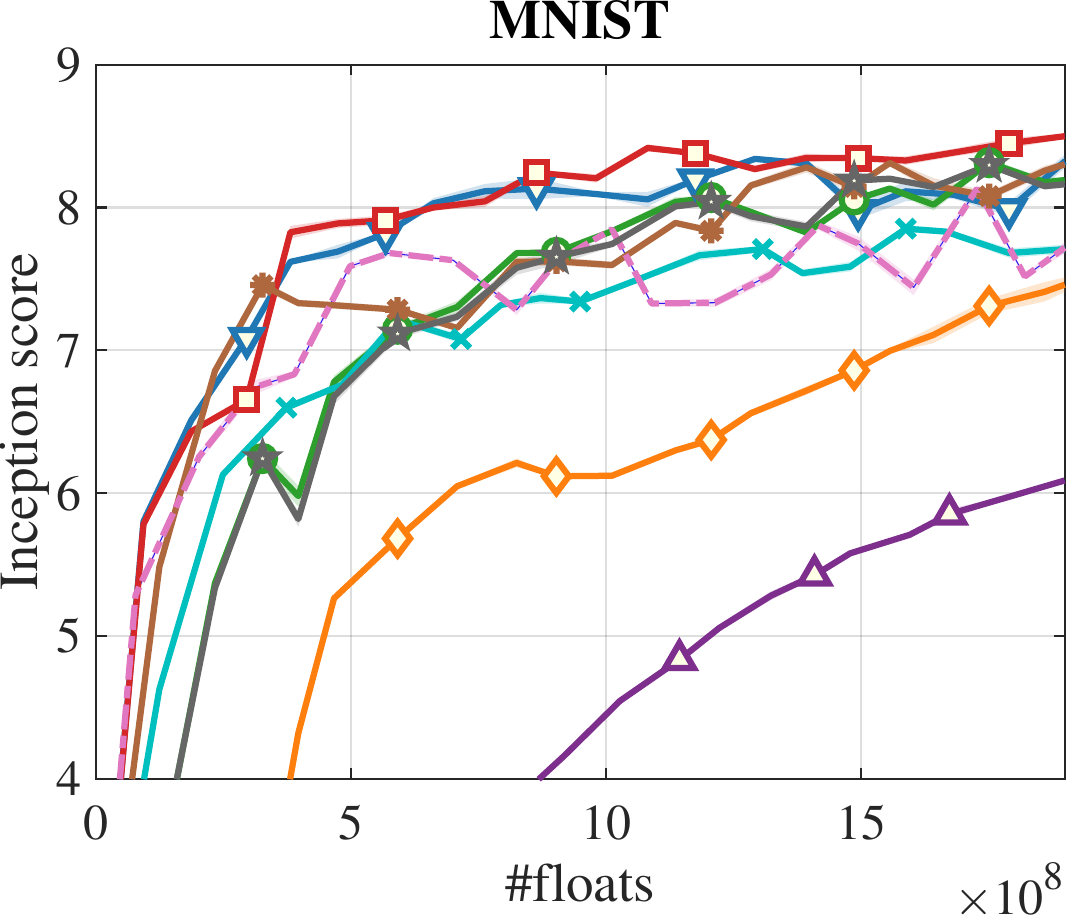}}
    \hfil
    \subfloat{\includegraphics[trim={0cm 0cm 0cm 0cm}, clip, width=.31\linewidth]{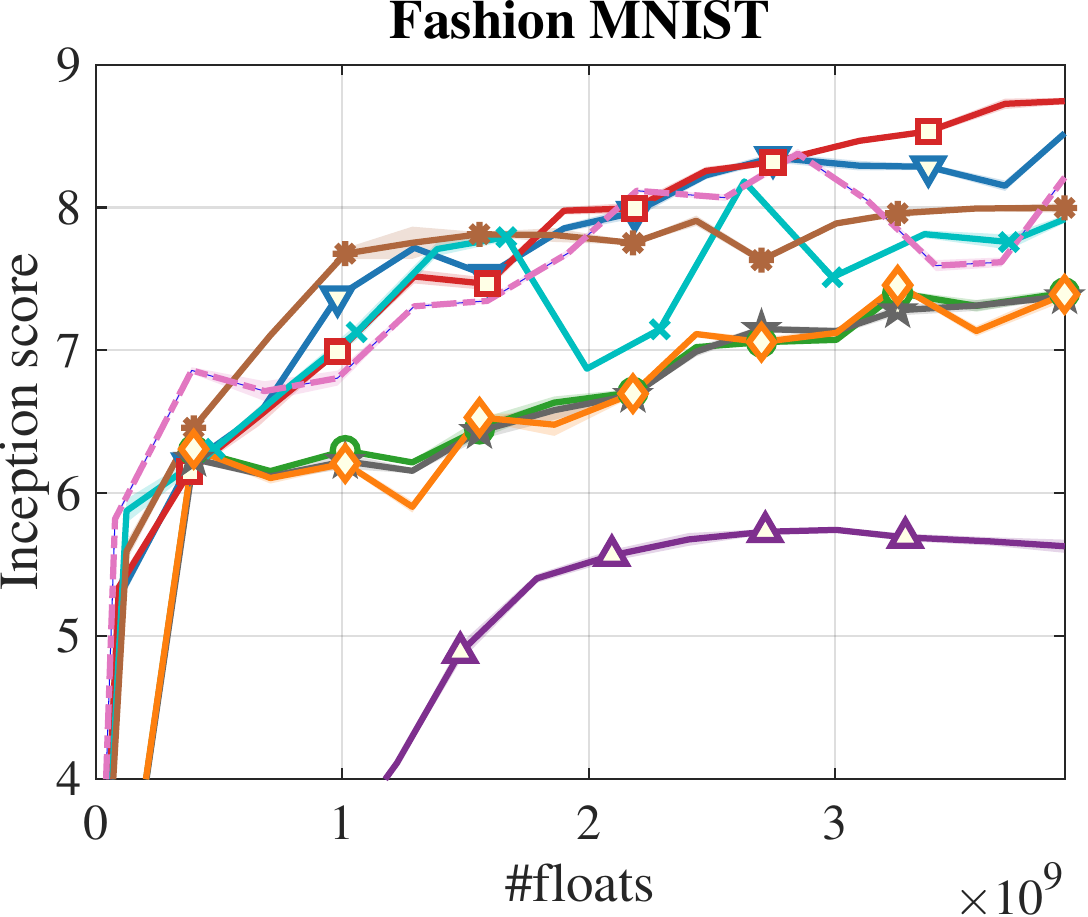}}
    \subfloat{\includegraphics[trim={0cm 0cm 0cm 0cm}, clip, width=.32\linewidth]{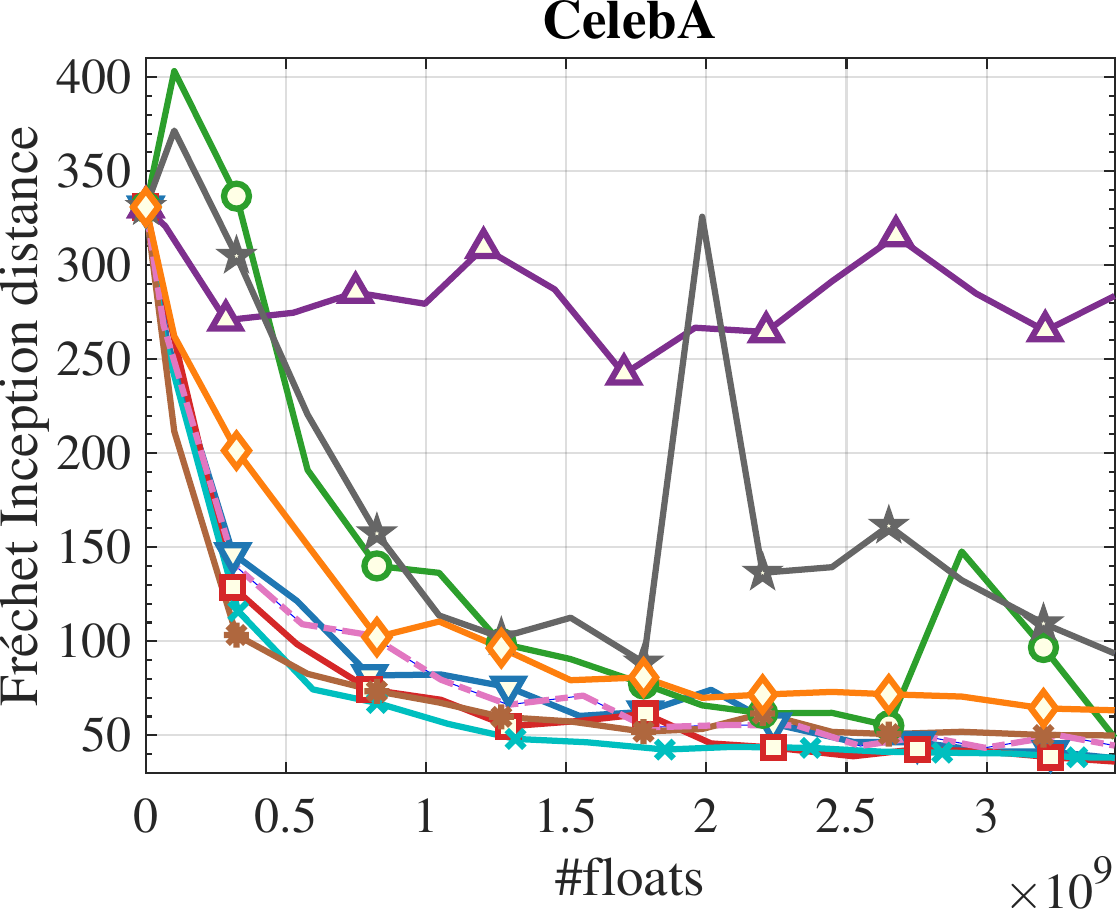}}
    \\[0.2em]
    \subfloat{\includegraphics[trim={-1.05cm 0cm 0cm 0cm}, clip, width=.75\linewidth]{./imgs_journal/legend}}
    \caption{
        The training curves on the GAN training task (left: MNIST, middle: Fashion MNIST, right: CelebA). The horizontal axis stands for the amount of communication and the vertical axis represents the IS on MNIST and Fashion MNIST (resp., FID on CelebA).
        Note that larger IS is better, while smaller FID is better.
    }
    \label{figure_gan}
\end{figure*}

\begin{figure*}[t]
    \centering
    \subfloat[{\AlgMB}]{\includegraphics[trim={2cm 1.5cm 1.5cm 1.5cm}, clip, width=.31\linewidth]{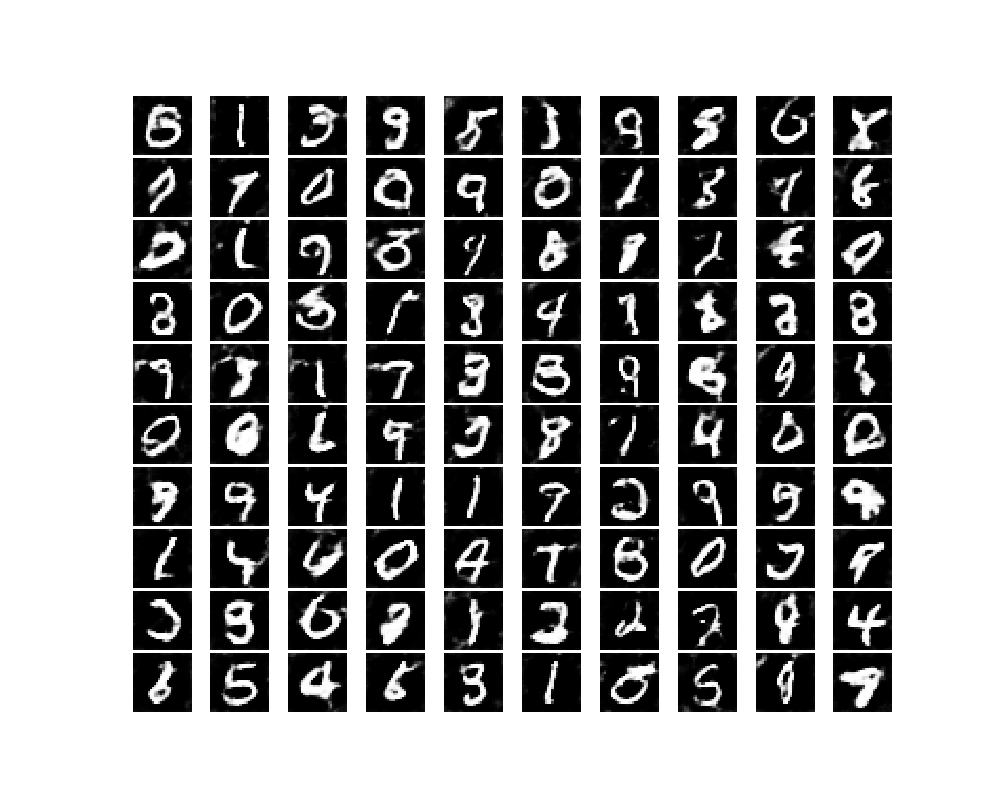}}
    \hfil
    \subfloat[{\AlgSTORM}]{\includegraphics[trim={2cm 1.5cm 1.5cm 1.5cm}, clip, width=.31\linewidth]{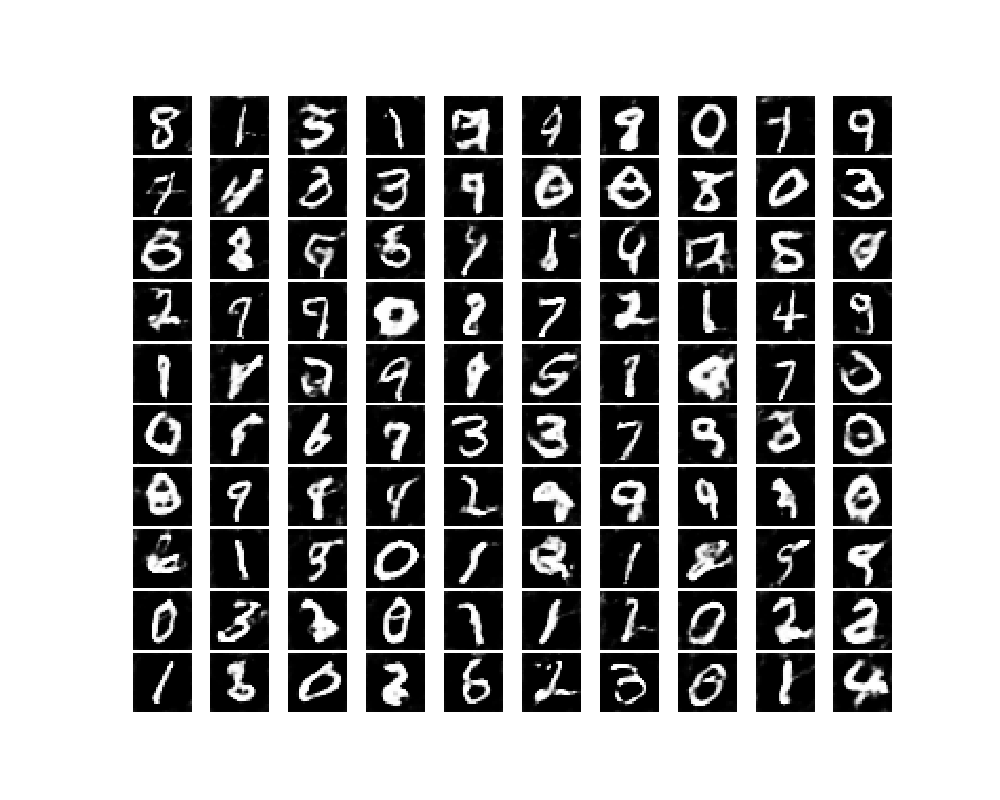}}
    \hfil
    \subfloat[{\AlgNaive}]{\includegraphics[trim={2cm 1.5cm 1.5cm 1.5cm}, clip, width=.31\linewidth]{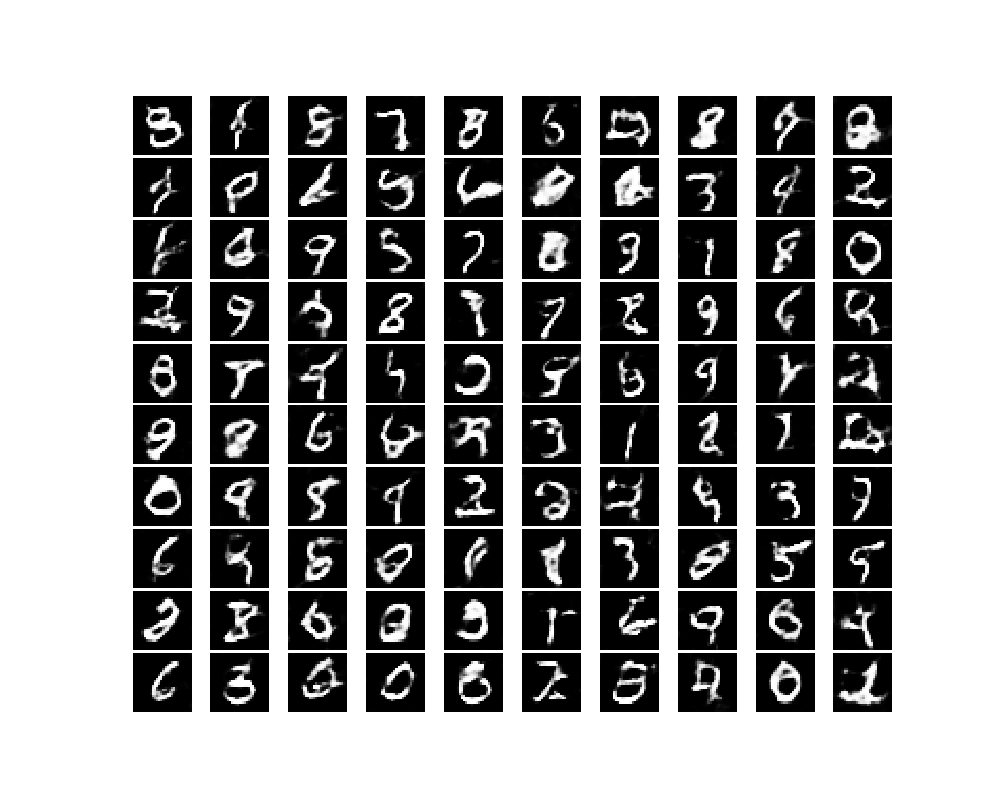}}
    \\
    \subfloat[Parallel SGDA]{\includegraphics[trim={2cm 1.5cm 1.5cm 1.5cm}, clip, width=.31\linewidth]{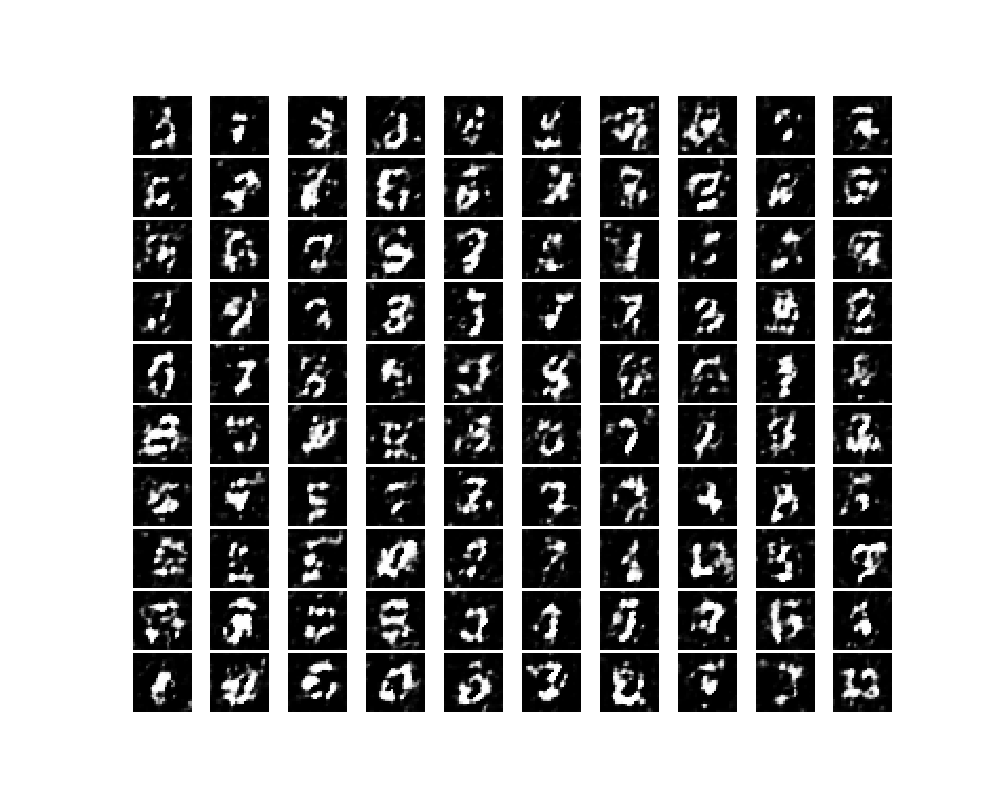}}
    \hfil
    \subfloat[CODASCA]{\includegraphics[trim={2cm 1.5cm 1.5cm 1.5cm}, clip, width=.31\linewidth]{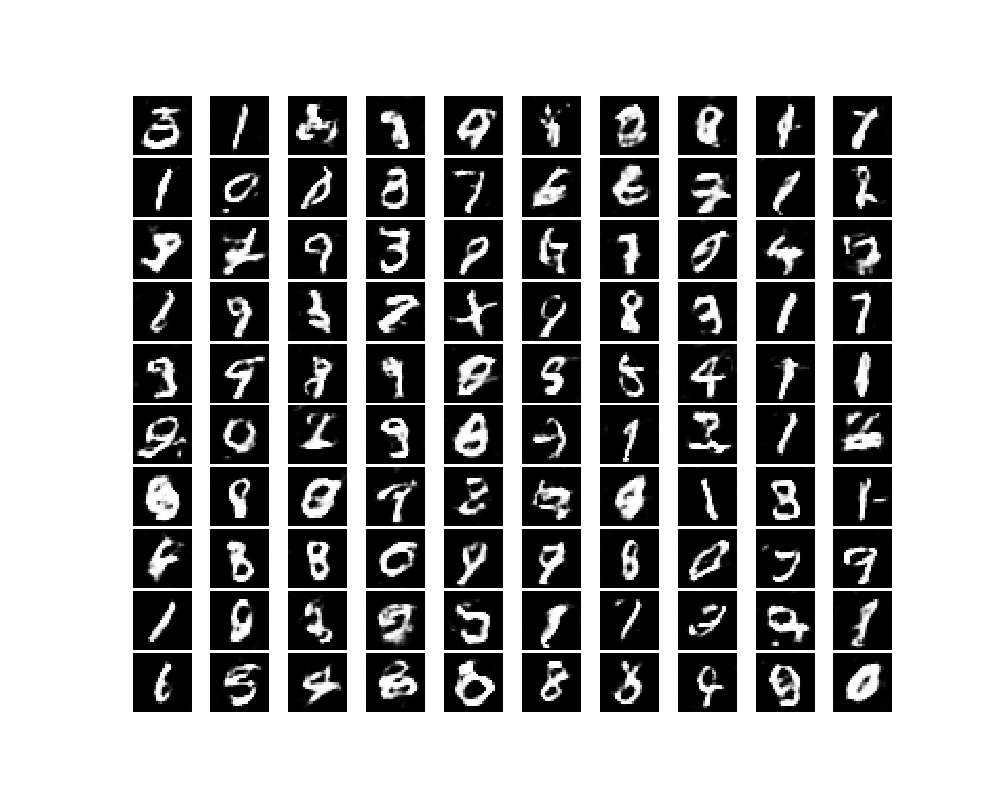}}
    \hfil
    \subfloat[CODA+]{\includegraphics[trim={2cm 1.5cm 1.5cm 1.5cm}, clip, width=.31\linewidth]{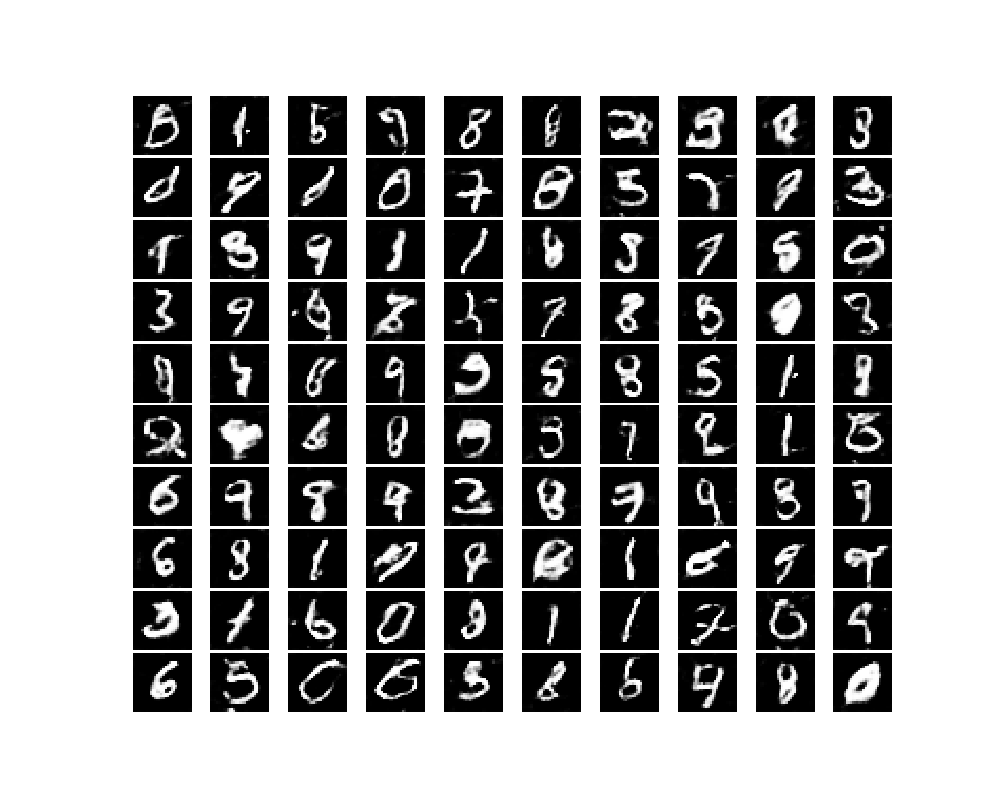}}
    \\
    \subfloat[Catalyst-Scaffold-S]{\includegraphics[trim={2cm 1.5cm 1.5cm 1.5cm}, clip, width=.31\linewidth]{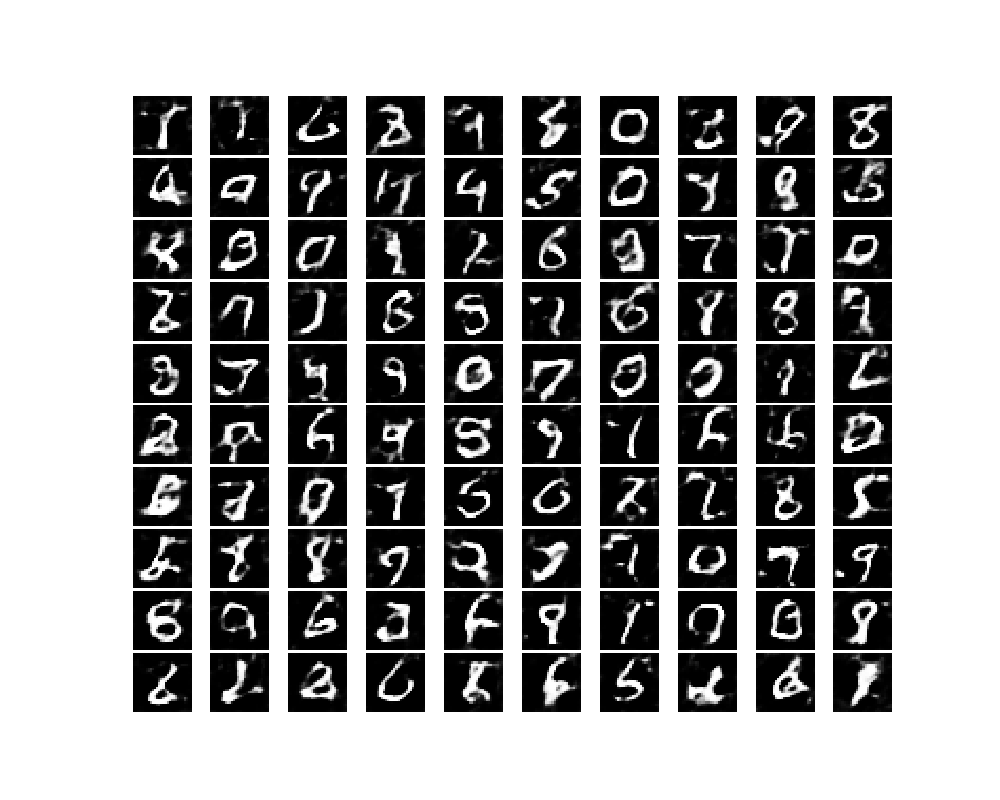}}
    \hfil
    \subfloat[Local SGDA+]{\includegraphics[trim={2cm 1.5cm 1.5cm 1.5cm}, clip, width=.31\linewidth]{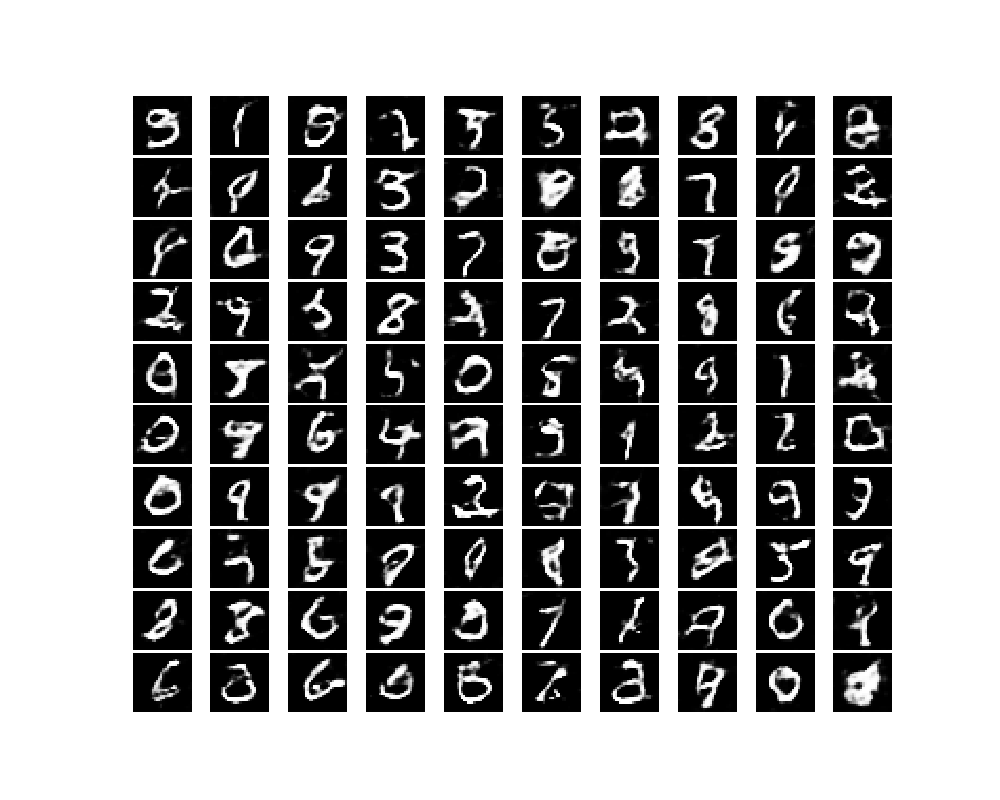}}
    \hfil
    \subfloat[Extra Step Local SGD]{\includegraphics[trim={2cm 1.5cm 1.5cm 1.5cm}, clip, width=.31\linewidth]{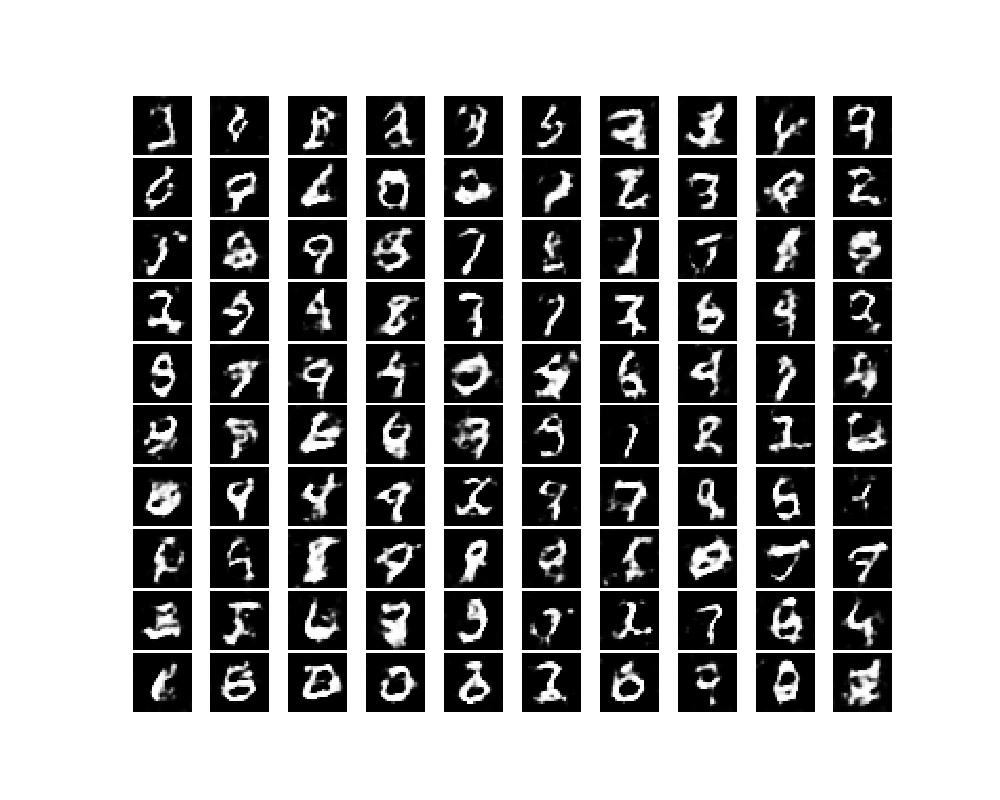}}
    \caption{
        Generative samples of the compared algorithms on MNIST.
        The samples are generated using the same set of random noise vectors.
    }
    \label{figure_generative_images}
\end{figure*}

\begin{figure*}[t]
    \centering
    \subfloat[{\AlgMB}]{\includegraphics[trim={2cm 1.5cm 1.5cm 1.5cm}, clip, width=.31\linewidth]{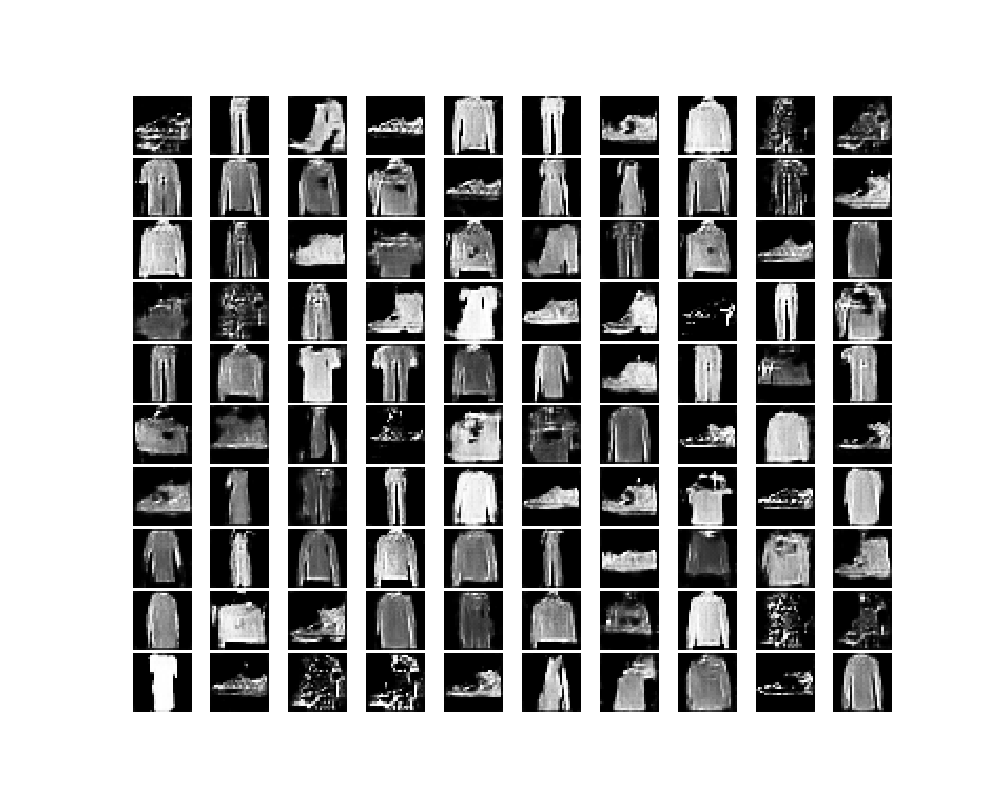}}
    \hfil
    \subfloat[{\AlgSTORM}]{\includegraphics[trim={2cm 1.5cm 1.5cm 1.5cm}, clip, width=.31\linewidth]{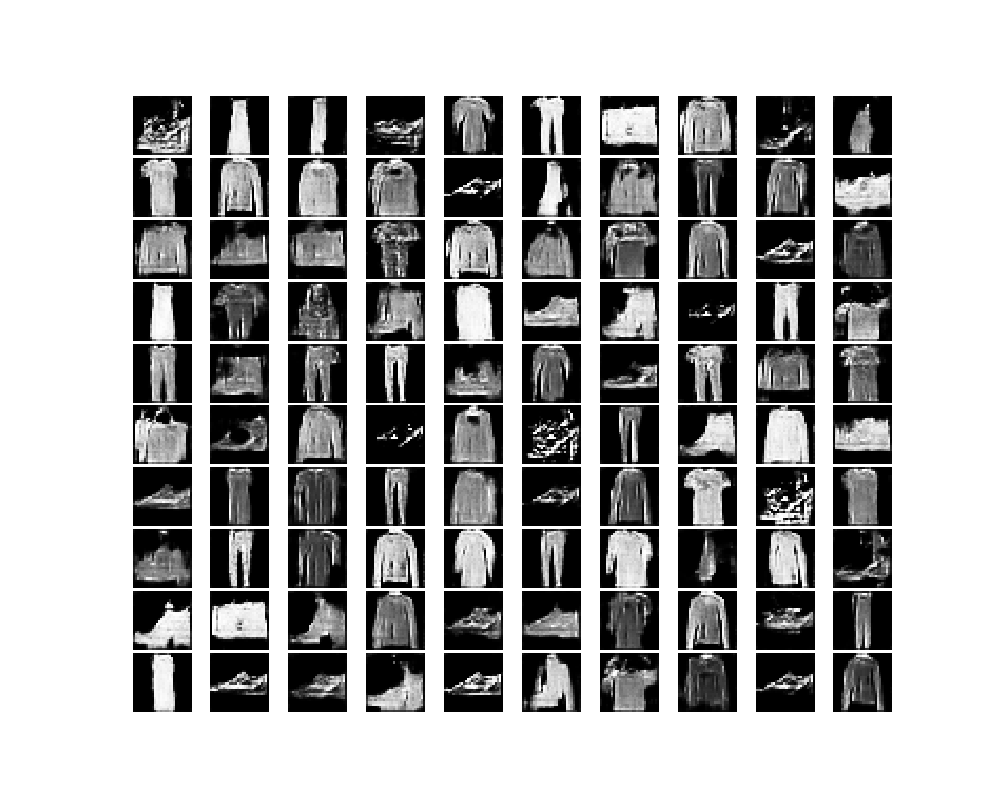}}
    \hfil
    \subfloat[{\AlgNaive}]{\includegraphics[trim={2cm 1.5cm 1.5cm 1.5cm}, clip, width=.31\linewidth]{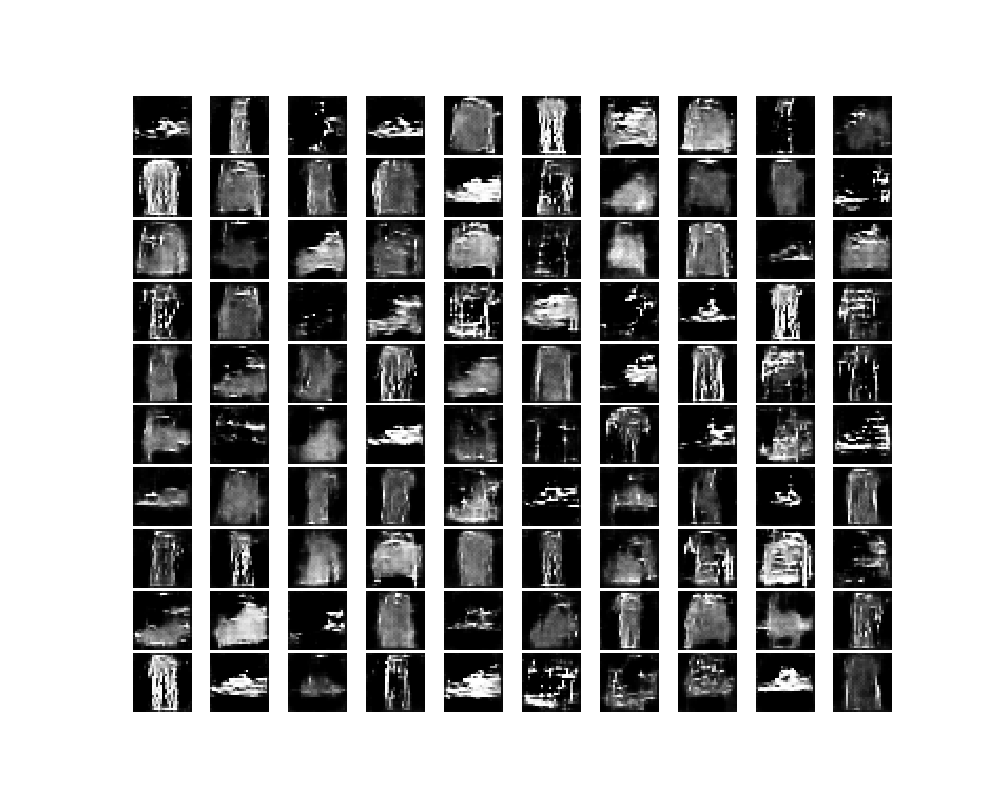}}
    \\
    \subfloat[Parallel SGDA]{\includegraphics[trim={2cm 1.5cm 1.5cm 1.5cm}, clip, width=.31\linewidth]{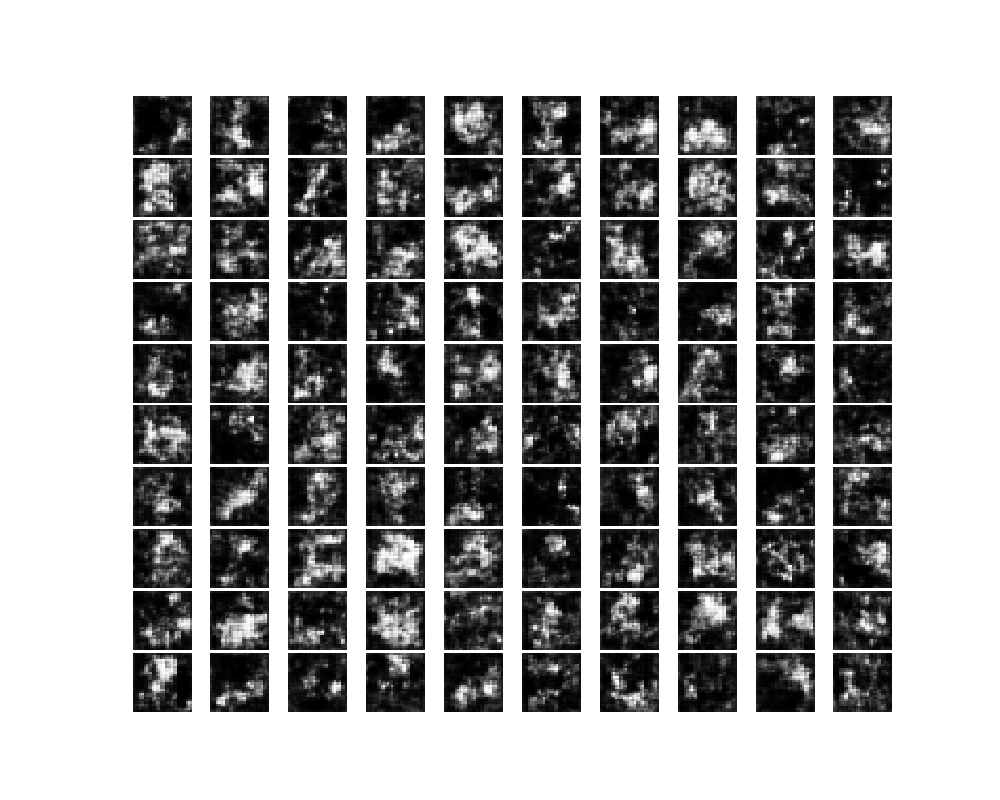}}
    \hfil
    \subfloat[CODASCA]{\includegraphics[trim={2cm 1.5cm 1.5cm 1.5cm}, clip, width=.31\linewidth]{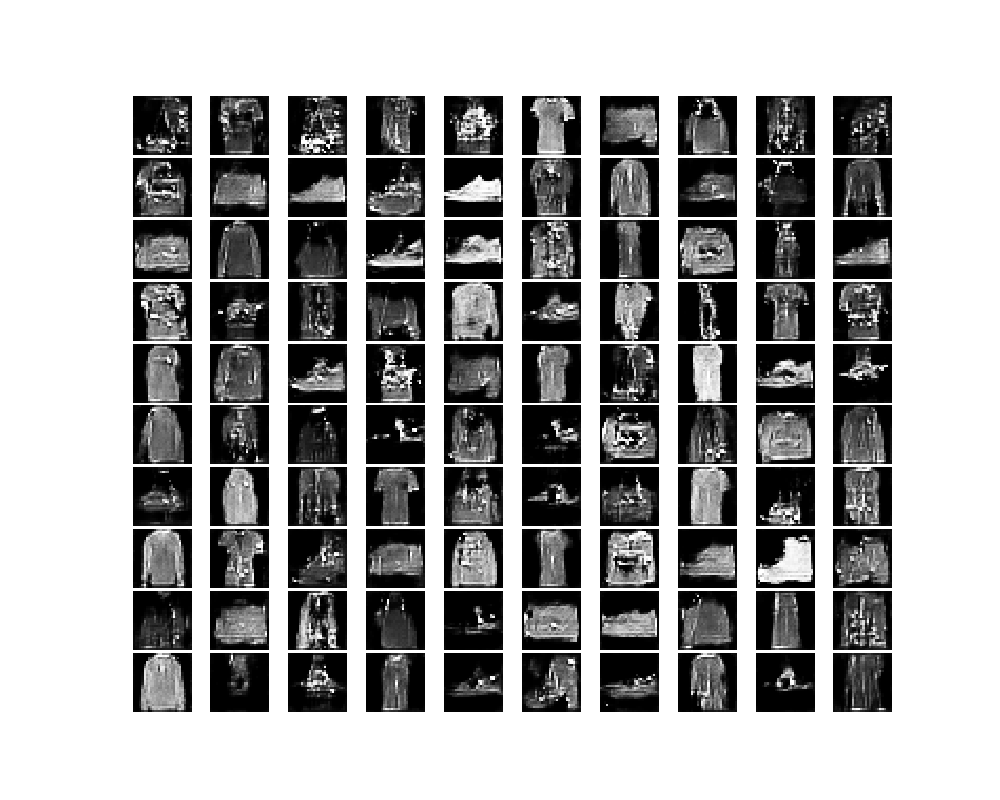}}
    \hfil
    \subfloat[CODA+]{\includegraphics[trim={2cm 1.5cm 1.5cm 1.5cm}, clip, width=.31\linewidth]{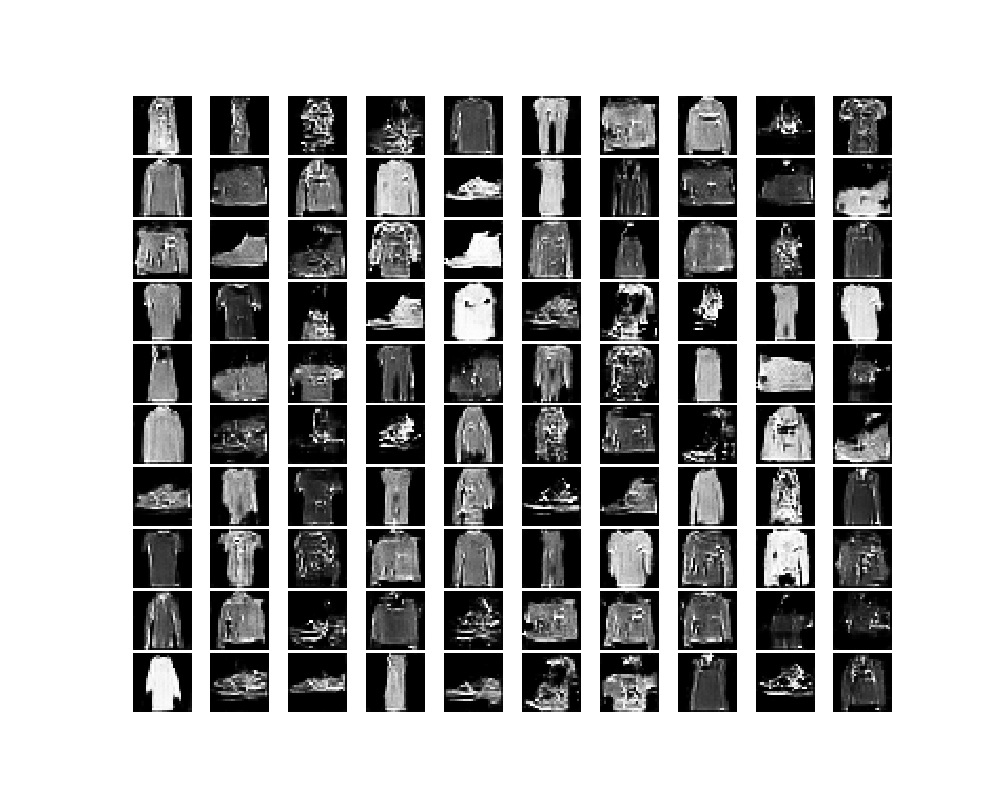}}
    \\
    \subfloat[Catalyst-Scaffold-S]{\includegraphics[trim={2cm 1.5cm 1.5cm 1.5cm}, clip, width=.31\linewidth]{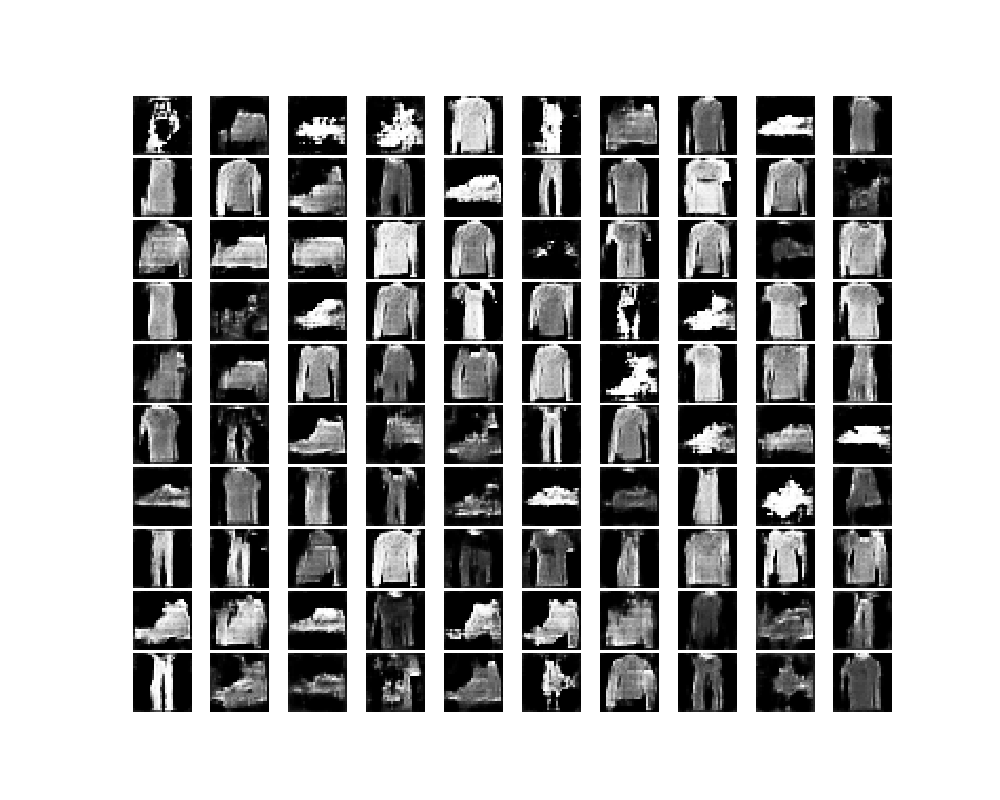}}
    \hfil
    \subfloat[Local SGDA+]{\includegraphics[trim={2cm 1.5cm 1.5cm 1.5cm}, clip, width=.31\linewidth]{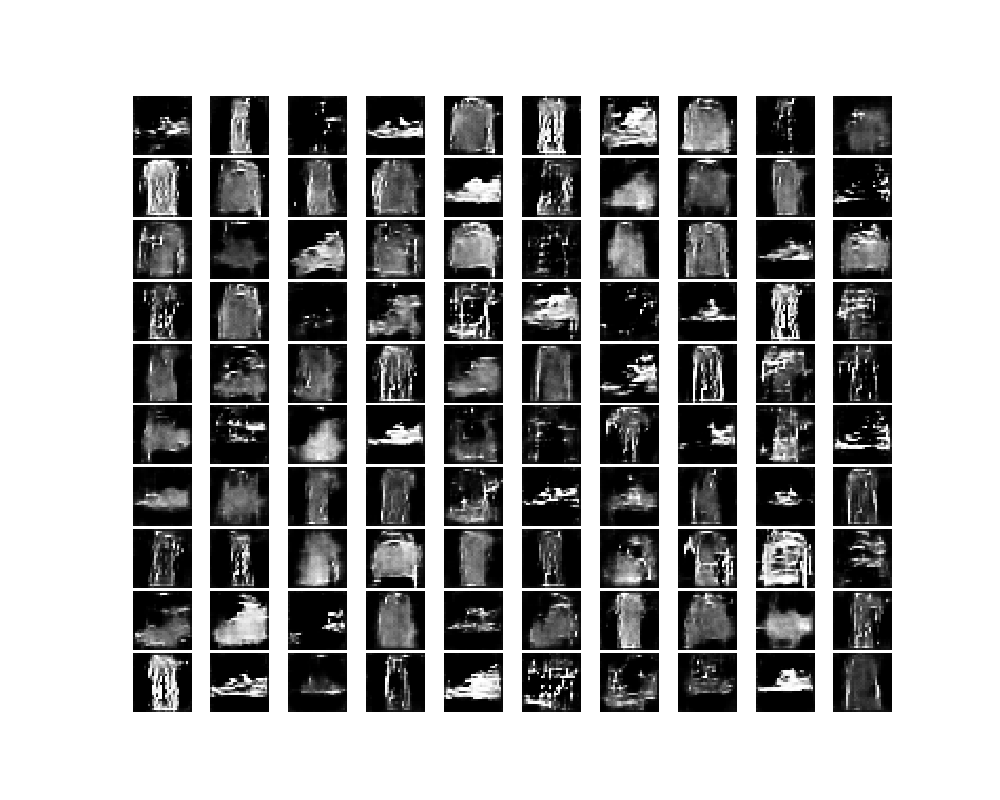}}
    \hfil
    \subfloat[Extra Step Local SGD]{\includegraphics[trim={2cm 1.5cm 1.5cm 1.5cm}, clip, width=.31\linewidth]{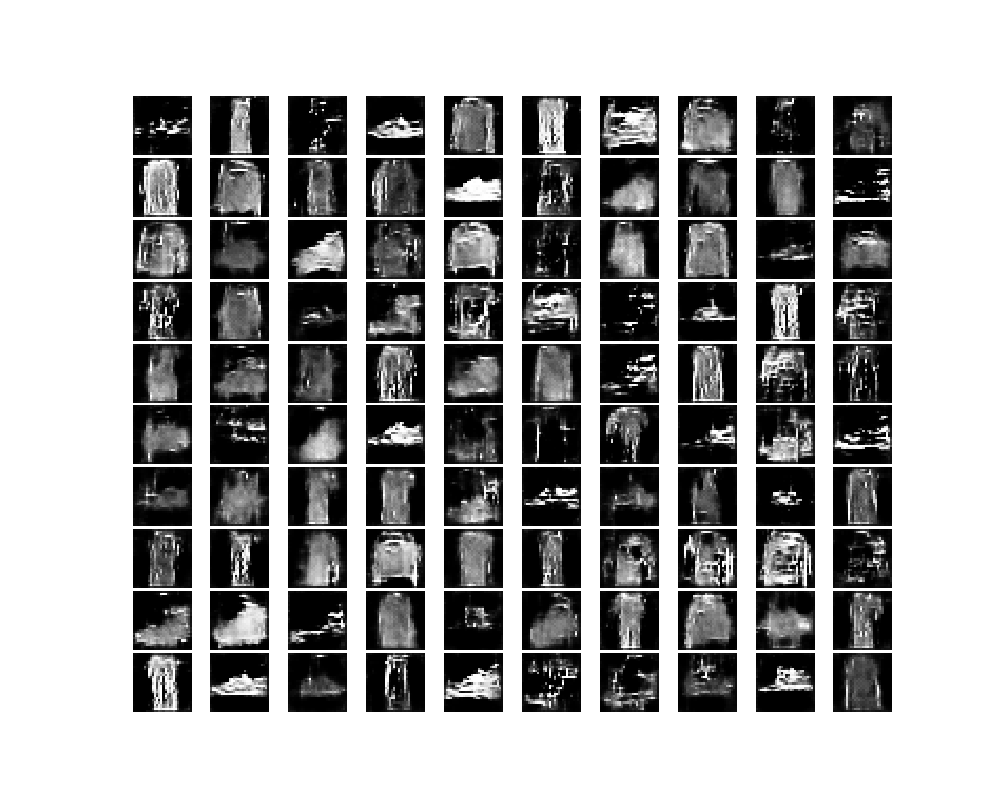}}
    \caption{
        Generative samples of the compared algorithms on Fashion MNIST.
        The samples are generated using the same set of random noise vectors.
    }
    \label{figure_generative_images_fashion_mnist}
\end{figure*}

\begin{figure*}[t]
    \centering
    \subfloat[{\AlgMB}]{\includegraphics[trim={2cm 1.5cm 1.5cm 1.5cm}, clip, width=.31\linewidth]{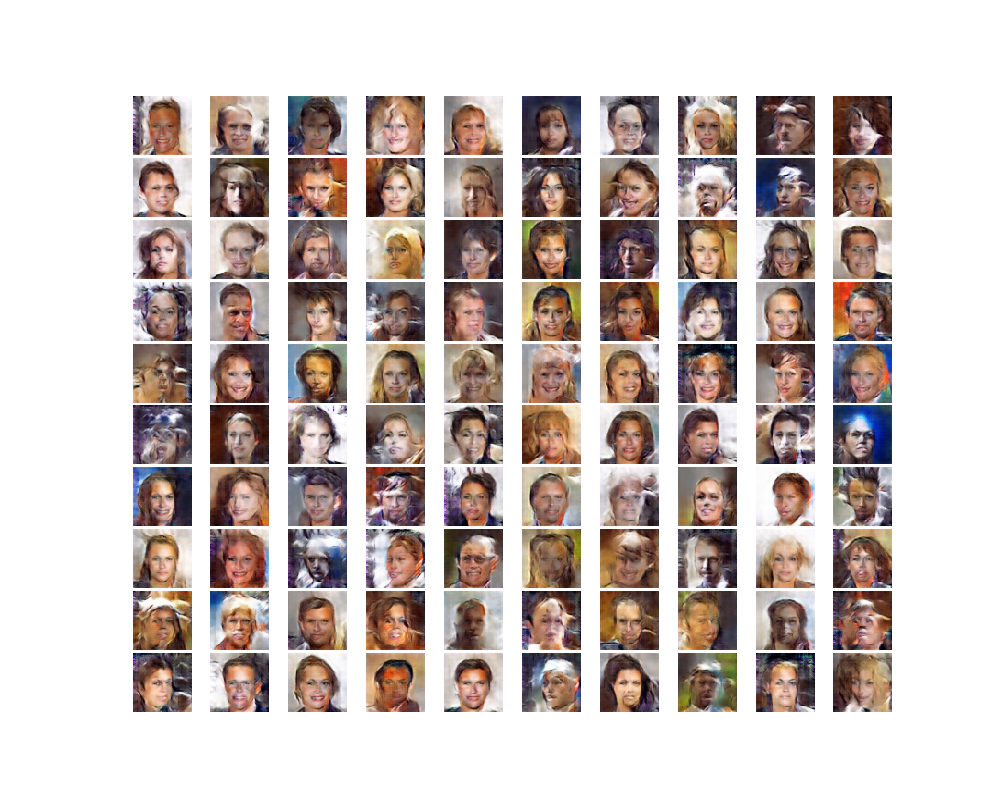}}
    \hfil
    \subfloat[{\AlgSTORM}]{\includegraphics[trim={2cm 1.5cm 1.5cm 1.5cm}, clip, width=.31\linewidth]{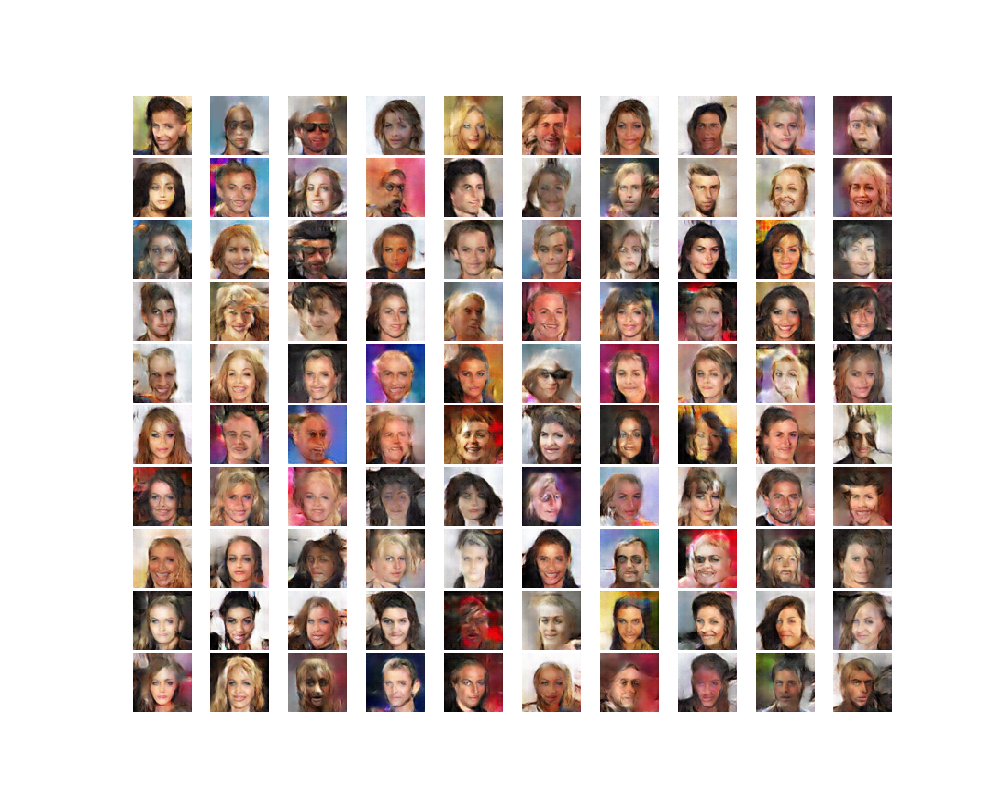}}
    \hfil
    \subfloat[{\AlgNaive}]{\includegraphics[trim={2cm 1.5cm 1.5cm 1.5cm}, clip, width=.31\linewidth]{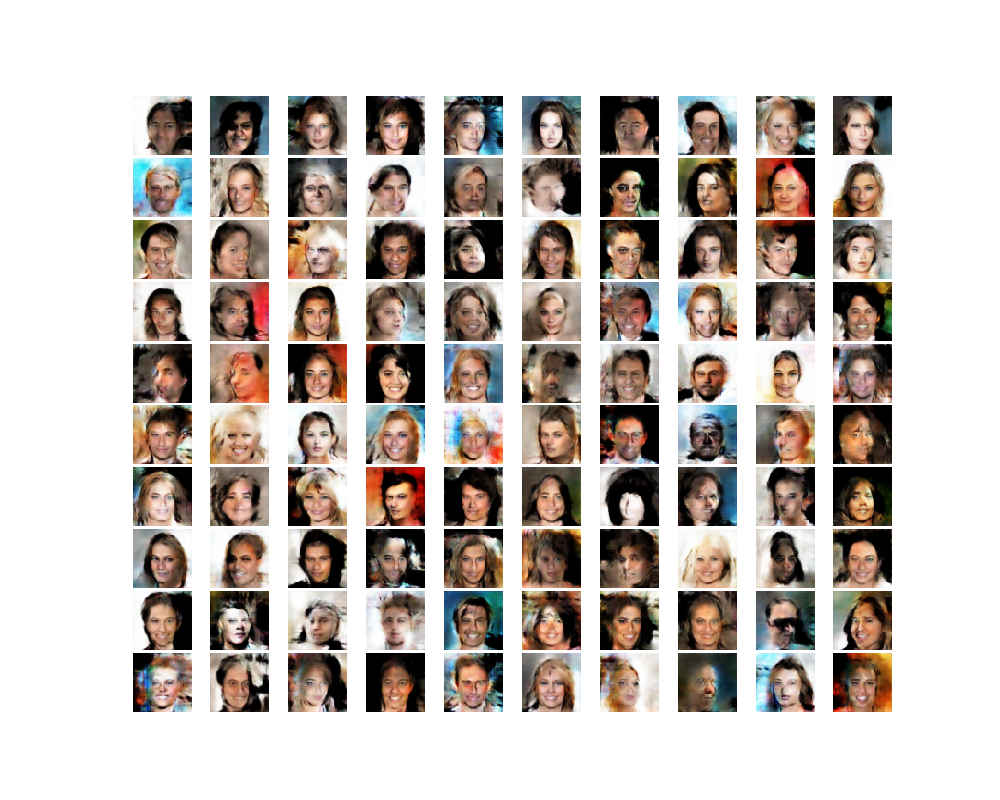}}
    \\
    \subfloat[Parallel SGDA]{\includegraphics[trim={2cm 1.5cm 1.5cm 1.5cm}, clip, width=.31\linewidth]{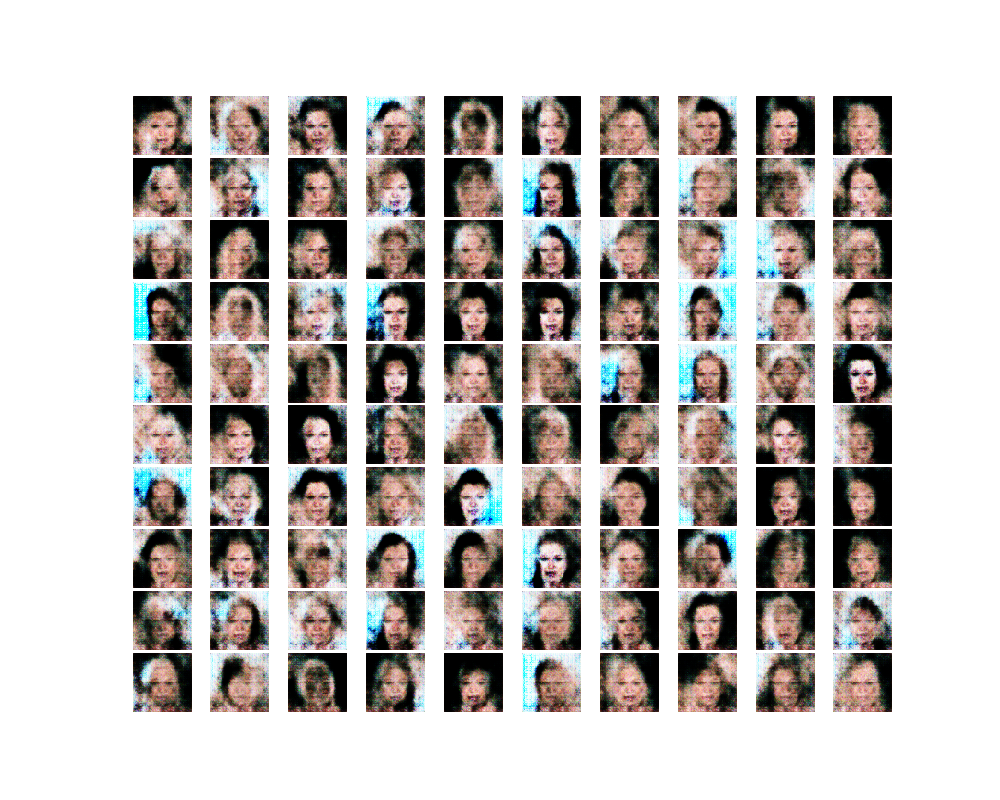}}
    \hfil
    \subfloat[CODASCA]{\includegraphics[trim={2cm 1.5cm 1.5cm 1.5cm}, clip, width=.31\linewidth]{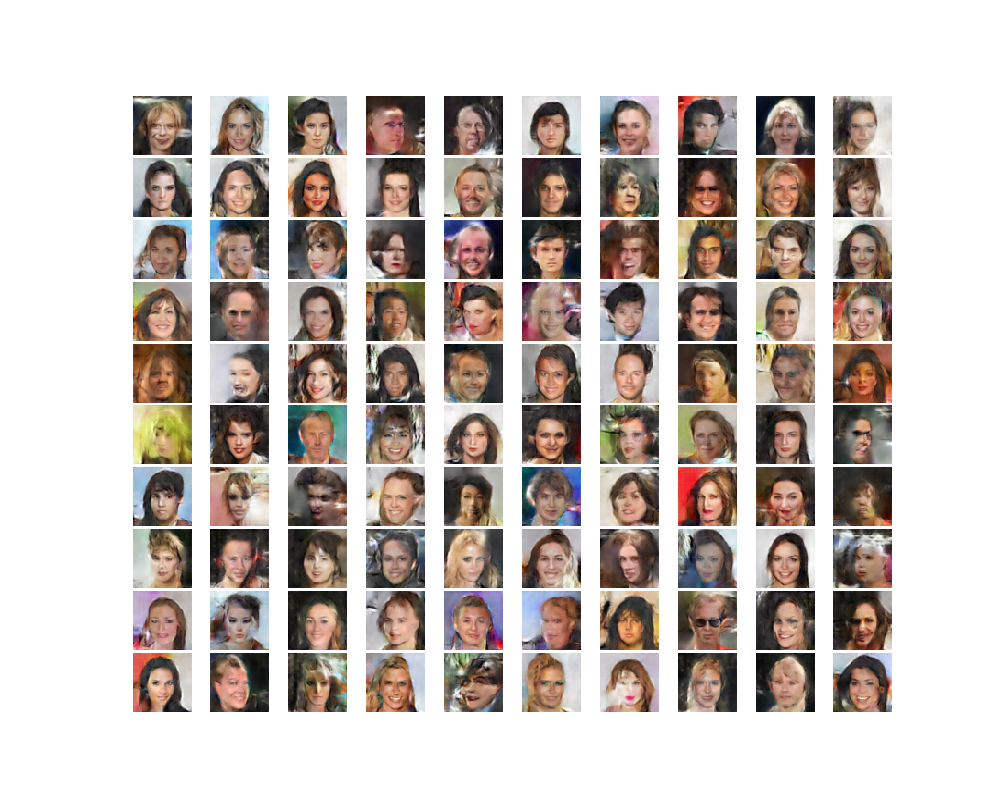}}
    \hfil
    \subfloat[CODA+]{\includegraphics[trim={2cm 1.5cm 1.5cm 1.5cm}, clip, width=.31\linewidth]{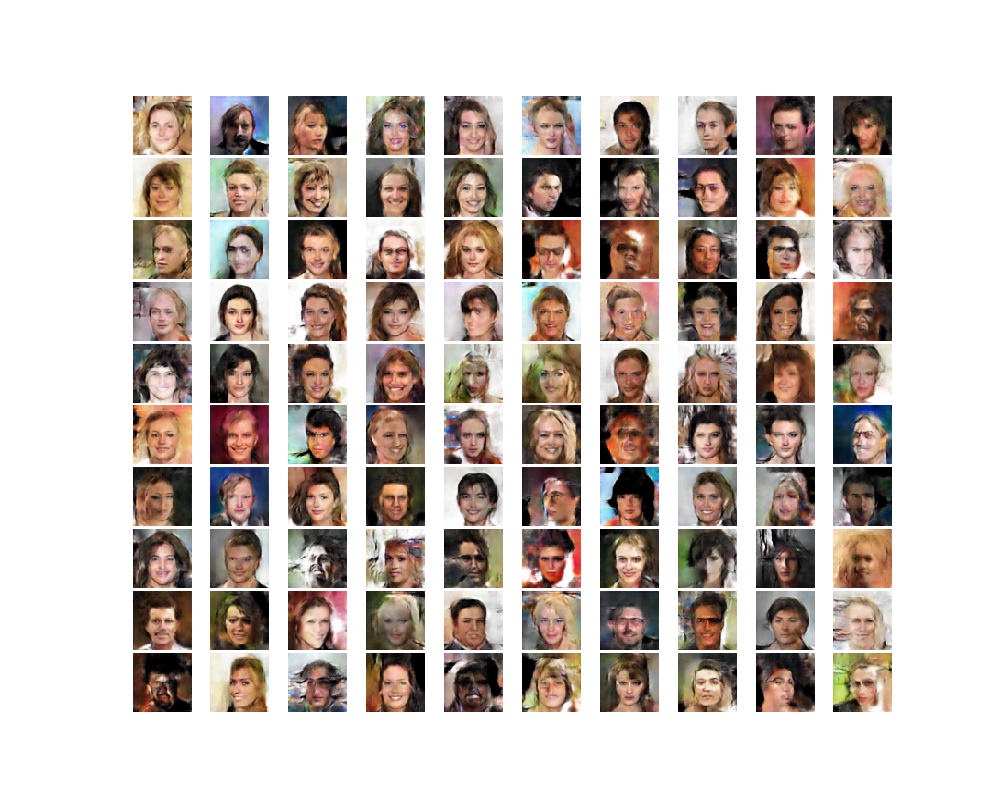}}
    \\
    \subfloat[Catalyst-Scaffold-S]{\includegraphics[trim={2cm 1.5cm 1.5cm 1.5cm}, clip, width=.31\linewidth]{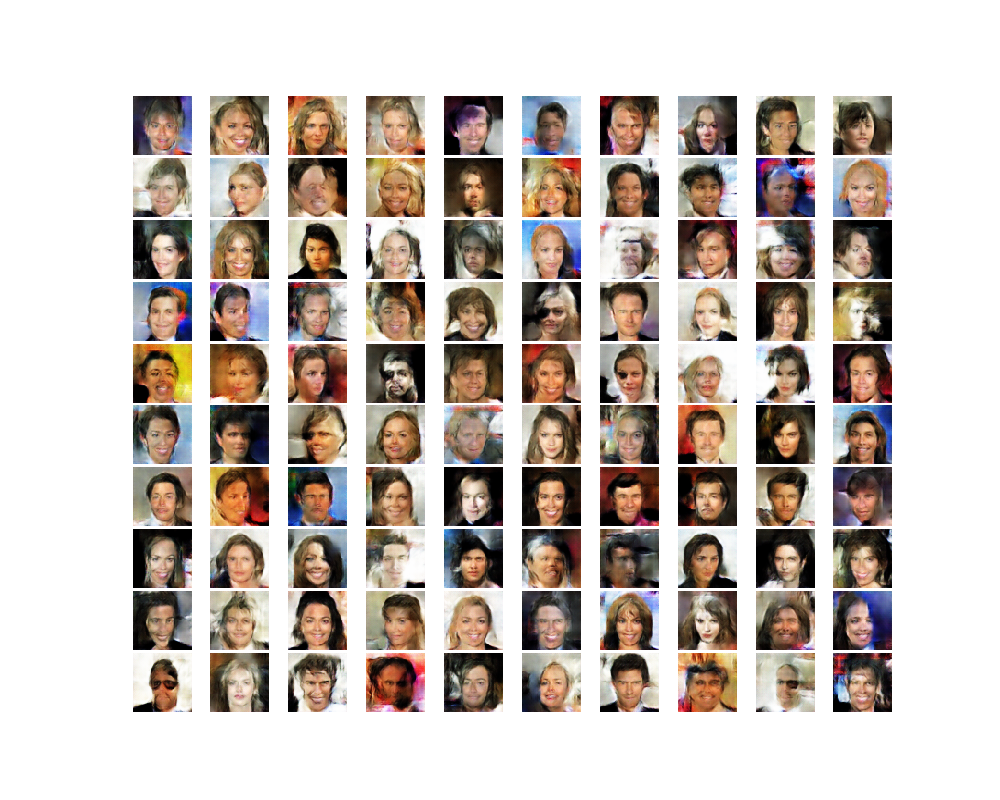}}
    \hfil
    \subfloat[Local SGDA+]{\includegraphics[trim={2cm 1.5cm 1.5cm 1.5cm}, clip, width=.31\linewidth]{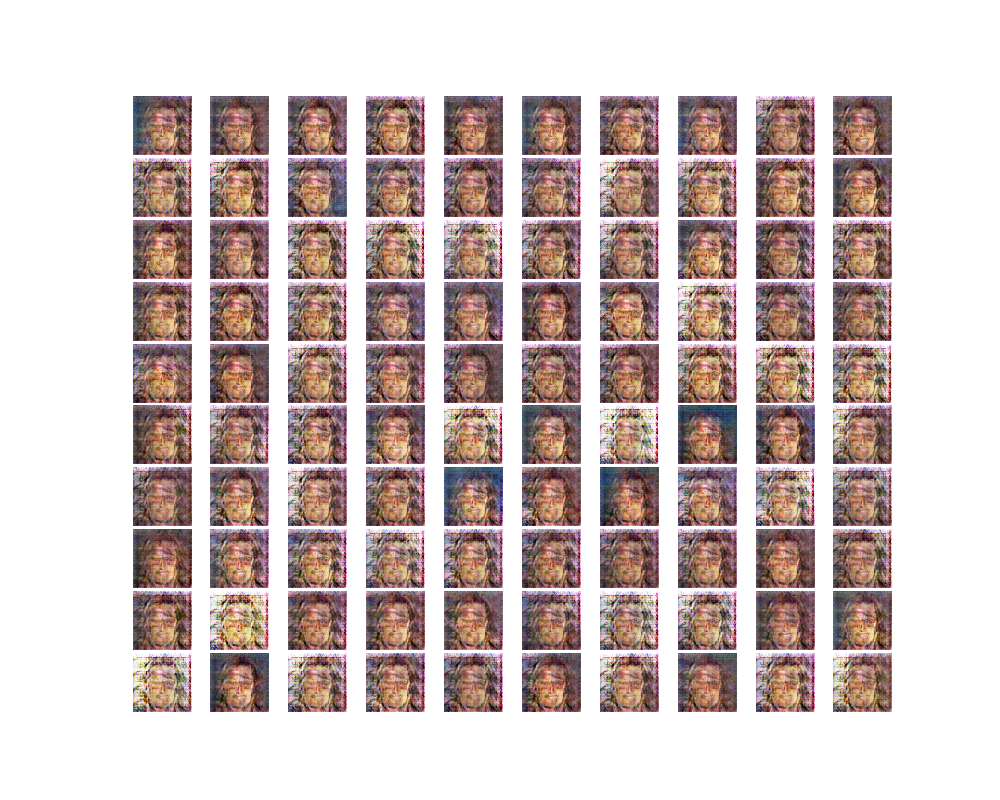}}
    \hfil
    \subfloat[Extra Step Local SGD]{\includegraphics[trim={2cm 1.5cm 1.5cm 1.5cm}, clip, width=.31\linewidth]{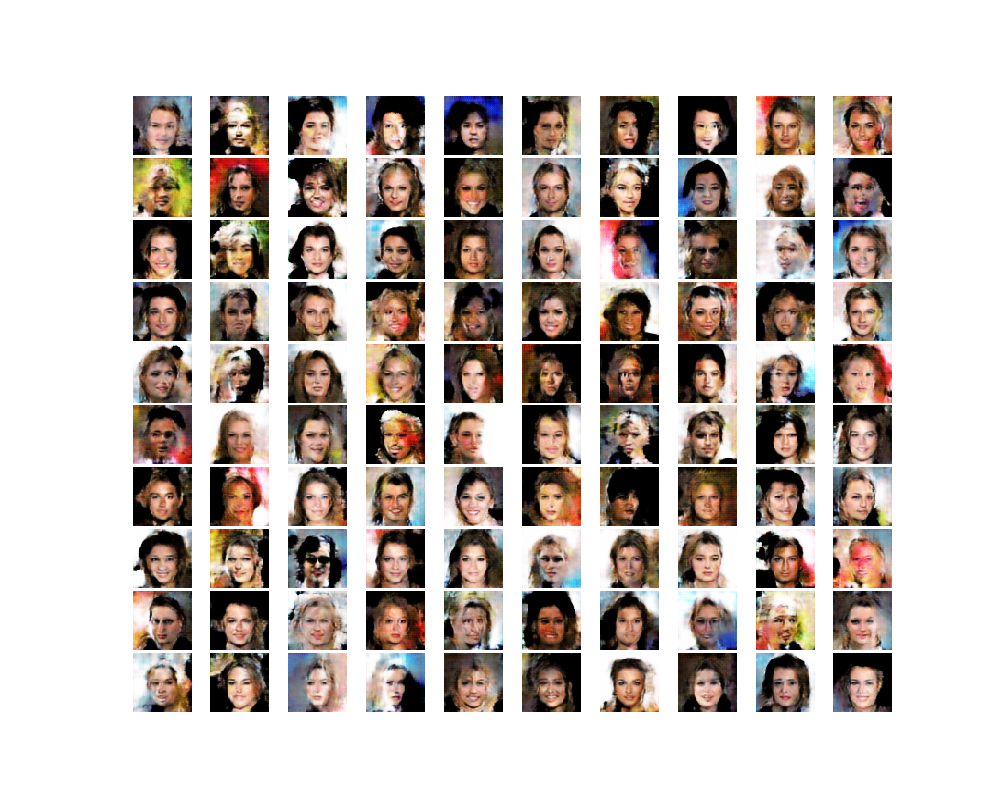}}
    \caption{
        Generative samples of the compared algorithms on CelebA.
        The samples are generated using the same set of random noise vectors.
    }
    \label{figure_generative_images_celeba}
\end{figure*}

Figure~\ref{figure_gan} and Figures~\ref{figure_generative_images}-\ref{figure_generative_images_celeba} complement the results of Table~\ref{table_gan}.
In Figure~\ref{figure_gan}, we present the training curve of each algorithm.
In Figures~\ref{figure_generative_images}-\ref{figure_generative_images_celeba}, we visualize sample images generated by all the compared algorithms after training.
We can see that both {\AlgMB} and {\AlgSTORM} generate high-quality sample images.

\begin{figure}[h]
    \captionsetup[subfloat]{farskip=1pt,captionskip=1pt}
    \centering
    \subfloat{\includegraphics[trim={0cm 0cm 0cm 0cm}, clip, width=.49\linewidth]{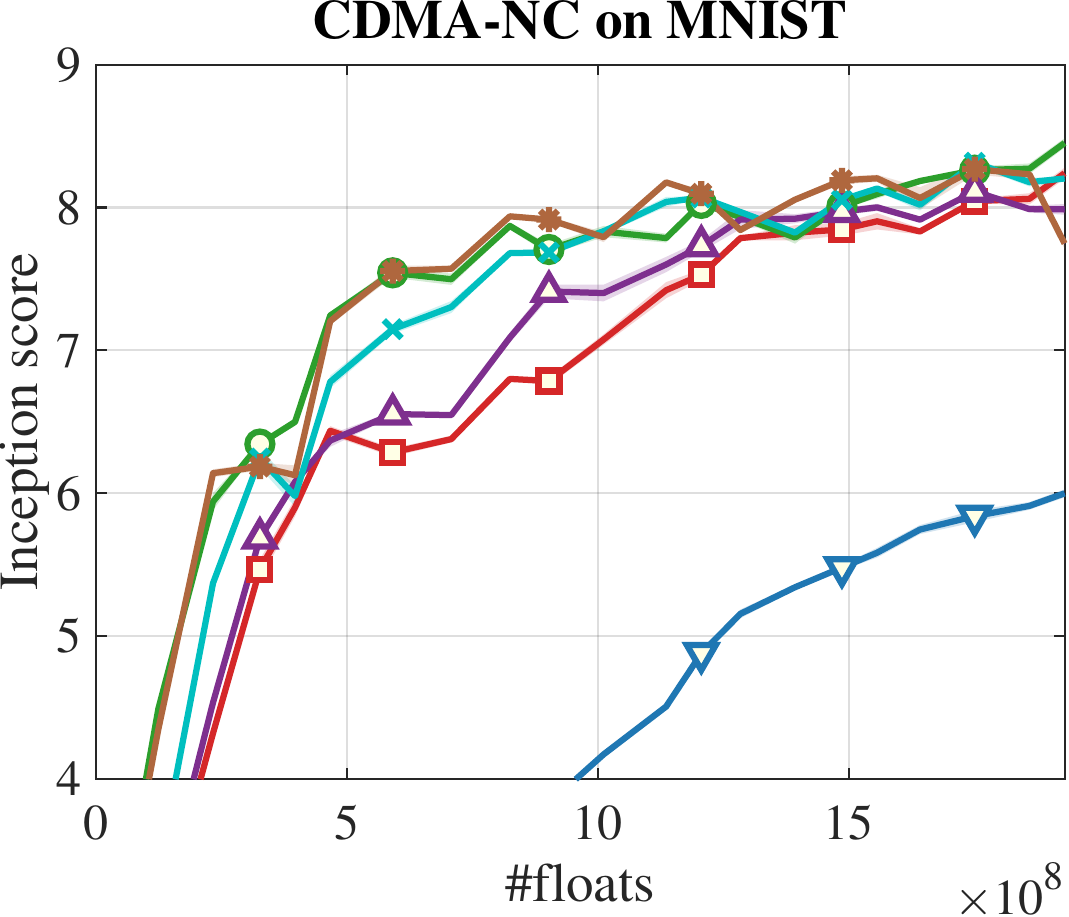}}
    \hfil
    \subfloat{\includegraphics[trim={0cm 0cm 0cm 0cm}, clip, width=.49\linewidth]{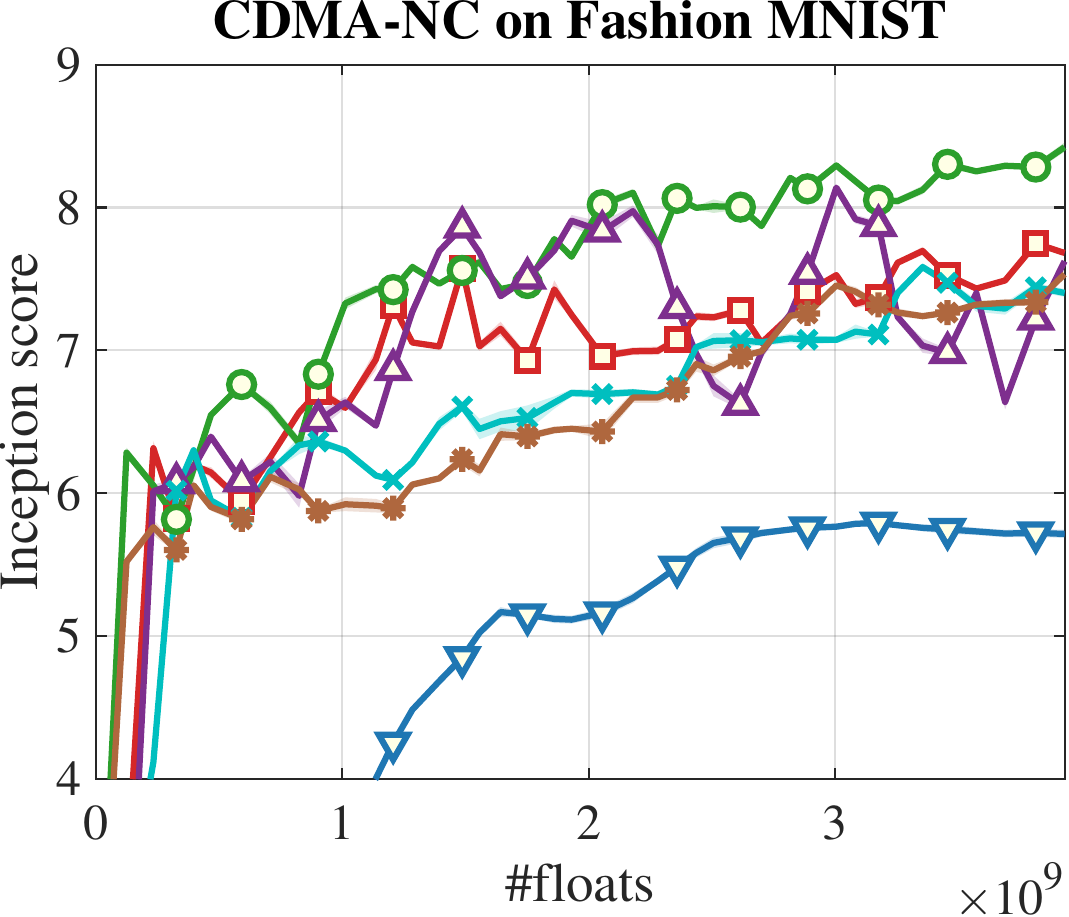}}
    \\[0.5em]
    \subfloat{\includegraphics[trim={0cm 0cm 0cm 0cm}, clip, width=.49\linewidth]{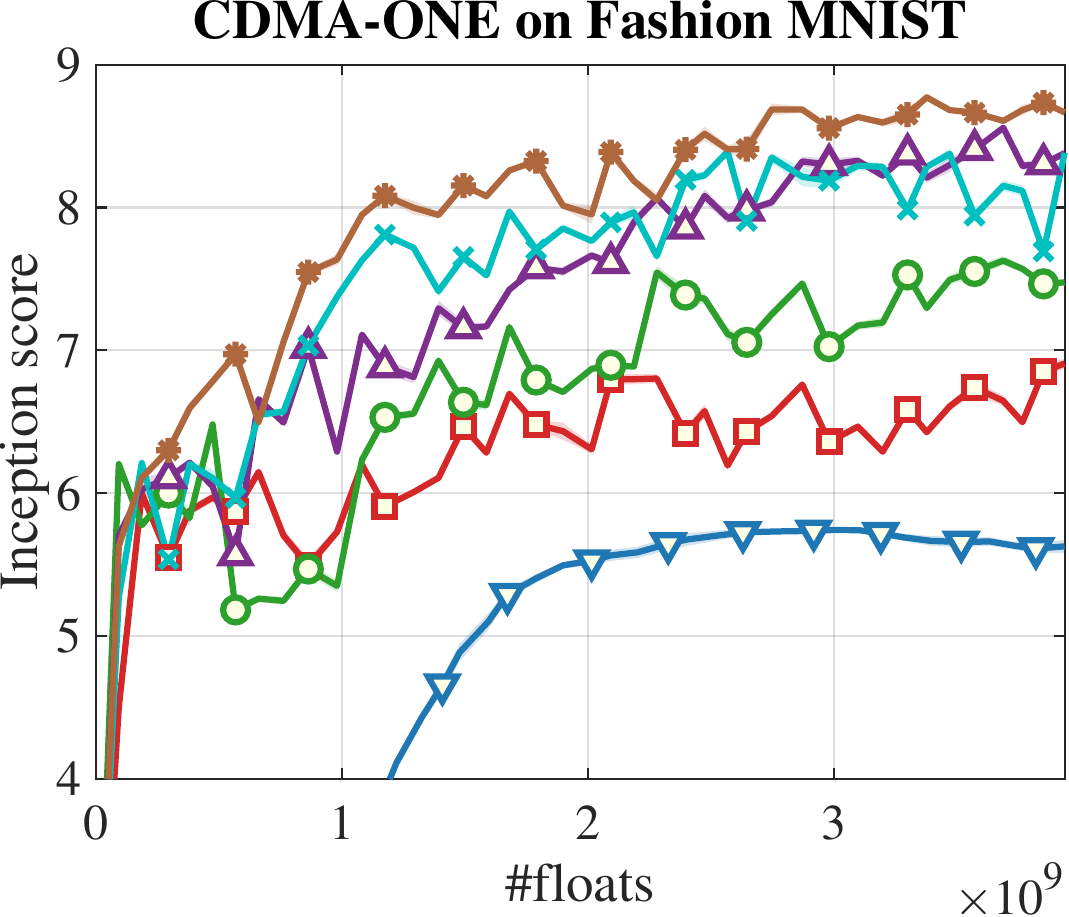}}
    \hfil
    \subfloat{\includegraphics[trim={0cm 0cm 0cm 0cm}, clip, width=.49\linewidth]{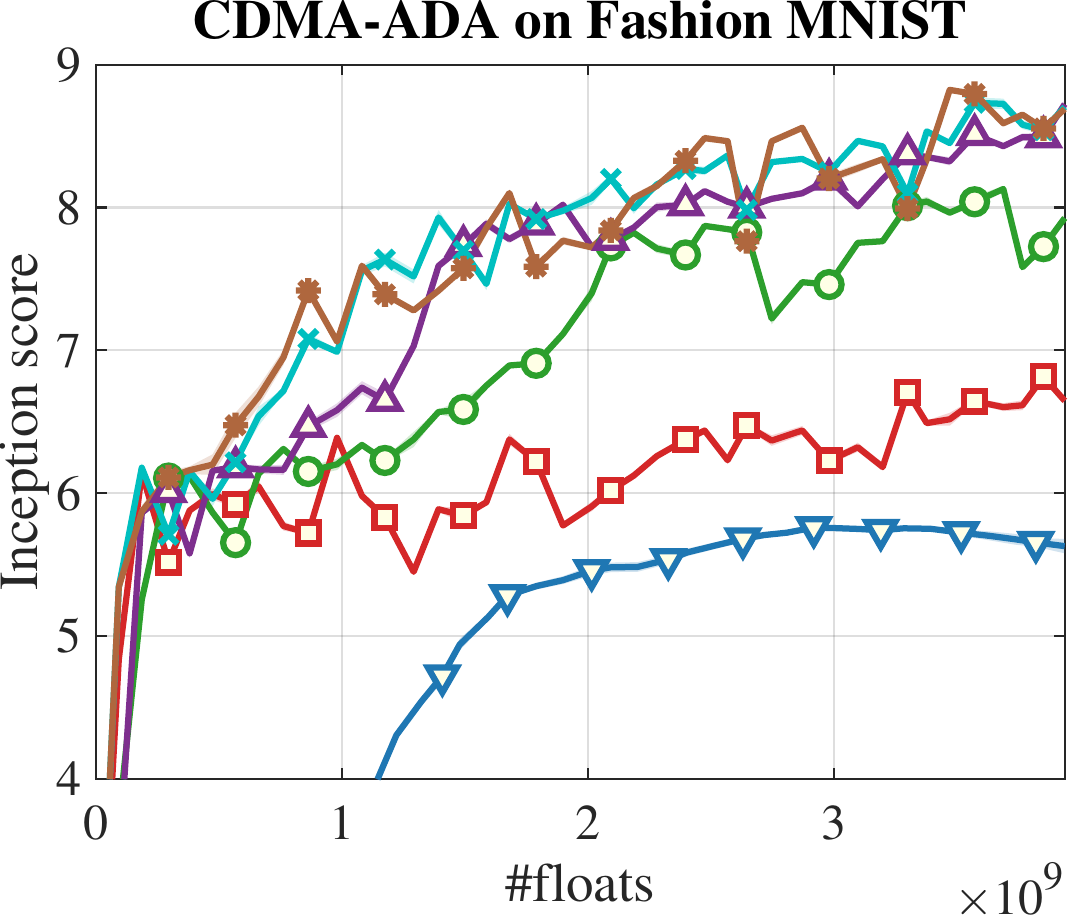}}
    \\[0.05em]
    \subfloat{\includegraphics[trim={-1.3cm 0cm 0cm 0cm}, clip, width=.98\linewidth]{./imgs_journal/legend_K}}
    \caption{
        Top: results of {\AlgNaive} with different values of $K$ for GAN training on MNIST and Fashion MNIST.
        Bottom: results of {\AlgMB} and {\AlgSTORM} with varying $K$ on Fashion MNIST. 
    }
    \label{figure_gan_K_more}
\end{figure}

Figure~\ref{figure_gan_K_more} presents additional results of the three versions of {\FrameworkAbbr} with different values of $K$, complementing the results presented in Figure~\ref{figure_gan_MNIST_K}.
The results in these two figures demonstrate that {\FrameworkAbbr} with multiple local steps performs significantly better than with only one local step.

\subsection{Specification of Hyperparameters}  \label{section_specification_of_hyperparameter}
The hyperparameters of {\AlgMB}, {\AlgNaive}, Parallel SGDA, and Local SGDA+ are the constant stepsizes $\eta$ and $\gamma$.
The Extra Step Local SGD algorithm has only one hyperparameter, the local step size $\eta_{\ell}$.
CODA+ involves three hyperparameters: the local step size $\eta_{\ell}$, the weight coefficient $\theta$ of the proximal-point subproblem, and the interval $K_0$ (in terms of the number of local updates) between two proximal point updates.
CODASCA and Catalyst-Scaffold-S have all the three hyperparameters of CODA+ as well as an additional global step size $\eta_{g}$.
Following~\citet{hou2021efficient}, we set $\eta_{g} = \eta_{\ell}$ for Catalyst-Scaffold-S.
The hyperparameters of {\AlgSTORM} are $\eta_t$, $\gamma_t$ and $\alpha_t$.
In our implementation, we set $\eta_t = \frac{c_{\eta}}{(t+1)^{\rho}}$, $\gamma_t = \frac{c_{\gamma}}{(t+1)^{\rho}}$, and $\alpha_t = \min\{1, \frac{c_{\alpha}}{(t+1)^{2 \rho}}\}$, where $c_{\eta}$, $c_{\gamma}$, $c_{\alpha}$, and $\rho$ are tunable parameters.
We note that Catalyst-Scaffold-S communicates with probability $p$ at each local step, while Parallel SGDA does not have local steps and other algorithms communicate after every $K$ local steps.
We set $p = 1 / K$ so that the average number of local steps of Catalyst-Scaffold-S in each round is equal to $K$,
where the value of $K$ has been specified in Section~\ref{section_experiments}.

\subsubsection{Hyperparameters for AUC maximization}

For the AUC maximization task, the hyperparameters are chosen as follows.
\begin{itemize}
    \item For {\AlgMB}, {\AlgNaive}, Parallel SGDA, and Local SGDA+, $\eta$ is chosen from $\{\num{1}, \num{7e-1}, \num{3.162e-1}, \num{2.5e-1}, \num{2e-1}, \num{1e-1}\}$\footnote{Note that $\num{3.162e-1} = 10^{-0.5}$.} and $\gamma$ is chosen from $\{\num{1}, \num{3.162e-1}, \num{1e-1}, \num{3.162e-2}, \num{1e-2}, \allowbreak \num{3.162e-3}, \num{1e-3}\}$.
    \item The step size $\eta_{\ell}$ of Extra Step Local SGD is tuned from $
    \{\num{1}, \num{3.162e-1}, \num{1e-1}, \num{3.162e-2}, \num{1e-2}, \num{3.162e-3}, \num{1e-3}, \allowbreak \num{3.162e-4}, \num{1e-4}\}$.
    \item For {\AlgSTORM}, $c_{\eta} \in \{\num{1}, \num{7e-1}, \num{3.162e-1}, \allowbreak \num{2.5e-1}, \num{2e-1}, \num{1e-1}\}$ and $c_{\gamma} \in \{\num{1}, \num{3.162e-1}, \allowbreak \num{1e-1}, \num{3.162e-2}, \num{1e-2}, \num{3.162e-3}, \num{1e-3}\}$, $c_{\alpha} \in \{0.95, 0.99, 5, 10\}$, and $\rho \in \{0, 1/5, 1/3\}$.
    We note that while our theory suggests that $\rho = 1/3$, in practice {\AlgSTORM} with $\rho=0$ or $\rho=1/5$ performs better in some cases.
    \item The hyperparameters $\eta_{\ell}$, $\theta$, and $K_0$ of CODA+, and Catalyst-Scaffold-S are chosen from $\{\num{1}, \num{3.162e-1}, \num{1e-1}, \num{3.162e-2}, \num{1e-2}\}$, $\{0, \num{1e-1}, \num{1e1}\}$, and $\{\num{2e3}, \num{4e3}\}$, respectively.
    \item For CODASCA, $\eta_{\ell}$, $\theta$, and $K_0$ are chosen in the same way as those of CODA+, and $\eta_g$ is tuned from $\{1.1, 1.0, 0.99\}$ following~\citet{yuan2021federated}.
\end{itemize}
Table~\ref{table_auc_parameters} shows the best hyperparameters for each algorithm on the AUC maximization experiment.

\subsubsection{Hyperparameters for robust adversarial network training}

For the robust adversarial network training task, the hyperparameters are chosen as follows.

\begin{itemize}
    \item For {\AlgMB}, {\AlgNaive}, Parallel SGDA, and Local SGDA+, $\eta$ is chosen from $\{\num{3.162e-2}, \num{1e-2}, \num{3.162e-3}, \num{1e-3}\}$ and $\gamma$ is chosen from $\{\num{1}, \num{3.162e-1}, \num{1e-1}, \num{3.162e-2}, \num{1e-3}\}$.
    \item The step size $\eta_{\ell}$ of Extra Step Local SGD is tuned from $
    \{\num{1}, \num{3.162e-1}, \num{1e-1}, \num{3.162e-2}, \num{1e-2}, \num{3.162e-3}, \num{1e-3}, \allowbreak \num{3.162e-4}, \num{1e-4}\}$.
    \item For {\AlgSTORM}, $c_{\eta} \in \{\num{3.162e-2}, \num{1e-2}, \num{3.162e-3}, , \num{1e-3}\}$ and $c_{\gamma} \in \{\num{1}, \num{3.162e-1}, \num{1e-1}, \num{3.162e-2}, \num{1e-2}\}$, $c_{\alpha} \in \{5, 10\}$, and $\rho \in \{1/5, 1/3\}$.
    \item The hyperparameters $\eta_{\ell}$, $\theta$, and $K_0$ of CODA+, and Catalyst-Scaffold-S are chosen from $\{\num{1e-1}, \num{3.162e-2}, \num{1e-2}, \num{3.162e-3}, \num{1e-3}\}$, $\{0, \num{1e-1}, \num{1e1}\}$, and $\{\num{2e3}, \num{4e3}\}$, respectively.
    \item For CODASCA, $\eta_{\ell}$, $\theta$, and $K_0$ are chosen in the same way as those of CODA+, and $\eta_g$ is tuned from $\{1.1, 1.0, 0.99\}$ following~\citet{yuan2021federated}.
\end{itemize}
Table~\ref{table_robustnn_parameters} shows the best hyperparameters for each algorithm on the robust adversarial network training experiment.

\subsubsection{Hyperparameters for GAN training}

For the GAN training task, the hyperparameters are chosen as follows.

\begin{itemize}
    \item For {\AlgMB}, {\AlgNaive}, Parallel SGDA, and Local SGDA+, both $\eta$ and $\gamma$ are chosen from $\{\num{1e-2}, \num{1e-3}, \num{1e-4}\}$.
    \item The step size $\eta_{\ell}$ of Extra Step Local SGD is tuned from $
    \{\num{1e-2}, \num{1e-3}, \num{1e-4}, \num{1e-5}\}$.
    \item For {\AlgSTORM}, both $c_{\eta}$ and $c_{\gamma}$ are selected from $\{\num{1e-2}, \num{1e-3}, \num{1e-4}\}$, $c_{\alpha} \in \{0.99, 5, 10\}$, and $\rho \in \{0, 1/5, 1/3\}$.
    \item The hyperparameters $\eta_{\ell}$, $\theta$, and $K_0$ of CODA+, and Catalyst-Scaffold-S are chosen from $\{\num{1e-2}, \num{1e-3}, \num{1e-4}\}$, $\{0, \num{1e-1}, \num{1e1}\}$, and $\{\num{2e3}, \num{4e3}\}$, respectively.
    \item For CODASCA, $\eta_{\ell}$, $\theta$, and $K_0$ are chosen in the same way as those of CODA+, and $\eta_g$ is tuned from $\{1.1, 1.0, 0.99\}$ following~\citet{yuan2021federated}.
\end{itemize}
Table~\ref{table_gan_parameters} and~\ref{table_gan_parameters_celeba} (resp., Table~\ref{table_gan_parameters_K}) summarize the best hyperparameters for each algorithm corresponding to the results in Figure~\ref{figure_gan} (resp., Figures~\ref{figure_gan_MNIST_K} and~\ref{figure_gan_K_more}).

\begin{table*}[!htb]
\centering
\caption{Best hyperparameters on the AUC maximization experiment.}
\label{table_auc_parameters}
\begin{tabular}{lll}
\toprule
Algorithm  & Best hyperparameters on MNIST & Best hyperparameters on CIFAR-10 \\
\midrule
{\AlgMB} & $\eta \!=\! \num{3.162e-1}$, $\gamma \!=\! \num{1}$ & $\eta \!=\! \num{2e-1}$, $\gamma \!=\! \num{3.162e-1}$ \\
{\AlgSTORM} & $c_{\eta} \!=\! \num{7e-1}$, $c_{\gamma} \!=\! \num{3.162e-1}$, $c_{\alpha} \!=\! 5$, $\rho \!=\! 1/5$ & $c_{\eta} \!=\! \num{2.5e-1}$, $c_{\gamma} \!=\! \num{3.162e-1}$, $c_{\alpha} \!=\! 0.95$, $\rho \!=\! 0$\\
{\AlgNaive} & $\eta \!=\! \num{3.162e-1}$, $\gamma \!=\! \num{1e-1}$ & $\eta \!=\! \num{1e-1}$, $\gamma \!=\! \num{1e-1}$ \\
Parallel SGDA & $\eta \!=\! \num{1}$, $\gamma \!=\! \num{1}$ & $\eta \!=\! \num{1}$, $\gamma \!=\! \num{3.162e-1}$ \\
CODASCA & $\eta_{\ell} \!=\! \num{3.162e-1}$, $\eta_{g} \!=\! \num{1}$, $\theta \!=\! 0$, $K_0 \!=\! \num{4e3}$ & $\eta_{\ell} \!=\! \num{3.162e-2}$, $\eta_{g} \!=\! \num{0.99}$, $\theta \!=\! 0$, $K_0 \!=\! \num{4e3}$ \\
CODA+ & $\eta_{\ell} \!=\! \num{3.162e-1}$, $\theta \!=\! 0$, $K_0 \!=\! \num{4e3}$ & $\eta_{\ell} \!=\! \num{1e-1}$, $\theta \!=\! 0$, $K_0 \!=\! \num{2e3}$ \\
Catalyst-Scaffold-S & $\eta_{\ell} \!=\! \num{1e-1}$, $\theta \!=\! 0$, $K_0 \!=\! \num{2e3}$ & $\eta_{\ell} \!=\! \num{3.162e-1}$, $\theta \!=\! 0$, $K_0 \!=\! \num{4e3}$ \\
Local SGDA+ & $\eta \!=\! \num{3.162e-1}$, $\gamma \!=\! \num{1e-1}$ & $\eta \!=\! \num{1e-1}$, $\gamma \!=\! \num{1e-1}$ \\
Extra Step Local SGD & $\eta_{\ell} \!=\! \num{3.162e-1}$ & $\eta_{\ell} \!=\! \num{3.162e-1}$ \\
\bottomrule
\end{tabular}
\end{table*}

\begin{table*}[!htb]
\centering
\caption{Best hyperparameters on the robust adversarial network training experiment.}
\label{table_robustnn_parameters}
\begin{tabular}{lll}
\toprule
Algorithm  & Best hyperparameters on MNIST & Best hyperparameters on Fashion MNIST \\
\midrule
{\AlgMB} & $\eta \!=\! \num{1e-3}$, $\gamma \!=\! \num{3.162e-1}$ & $\eta \!=\! \num{1e-3}$, $\gamma \!=\! \num{1e-1}$ \\
{\AlgSTORM} & $c_{\eta} \!=\! \num{1e-2}$, $c_{\gamma} \!=\! \num{1}$, $c_{\alpha} \!=\! 5$, $\rho \!=\! 1/3$ & $c_{\eta} \!=\! \num{1e-2}$, $c_{\gamma} \!=\! \num{1}$, $c_{\alpha} \!=\! 5$, $\rho \!=\! 1/3$ \\
{\AlgNaive} & $\eta \!=\! \num{3.162e-4}$, $\gamma \!=\! \num{3.162e-2}$ & $\eta \!=\! \num{1e-3}$, $\gamma \!=\! \num{3.162e-2}$ \\
Parallel SGDA & $\eta \!=\! \num{1e-2}$, $\gamma \!=\! \num{1}$ & $\eta \!=\! \num{1e-2}$, $\gamma \!=\! \num{1e-1}$ \\
CODASCA & $\eta_{\ell} \!=\! \num{3.162e-3}$, $\eta_{g} \!=\! \num{1}$, $\theta \!=\! \num{0}$, $K_0 \!=\! \num{2e3}$ & $\eta_{\ell} \!=\! \num{3.162e-3}$, $\eta_{g} \!=\! \num{1.1}$, $\theta \!=\! \num{1e-1}$, $K_0 \!=\! \num{4e3}$ \\
CODA+ & $\eta_{\ell} \!=\! \num{1e-3}$, $\theta \!=\! \num{0}$, $K_0 \!=\! \num{2e3}$ & $\eta_{\ell} \!=\! \num{1e-3}$, $\theta \!=\! \num{1e-1}$, $K_0 \!=\! \num{4e3}$ \\
Catalyst-Scaffold-S & $\eta_{\ell} \!=\! \num{1e-3}$, $\theta \!=\! \num{1e-1}$, $K_0 \!=\! \num{4e3}$ & $\eta_{\ell} \!=\! \num{1e-3}$, $\theta \!=\! \num{0}$, $K_0 \!=\! \num{4e3}$ \\
Local SGDA+ & $\eta \!=\! \num{3.162e-4}$, $\gamma \!=\! \num{3.162e-2}$ & $\eta \!=\! \num{1e-3}$, $\gamma \!=\! \num{3.162e-2}$ \\
Extra Step Local SGD & $\eta_{\ell} \!=\! \num{3.162e-4}$ & $\eta_{\ell} \!=\! \num{3.162e-4}$ \\ 
\bottomrule
\end{tabular}
\end{table*}

\begin{table*}[!htb]
\centering
\caption{Best hyperparameters for the GAN training experiment on MNIST and Fashion MNIST corresponding to the results in Table~\ref{table_gan} and Figure~\ref{figure_gan}.}
\label{table_gan_parameters}
\begin{tabular}{lll}
\toprule
Algorithm  & Best hyperparameters on MNIST & Best hyperparameters on Fashion MNIST \\
\midrule
{\AlgMB} & $\eta \!=\! \num{1e-2}$, $\gamma \!=\! \num{1e-2}$ & $\eta \!=\! \num{1e-2}$, $\gamma \!=\! \num{1e-3}$ \\
{\AlgSTORM} & $c_{\eta} \!=\! \num{1e-2}$, $c_{\gamma} \!=\! \num{1e-2}$, $c_{\alpha} \!=\! 0.99$, $\rho \!=\! 0$ & $c_{\eta} \!=\! \num{1e-2}$, $c_{\gamma} \!=\! \num{1e-3}$, $c_{\alpha} \!=\! 0.99$, $\rho \!=\! 0$ \\
{\AlgNaive} & $\eta \!=\! \num{1e-2}$, $\gamma \!=\! \num{1e-3}$ & $\eta \!=\! \num{1e-3}$, $\gamma \!=\! \num{1e-3}$ \\
Parallel SGDA & $\eta \!=\! \num{1e-2}$, $\gamma \!=\! \num{1e-2}$ & $\eta \!=\! \num{1e-3}$, $\gamma \!=\! \num{1e-2}$ \\
CODASCA & $\eta_{\ell} \!=\! \num{1e-2}$, $\eta_{g} \!=\! \num{1}$, $\theta \!=\! \num{1e-1}$, $K_0 \!=\! \num{4e3}$ & $\eta_{\ell} \!=\! \num{1e-2}$, $\eta_{g} \!=\! \num{1.1}$, $\theta \!=\! \num{0}$, $K_0 \!=\! \num{4e3}$ \\
CODA+ & $\eta_{\ell} \!=\! \num{1e-2}$, $\theta \!=\! 0$, $K_0 \!=\! \num{4e3}$ & $\eta_{\ell} \!=\! \num{1e-2}$, $\theta \!=\! \num{0}$, $K_0 \!=\! \num{4e3}$ \\
Catalyst-Scaffold-S & $\eta_{\ell} \!=\! \num{1e-2}$, $\theta \!=\! 0$, $K_0 \!=\! \num{2e3}$ & $\eta_{\ell} \!=\! \num{1e-2}$, $\theta \!=\! \num{0}$, $K_0 \!=\! \num{4e3}$ \\
Local SGDA+ & $\eta \!=\! \num{1e-2}$, $\gamma \!=\! \num{1e-3}$ & $\eta \!=\! \num{1e-3}$, $\gamma \!=\! \num{1e-3}$\\
Extra Step Local SGD & $\eta_{\ell} \!=\! \num{1e-3}$ & $\eta_{\ell} \!=\! \num{1e-3}$ \\
\bottomrule
\end{tabular}
\end{table*}

\begin{table*}[!htb]
\centering
\caption{Best hyperparameters for the GAN training experiment on CelebA corresponding to the results in Table~\ref{table_gan} and Figure~\ref{figure_gan}.}
\label{table_gan_parameters_celeba}
\begin{tabular}{lll}
\toprule
Algorithm  & Best hyperparameters on CelebA \\
\midrule
{\AlgMB} & $K=20$, $\eta \!=\! \num{1e-2}$, $\gamma \!=\! \num{1e-2}$ \\
{\AlgSTORM} & $K=25$, $c_{\eta} \!=\! \num{1e-2}$, $c_{\gamma} \!=\! \num{1e-2}$, $c_{\alpha} \!=\! 0.99$, $\rho \!=\! 0$  \\
{\AlgNaive} & $K=20$, $\eta \!=\! \num{1e-3}$, $\gamma \!=\! \num{1e-2}$  \\
Parallel SGDA & $\eta \!=\! \num{1e-2}$, $\gamma \!=\! \num{1e-2}$ \\
CODASCA & $K=25$, $\eta_{\ell} \!=\! \num{1e-2}$, $\eta_{g} \!=\! \num{1.1}$, $\theta \!=\! \num{0}$, $K_0 \!=\! \num{4e3}$ \\
CODA+ & $K=20$, $\eta_{\ell} \!=\! \num{1e-2}$, $\theta \!=\! 0$, $K_0 \!=\! \num{4e3}$ \\
Catalyst-Scaffold-S & $K=20$, $\eta_{\ell} \!=\! \num{1e-2}$, $\theta \!=\! 0$, $K_0 \!=\! \num{2e3}$ \\
Local SGDA+ & $K=20$, $\eta \!=\! \num{1e-3}$, $\gamma \!=\! \num{1e-2}$ \\
Extra Step Local SGD & $K=10$, $\eta_{\ell} \!=\! \num{1e-2}$ \\
\bottomrule
\end{tabular}
\end{table*}

\begin{table*}[!htb]
\centering
\caption{Best hyperparameters for the GAN training experiment corresponding to the results in Figures~\ref{figure_gan_MNIST_K} and~\ref{figure_gan_K_more}.}
\label{table_gan_parameters_K}
\begin{tabular}{llll}
\toprule
Algorithm  & $K$ & Best hyperparameters on MNIST & Best hyperparameters on Fashion MNIST \\
\midrule
\multirow{6}{*}{{\AlgMB}} & $K=1$ & $\eta \!=\! \num{1e-3}$, $\gamma \!=\! \num{1e-2}$ & $\eta \!=\! \num{1e-2}$, $\gamma \!=\! \num{1e-2}$ \\
 & $K=5$ & $\eta \!=\! \num{1e-2}$, $\gamma \!=\! \num{1e-2}$ & $\eta \!=\! \num{1e-2}$, $\gamma \!=\! \num{1e-2}$ \\
 & $K=10$ & $\eta \!=\! \num{1e-2}$, $\gamma \!=\! \num{1e-2}$ & $\eta \!=\! \num{1e-2}$, $\gamma \!=\! \num{1e-2}$ \\
 & $K=15$ & $\eta \!=\! \num{1e-2}$, $\gamma \!=\! \num{1e-2}$ & $\eta \!=\! \num{1e-2}$, $\gamma \!=\! \num{1e-2}$ \\
 & $K=20$ & $\eta \!=\! \num{1e-2}$, $\gamma \!=\! \num{1e-3}$ & $\eta \!=\! \num{1e-2}$, $\gamma \!=\! \num{1e-2}$ \\
 & $K=25$ & $\eta \!=\! \num{1e-2}$, $\gamma \!=\! \num{1e-2}$ & $\eta \!=\! \num{1e-2}$, $\gamma \!=\! \num{1e-2}$ \\
\midrule
\multirow{6}{*}{{\AlgSTORM}} & $K=1$ & $c_{\eta} \!=\! \num{1e-3}$, $c_{\gamma} \!=\! \num{1e-2}$, $c_{\alpha} \!=\! 0.99$, $\rho \!=\! 0$ & $c_{\eta} \!=\! \num{1e-2}$, $c_{\gamma} \!=\! \num{1e-2}$, $c_{\alpha} \!=\! 0.99$, $\rho \!=\! 0$ \\
 & $K=5$ & $c_{\eta} \!=\! \num{1e-2}$, $c_{\gamma} \!=\! \num{1e-2}$, $c_{\alpha} \!=\! 0.99$, $\rho \!=\! 0$ & $c_{\eta} \!=\! \num{1e-2}$, $c_{\gamma} \!=\! \num{1e-2}$, $c_{\alpha} \!=\! 0.99$, $\rho \!=\! 0$ \\
 & $K=10$ & $c_{\eta} \!=\! \num{1e-2}$, $c_{\gamma} \!=\! \num{1e-3}$, $c_{\alpha} \!=\! 0.99$, $\rho \!=\! 0$ & $c_{\eta} \!=\! \num{1e-2}$, $c_{\gamma} \!=\! \num{1e-2}$, $c_{\alpha} \!=\! 0.99$, $\rho \!=\! 0$ \\
 & $K=15$ & $c_{\eta} \!=\! \num{1e-2}$, $c_{\gamma} \!=\! \num{1e-3}$, $c_{\alpha} \!=\! 0.99$, $\rho \!=\! 0$ & $c_{\eta} \!=\! \num{1e-2}$, $c_{\gamma} \!=\! \num{1e-2}$, $c_{\alpha} \!=\! 0.99$, $\rho \!=\! 0$ \\
 & $K=20$ & $c_{\eta} \!=\! \num{1e-2}$, $c_{\gamma} \!=\! \num{1e-3}$, $c_{\alpha} \!=\! 0.99$, $\rho \!=\! 0$ & $c_{\eta} \!=\! \num{1e-2}$, $c_{\gamma} \!=\! \num{1e-2}$, $c_{\alpha} \!=\! 0.99$, $\rho \!=\! 0$ \\
 & $K=25$ & $c_{\eta} \!=\! \num{1e-2}$, $c_{\gamma} \!=\! \num{1e-3}$, $c_{\alpha} \!=\! 0.99$, $\rho \!=\! 0$ & $c_{\eta} \!=\! \num{1e-2}$, $c_{\gamma} \!=\! \num{1e-2}$, $c_{\alpha} \!=\! 0.99$, $\rho \!=\! 0$ \\
\midrule
\multirow{6}{*}{{\AlgNaive}} & $K=1$ & $\eta \!=\! \num{1e-2}$, $\gamma \!=\! \num{1e-2}$ & $\eta \!=\! \num{1e-3}$, $\gamma \!=\! \num{1e-2}$ \\
 & $K=5$ & $\eta \!=\! \num{1e-2}$, $\gamma \!=\! \num{1e-2}$ & $\eta \!=\! \num{1e-2}$, $\gamma \!=\! \num{1e-2}$ \\
 & $K=10$ & $\eta \!=\! \num{1e-2}$, $\gamma \!=\! \num{1e-2}$ & $\eta \!=\! \num{1e-2}$, $\gamma \!=\! \num{1e-2}$ \\
 & $K=15$ & $\eta \!=\! \num{1e-2}$, $\gamma \!=\! \num{1e-3}$ & $\eta \!=\! \num{1e-2}$, $\gamma \!=\! \num{1e-3}$ \\
 & $K=20$ & $\eta \!=\! \num{1e-2}$, $\gamma \!=\! \num{1e-3}$ & $\eta \!=\! \num{1e-3}$, $\gamma \!=\! \num{1e-3}$ \\
 & $K=25$ & $\eta \!=\! \num{1e-2}$, $\gamma \!=\! \num{1e-3}$ & $\eta \!=\! \num{1e-3}$, $\gamma \!=\! \num{1e-2}$ \\
\bottomrule
\end{tabular}
\end{table*}

\end{document}